# Support Vector Machine Guided Reproducing Kernel Particle Method for Image-Based Modeling of Microstructures

**Yanran Wang [1], Jonghyuk Baek [1], Yichun Tang [2], Jing Du [2], Mike Hillman [3], J. S. Chen [1,*]**

[1] Department of Structural Engineering, University of California San Diego, La Jolla, CA 92093-0085, USA

[2] Department of Mechanical Engineering, Pennsylvania State University, University Park, PA 16802, USA

[3] Karagozian & Case, Inc., Glendale, CA 91203, USA

## Abstract

This work presents an approach for automating the discretization and approximation procedures in constructing digital representations of composites from Micro-CT images featuring intricate microstructures. The proposed method is guided by the Support Vector Machine (SVM) classification, offering an effective approach for discretizing microstructural images. An SVM soft margin training process is introduced as a classification of heterogeneous material points, and image segmentation is accomplished by identifying support vectors through a local regularized optimization problem. In addition, an Interface-Modified Reproducing Kernel Particle Method (IM-RKPM) is proposed for appropriate approximations of weak discontinuities across material interfaces. The proposed method modifies the smooth kernel functions with a regularized heavy-side function concerning the material interfaces to alleviate Gibb's oscillations. This IM-RKPM is formulated without introducing duplicated degrees of freedom associated with the interface nodes commonly needed in the conventional treatments of weak discontinuities in the meshfree methods. Moreover, IM-RKPM can be implemented with various domain integration techniques, such as Stabilized Conforming Nodal Integration (SCNI). The extension of the proposed method to 3-dimension is straightforward, and the effectiveness of the proposed method is validated through the image-based modeling of polymer-ceramic composite microstructures.

* Corresponding author.

*E-mail address*: js-chen@eng.ucsd.edu (J.S. Chen).

# 1 Introduction

In recent years, a variety of non-destructive imaging techniques, such as micro-X-ray computed tomography (micro-CT), have emerged as powerful alternatives to obtain detailed information about the microstructure and internal deformation of composite materials [1]–[4]. Nevertheless, modeling microstructures remains challenging owing to their geometrical and topological complexities and heterogeneity, making the body-fitted mesh generation for mesh-based methods extremely tedious and time-consuming, especially in the three-dimension model construction. An example of a 2D slice of micro-CT image of a polymer-ceramic composite specimen (polymer matrix reinforced by ceramic particles) and its corresponding body-fitted finite element mesh is shown in Figure 1, demonstrating the meshing complexity.

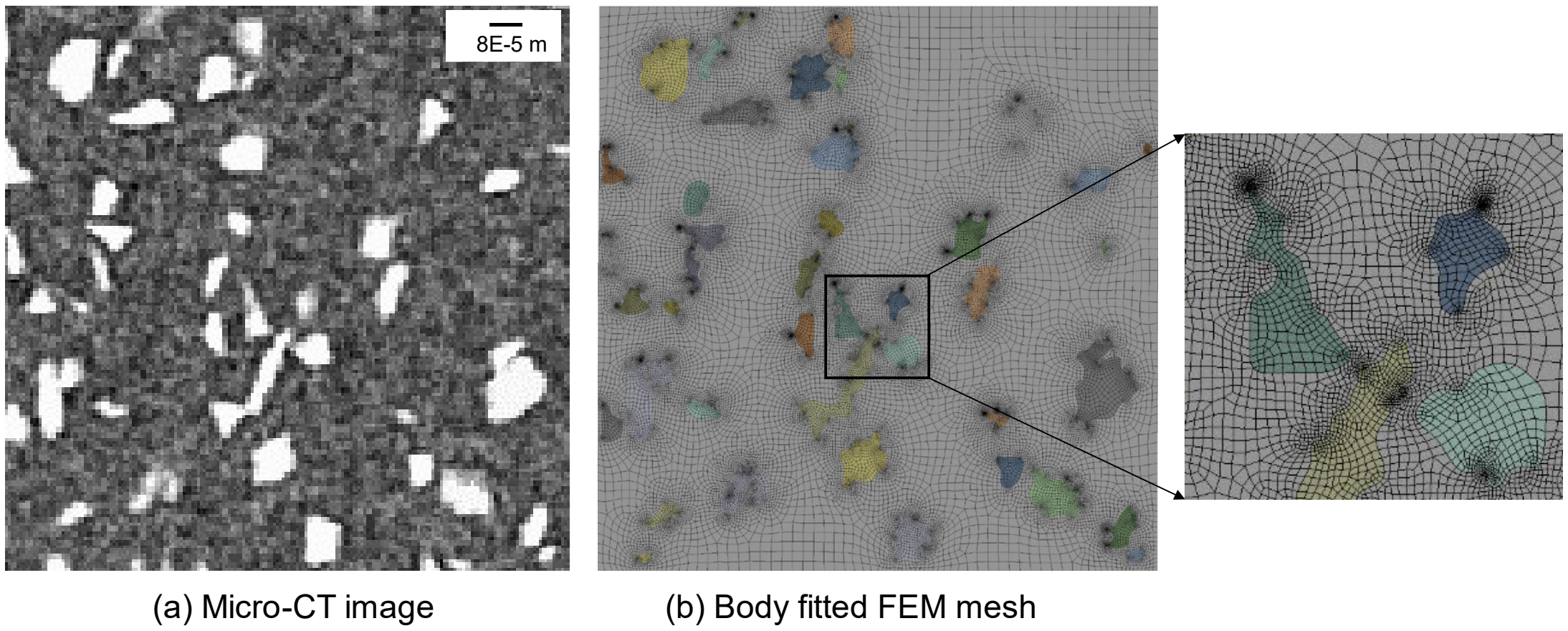

**Figure 1: Micro-CT image of a polymer-ceramic composite microstructure and its corresponding body-fitted finite element mesh**

Various image segmentation techniques have been developed over the past several decades, including region-based and classification-based methods [5]. Global and local thresholding [6] is a simple region-based method that uses a threshold value to separate objects from the background, but it can lead to poor results if the threshold is not chosen correctly. The region growing method [7] is another region-based approach that relies on user-selected seed pixels and offers advantages over thresholding, but the numerical results can be sensitive to the selection of initial seed points. On the contrary, classification-based methods generally adopt a global approach for image segmentation, whereby an automatic pattern recognition process is utilized in the context of supervised learning based on manually segmented training datasets. K-Nearest Neighbor (KNN) is a simple, non-parametric supervised learning model that makes predictions based on the k-nearest neighbors in the training data, but it usually requires a large amount of training data to suppress high variance problems [8]. Tree-based algorithms, including decision trees and random forests, are another widely used supervised learning techniques in which the training data is partitioned into smaller subsets without much data pre-processing and with high interpretability and computational efficiency. However, these methods may have limited ability to predict unseen data, restricted decision boundary expressiveness, and can be sensitive to imbalanced data [9]. Recently, deep learning algorithms have enabled to develop state-of-the-art image segmentation methods, especially those based on convolutional neural networks (CNNs), which can automatically learn features from raw images with minimal human interaction. However, these methods require large amounts of labeled datasets with extensive training and are mathematically more challenging to interpret due to the highly non-linear relationships between input features and output labels [10], [11].

The present work employs the Support Vector Machine (SVM) algorithms as the image segmentation method to guide the numerical model generation. SVM is a classification-based machine learning algorithm built on solid mathematical foundation and optimization frameworks [12], [13]. Compared to other supervised algorithms, SVM is advantageous because it generates a unique maximum-margined global hyperplane for separating training datasets, providing a global solution for data classification. Additionally, it is not sensitive to the underlying probabilistic distribution of the training dataset, ensuring high performance for limited, noisy, or imbalanced datasets [14]. One apparent limitation of the standard SVM is that it requires $O(l^3)$ operations, where $l$ is the length of the training dataset, to solve a complex quadratic programming problem (QPP) with inequality constraints. Various approaches have been proposed to overcome this limitation, such as the training decomposition method [15] and the reduced support vector machine (R-SVM) algorithm [16], which significantly improves SVM's training speed. Additionally, more efficient formulations of SVM have been introduced, such as the Least Square SVM (LS-SVM) algorithm [17] and the Lagrangian SVM algorithm [18]. The LS-SVM algorithm optimizes a dual problem directly using a least-square loss function, replacing the hinge loss function in the original SVM's formulation to reformulate the complex QPP as a linear system of equations. In contrast, the Lagrangian SVM algorithm utilizes an implicit Lagrangian for the dual of the standard quadratic program of a linear SVM, leading to the minimization of an unconstrained differentiable convex function in the space of dimensionality equal to the number of training datasets. Both mentioned algorithms eliminate the necessity of complicated programming problem solvers, making them feasible for classifying large datasets. In addition to the binary SVM classifier, extensive research has been done to extend SVM to multi-class classification. The one-vs-all (OVA) method, one-vs-one (OVO) method, error-correcting output codes (ECOC), and directed acyclic graphs (DAGs) are among the most widely used approaches to handle multi-class classification with SVM [19]. The traditional binary SVM algorithm is adopted in this work for its effective applicability to the two-phase materials.

Numerical modeling of heterogeneous materials remains challenging for both mesh-based methods discretized with body-fitted discretization and meshfree methods formulated with smooth approximations. For the Finite Element Method (FEM), incomplete handling of discontinuities in mesh construction can lead to suboptimal convergence [20], and aligning meshes with interfaces is a non-trivial task for composites with complex microstructures and significant variations in constituent moduli. The meshfree methods utilize point-wise discretization instead of carefully constructed body-fitted meshes. However, meshfree methods such as element-free Galerkin (EFG) [21] and reproducing kernel particle method (RKPM) [22]–[24] typically suffer from Gibb's-like oscillation in the approximation when modeling weak continuities in composite materials, as their smooth approximation functions with overlapping local supports fail to capture gradient jump conditions across material interfaces [25]. Considerable effort has been dedicated to developing effective techniques for dealing with interface discontinuities. Since the proposed work is under the Galerkin meshfree framework, the review of methods developed based on mesh-based context to address interface discontinuities is omitted here. Reviews on some key non body-fitted FEM developments for interface discontinuities can be found in [26], [27].

Two primary approaches in meshfree methods have been proposed for handling material interface weak discontinuities. The first approach involves introducing discontinuities in the meshless approximation function. Krongauz and Belytschko proposed two types of jump enrichment functions into the conventional Moving Least Squares (MLK) or Reproducing Kernel (RK) approximation of the field variables [25]. The enrichment functions introduce discontinuous derivatives into solutions along material interfaces, but additional unknowns must be solved in this method. Chen et al. [28] introduced the jump enrichment functions into the RK shape function based upon enforcing the consistency conditions, which is termed the interface-enriched reproducing kernel approximation (I-RKPM). However, coupling interface-enriched RK shape functions with the standard RK shape function requires duplicated unknowns. In addition, Masuda and Noguchi introduced a discontinuous derivative basis functions to replace the conventional polynomial basis function used in the MLS approximation [29]. Another class of methods introduces modifications to the weak formulation to consider the effects of discontinuity in a weak sense. Codes and Moran treated material interface discontinuities by a Lagrange Multiplier technique so that the approximations are disjoint across the interfaces while the Lagrange Multiplier imposes the interface continuity constraints into the variational formulation of the meshfree discretization [30]. This approach introduces additional degrees of freedom to be solved associated with the Lagrange Multiplier, and stability conditions need additional attention. On the other hand, the discontinuous Galerkin (DG) formulation has also been considered, where the continuity of a field variable and its resulting interface flux or traction across interfaces are imposed in the weak form [31], [32]. Wang et al. proposed a DG reformulation of the EFG and RKPM to address interface discontinuity problems of composite materials [33]. This approach avoids duplicated unknowns, and by decomposing the domain into patches, the gradient jump of the dependent variable is captured by the boundary of the adjacent patches while the continuity condition is realized weakly through an augmented variational form with associated flux or traction crossing material interfaces. Additionally, other meshfree methods have also been proposed for non-body-fitted discretization of heterogeneous media, such as the immersed methods [34], [35]. However, these methods require special care of interface oscillations due to the employment of volumetric constraints on the foreground and background discretization.

The current work introduces a novel Interface-Modified Reproducing Kernel Particle Method (IM-RKPM) to properly handle weak discontinuities in composite materials across material interfaces. The proposed approach utilizes signed distance functions obtained from SVM classified Micro-CT images to introduce regularized weak discontinuities to the kernel functions for arbitrary interface geometries. No duplicated unknowns, special enrichment functions, or complicated reformulation of the RK shape functions are required in the proposed approach, offering automated model construction capabilities for modeling complex microstructures.

The remainder of the paper is organized as follows. Section 2 provides basic equations for the model problem and the Reproducing Kernel Particle Method, and the associated numerical domain integration techniques are also discussed in this section. A brief introduction of the SVM formulation and the proposed SVM and RK-guided procedures for image segmentation of heterogenous materials are introduced in Section 3. Section 4 presents an interface-modified kernel function and the Interface-Modified Reproducing Kernel Particle Method formulation to introduce weak discontinuities in the image-based modeling of composite microstructures. In addition, two validation examples are presented to examine the proposed method's effectiveness. In section 5, two numerical examples for image-based modeling of microstructures are demonstrated, and the paper concludes with a discussion and summary in Section 6.

## 2 Basic Equations

### 2.1 Model problem

Let a model elasticity problem be defined on a domain $\Omega$ with its boundary assigned as $\partial\Omega = \partial\Omega_g \cup \partial\Omega_h$, $\partial\Omega_g \cap \partial\Omega_h = \emptyset$, where the subscripts $g$ and $h$ denote the Dirichlet and Neumann boundaries, respectively. The strong form for heterogeneous elastic media can be described as:

$$\begin{aligned} \nabla \cdot \boldsymbol{\sigma} + \boldsymbol{s} &= \boldsymbol{0} \text{ in } \Omega \\ \boldsymbol{u} &= \bar{\boldsymbol{u}} \text{ on } \partial\Omega_g \\ \boldsymbol{n} \cdot \boldsymbol{\sigma} &= \boldsymbol{h} \text{ on } \partial\Omega_h \end{aligned} \tag{1}$$

where $\boldsymbol{u}$ represents the unknown displacement field, $\boldsymbol{\sigma}$ is the Cauchy stress tensor, $\boldsymbol{s}$ is the body force vector, $\bar{\boldsymbol{u}}$ and $\boldsymbol{t}$ are the prescribed displacement vector and the applied surface traction vector on the Dirichlet and Neumann boundaries, respectively, and $\boldsymbol{n}$ is the unit outward normal of the Neumann boundaries. The elastic constitutive relationship for heterogeneous materials is represented as:

$$\boldsymbol{\sigma}(\boldsymbol{x}) = \boldsymbol{C}(\boldsymbol{x}) : \boldsymbol{\varepsilon}\big(\boldsymbol{u}(\boldsymbol{x})\big) \tag{2}$$

Here $\boldsymbol{C}(\boldsymbol{x})$ is the elasticity tensor defined as:

$$\boldsymbol{C}(\boldsymbol{x}) = \begin{cases} \boldsymbol{C}^{\mathbf{1}}, & \boldsymbol{x} \in \Omega^1 \\ \boldsymbol{C}^{\mathbf{2}}, & \boldsymbol{x} \in \Omega^2 \end{cases} \tag{3}$$

where $\Omega^i$ are material sub-domains to be segmented by the SVM classification of microstructure image pixels.

The weak formulation is to find $\boldsymbol{u}(\boldsymbol{x}) \in U \subset H_g^1$, such that for all weight function $\boldsymbol{v}(\boldsymbol{x}) \in V \subset H_0^1$,

$$\int_\Omega \boldsymbol{\varepsilon}(\boldsymbol{v}) : \boldsymbol{\sigma}(\boldsymbol{u}) d\Omega = \int_\Omega \boldsymbol{v} \cdot \boldsymbol{s} d\Omega + \int_{\partial\Omega_h} \boldsymbol{v} \cdot \boldsymbol{h} d\Gamma \tag{4}$$

The Galerkin formulation seeks the trial solution function $\boldsymbol{u}^h \in U^h \subset U$, so that for all weight function $\boldsymbol{v}^h \in V^h \subset V$,

$$\int_\Omega \boldsymbol{\varepsilon}\big(\boldsymbol{v}^h\big) : \boldsymbol{\sigma}\big(\boldsymbol{u}^h\big) d\Omega = \int_\Omega \boldsymbol{v}^h \cdot \boldsymbol{s} d\Omega + \int_{\partial\Omega_h} \boldsymbol{v}^h \cdot \boldsymbol{h} d\Gamma \tag{5}$$

### 2.2 Reproducing Kernel Approximation

Let a closed domain $\bar{\Omega} = \Omega \cup \partial\Omega \subset \mathbb{R}^d$ be discretized by a set of $NP$ nodes denoted by $\mathbb{S}^{\mathrm{RK}} = \{\boldsymbol{x}_1, \boldsymbol{x}_2, \dots, \boldsymbol{x}_{NP} \mid \boldsymbol{x}_I \in \bar{\Omega}\}$, and let the approximation of a field variable $\boldsymbol{u}(\boldsymbol{x})$ in $\bar{\Omega}$ be denoted by $\boldsymbol{u}^h(\boldsymbol{x})$. The RK approximation of the field variable $\boldsymbol{u}(\boldsymbol{x})$ based on the discrete points in the set $\mathbb{S}^{\mathrm{RK}}$ is formulated as follows:

$$u_i^h(\boldsymbol{x}) = \sum_{I=1}^{NP} \Psi_I(\boldsymbol{x}) d_{iI} \tag{6}$$

where $\Psi_I$ denotes the RK shape function with support centered at the node $\boldsymbol{x}_I$ and $d_{iI}$ is the nodal coefficient in $i^{th}$ dimension to be sought. Moreover, let a node $I$ be associated with a subdomain $\Omega_I$, over which a kernel function $\phi_a(\boldsymbol{x}-\boldsymbol{x}_I)$ with a compact support $a$ is defined, such that $\overline{\Omega} \subset \bigcup_{I\in\mathbb{S}^{\mathrm{RK}}} \Omega_I$ holds. The RK approximation function is constructed as:

$$\begin{aligned}\Psi_I(\boldsymbol{x}) = C(\boldsymbol{x};\boldsymbol{x}-\boldsymbol{x}_I)\phi_a(\boldsymbol{x}-\boldsymbol{x}_I) = \left(\sum_{|\alpha|\le n} (\boldsymbol{x}-\boldsymbol{x}_I)^{\alpha}\, b_{\alpha}(\boldsymbol{x})\right)\phi_a(\boldsymbol{x}-\boldsymbol{x}_I) \\ \equiv \boldsymbol{H}^{\mathrm{T}}(\boldsymbol{x}-\boldsymbol{x}_{\mathbf{I}})\boldsymbol{b}(\boldsymbol{x})\phi_{\mathrm{a}}(\boldsymbol{x}-\boldsymbol{x}_{\mathbf{I}})\end{aligned} \tag{7}$$

$$\boldsymbol{H}^T(\boldsymbol{x}-\boldsymbol{x}_I) = [1, x_1 - x_{1I}, x_2 - x_{2I}, \dots, (x_3 - x_{3I})^n] \tag{8}$$

where α is a multi-index notation such that $\alpha = (\alpha_1, \alpha_2, \dots, \alpha_d)$ with a length defined as $|\alpha| = \alpha_1 + \alpha_2 + \cdots + \alpha_d$ , and $\boldsymbol{x}^{\alpha} \equiv \boldsymbol{x}_1^{\alpha_1} \cdot \boldsymbol{x}_2^{\alpha_2}, \dots, \boldsymbol{x}_d^{\alpha_d}$ , $b_{\alpha} = b_{\alpha_1\alpha_2\cdots\alpha_d}$ . The term $C(\boldsymbol{x};\boldsymbol{x}-\boldsymbol{x}_I) = \boldsymbol{H}^{\mathrm{T}}(\boldsymbol{x}-\boldsymbol{x}_{\mathbf{I}})\boldsymbol{b}(\boldsymbol{x})$ is called the correction function of the kernel $\phi_a(\boldsymbol{x}-\boldsymbol{x}_I)$ designed to introduced completeness to the RK approximation. The terms $\{(\boldsymbol{x}-\boldsymbol{x}_I)^{\alpha}\}_{|\alpha|\le n}$ form a set of basis functions, and $\boldsymbol{H}^T(\boldsymbol{x}-\boldsymbol{x}_I)$ is the corresponding vector of basis functions to the order $n$. The vector $\boldsymbol{b}(\boldsymbol{x})$ is the coefficient vector of $\{b_{\alpha}(\boldsymbol{x})\}_{|\alpha|\le n}$ and is solved by enforcing the following discrete reproducing conditions [36]:

$$\sum_{I=1}^{NP} \Psi_I(\boldsymbol{x})\boldsymbol{x}_I^{\alpha} = \boldsymbol{x}^{\alpha}, \quad |\alpha| \le n \tag{9}$$

or equivalently,

$$\sum_{I=1}^{NP} \Psi_I(\boldsymbol{x})(\boldsymbol{x}-\boldsymbol{x}_I)^{\alpha} = \delta_{0\alpha}, \quad |\alpha| \le n \tag{10}$$

After inserting Eq. (10) into Eq. (7), $\boldsymbol{b}(\boldsymbol{x})$ is obtained as:

$$\boldsymbol{b}(\boldsymbol{x}) = \boldsymbol{M}^{-1}(\boldsymbol{x})\boldsymbol{H}(\boldsymbol{0})\phi_{\mathrm{a}}(\boldsymbol{x}-\boldsymbol{x}_{\mathbf{I}}) \tag{11}$$

where $\boldsymbol{M}(\boldsymbol{x})$ is the moment matrix and is formulated as:

$$\boldsymbol{M}(\boldsymbol{x}) = \sum_{I=1}^{NP} \boldsymbol{H}(\boldsymbol{x}-\boldsymbol{x}_{\mathrm{I}})\boldsymbol{H}^{\mathrm{T}}(\boldsymbol{x}-\boldsymbol{x}_{\mathrm{I}})\phi_a(\boldsymbol{x}-\boldsymbol{x}_{\mathrm{I}}) \tag{12}$$

Finally, the RK shape function is obtained as:

$$\Psi_I(\boldsymbol{x}) = \boldsymbol{H}^{\mathrm{T}}(\boldsymbol{0})\boldsymbol{M}^{-1}(\boldsymbol{x})\boldsymbol{H}(\boldsymbol{x}-\boldsymbol{x}_{\mathrm{I}})\phi_{\mathrm{a}}(\boldsymbol{x}-\boldsymbol{x}_{\mathrm{I}}) \tag{13}$$

Examples of 1-dimensional kernel functions are shown in Figure 2, and the 1- and 2-dimensional RK shape functions constructed based on cubic B-spline kernel function and linear basis functions are shown in Figure 3. The locality and the smoothness of the RK approximation functions are determined by the kernel function, while the order of completeness in the approximation is determined by the order of basis functions $n$. Interested readers are referred to [22]–[24], [37]–[39] for basic properties of reproducing kernel approximation.

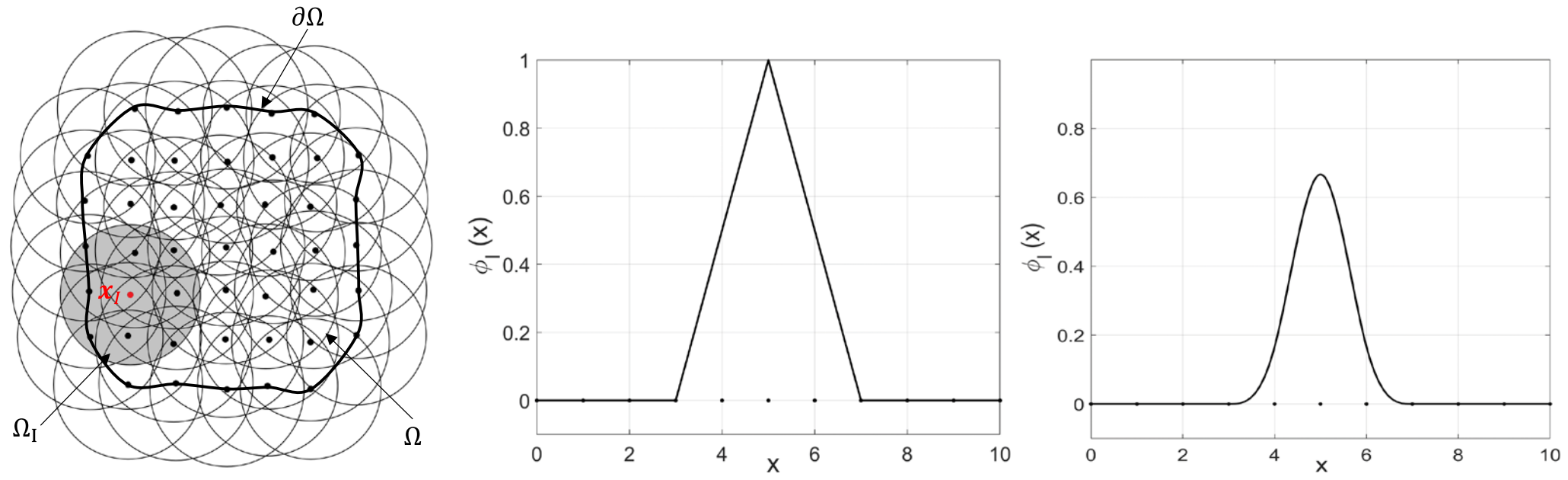

(a) RK discretization and nodal influence domains (b) Tent kernel function (c) Cubic B-spline kernel function

**Figure 2: RK domain discretization and examples of kernel functions**

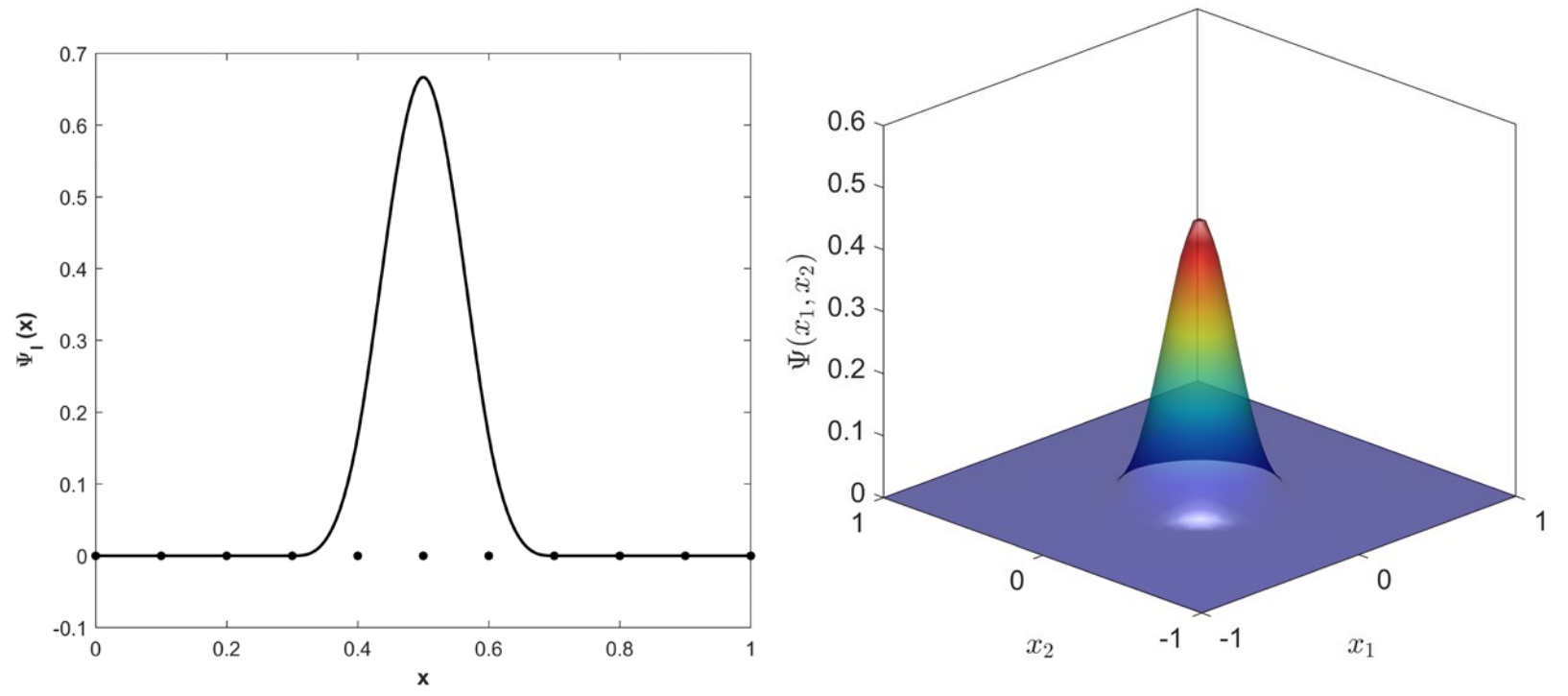

(a) 1D RK shape function centered at node $x_I = 0.5$ (b) 2D RK shape function centered at origin

**Figure 3: Examples of 1D and 2D RK shape functions constructed based on the cubic B-spline kernel function and linear basis functions**

## 2.3 Numerical domain integration

Due to the rational nature and arbitrary local supports, introducing RK approximations in the Galerkin weak form requires special attention. The conventional Gauss integration on background integration cells leads to a sub-optimal convergence unless significantly high-order quadrature rules are used, which is computationally infeasible especially in three-dimension [37]. Several domain integration techniques have been proposed, such as stabilized conforming and non-conforming nodal integrations [40]–[45] and variationally consistent integration [46], along with various stabilization methods [47]–[49]. The Gauss Integration (GI) and stabilized conforming nodal integration methods (SCNI) are summarized in this section.

### 2.3.1 Gauss integration

Gauss integration is introduced by subdividing the domain into background integration cells independent of the RK discretized nodes. For pixel discretization points, uniform integration cells illustrated in Figure 4 can be used for simplicity.

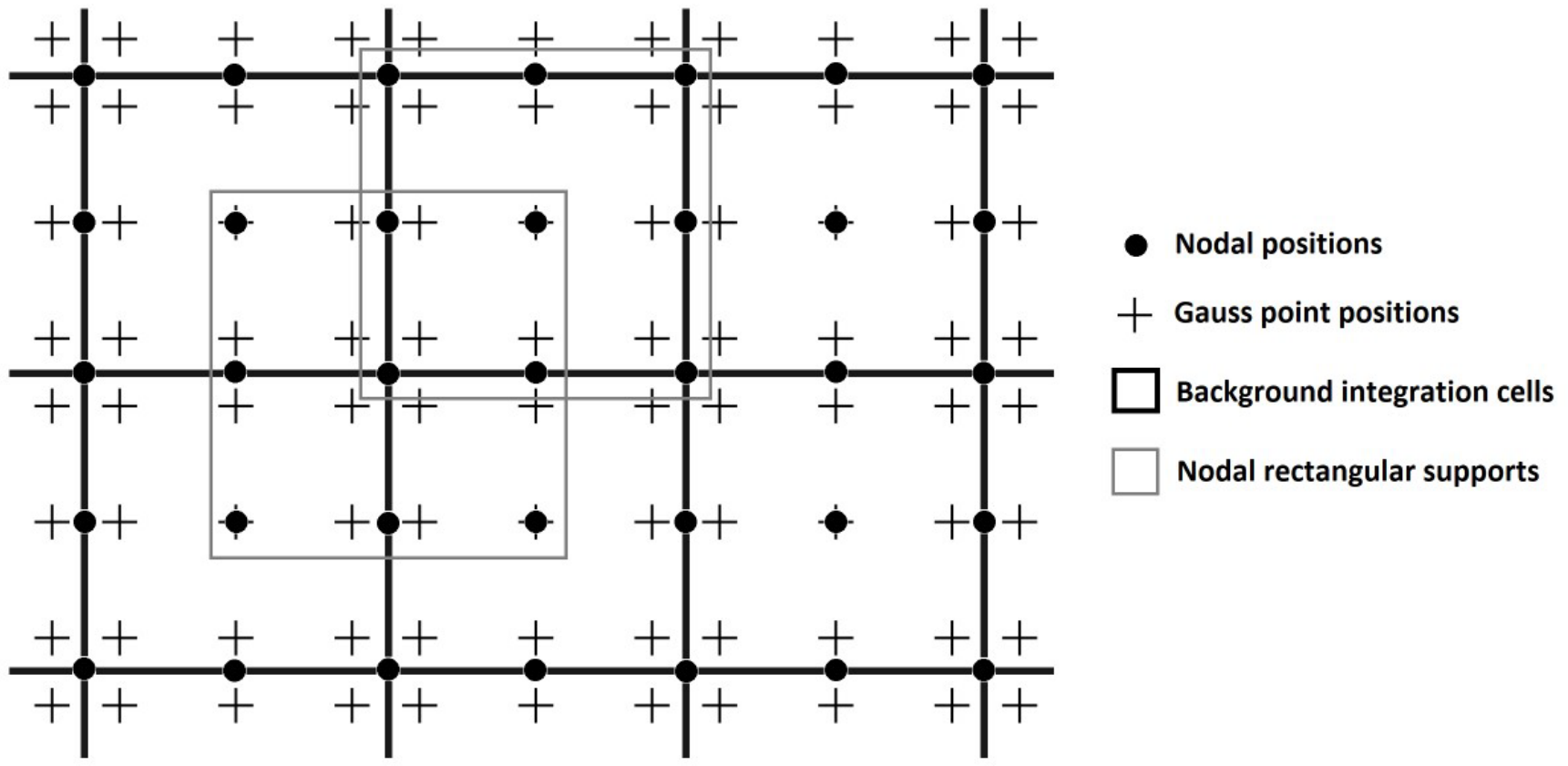

**Figure 4: Schematic of background Gauss integration cells and Gauss points distributions**

While Gauss integration is exceptionally efficient in integrating polynomials, meshfree shape functions are typically rational functions with overlapping supports. Additionally, the local domain of numerical integration in general misaligns with the nodal compact supports, which introduces significant error in Gauss integration, and high order quadrature rule is required to ensure optimal convergence [46], [50]. It is recommended in the literature [46] to use up to 5-point Quadrature rule to obtain desirable accuracy, which is computationally demanding, especially for high-dimensional problems. More detailed Gauss domain integration procedures can be referred to early Galerkin meshfree literature [22], [24], [51].

### 2.3.2 Stabilized Conforming Nodal Integration

One solution that significantly eases the computational cost of Gauss domain integration is to use the discretized nodes as integration points [47], referred to as the direct nodal integration (DNI). The DNI technique is appealing because of its simplicity and efficiency, as it does not require a background integration mesh, which makes numerical approximation truly "mesh-free." Nevertheless, since DNI is similar to a one-point quadrature rule, the under-integration of the weak form results in improper zero energy modes in most situations [40], [47].

Chen et al. [40] introduced the Stabilized Conforming Nodal Integration (SCNI) method as an enhancement of the DNI by fulfilling the consistency conditions between the approximations and the numerical integration of the weak form known as the integration constraints (IC). The SCNI method is formulated to exactly meet the first-order integration constraint and also to remedy the rank deficiency in the DNI method by introducing the following smoothed gradient in the Galerkin approximation:

$$\tilde{\nabla}\Psi_I(\boldsymbol{x}_L) = \frac{1}{W_L}\int_{\Omega_L} \Psi_I(\boldsymbol{x})\mathrm{d}\,\Omega = \frac{1}{W_L}\int_{\partial\Omega_L} \Psi_I(\boldsymbol{x})\boldsymbol{n}\,\mathrm{d}\Gamma \tag{14}$$

where $W_L$ denotes the areas of the nodal representative conforming smoothing cells, and $\boldsymbol{n}$ represents the unit outward normal of the smoothing cell boundaries. A convenient way of generating conforming smoothing cells is to create the Voronoi diagram according to the domain boundaries and nodal coordinates, as illustrated below in Figure 5, where the boundary integral in the smoothed gradient in Eq. (14) is carried out by the cell boundary quadrature points.

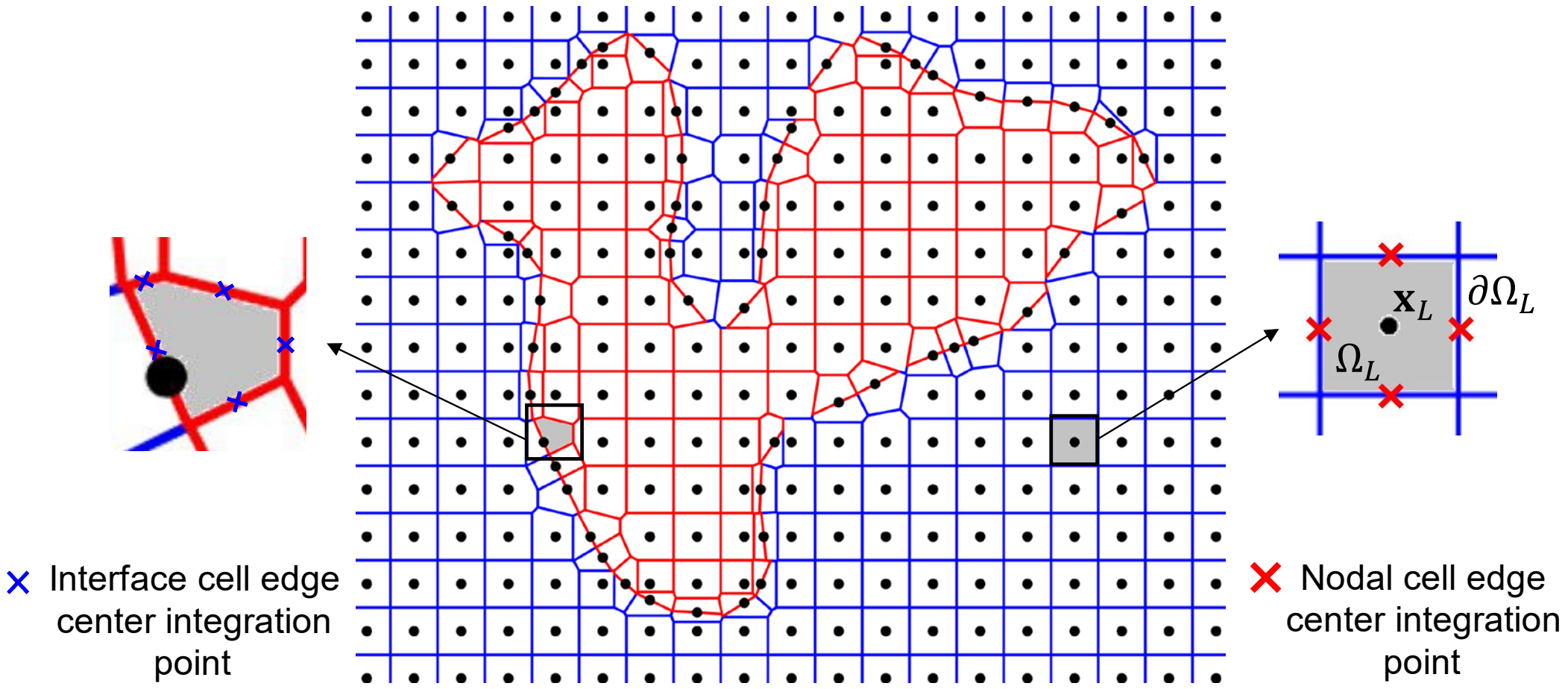

**Figure 5: Voronoi tessellation of domain and representative nodal cell**

The associated gradient matrix $\widetilde{\boldsymbol{B}}(\boldsymbol{x_L})$ of RK approximation evaluated at nodal integration point $\boldsymbol{x_L}$ is now expressed in terms of smoothed gradient as:

$$\widetilde{\boldsymbol{B}}_I(\boldsymbol{x}_L) = \begin{bmatrix} \tilde{b}_{I1}(\boldsymbol{x}_L) & 0 \\ 0 & \tilde{b}_{I2}(\boldsymbol{x}_L) \\ \tilde{b}_{I2}(\boldsymbol{x}_L) & \tilde{b}_{I1}(\boldsymbol{x}_L) \end{bmatrix} \tag{15}$$

$$\tilde{b}_{Ii}(\boldsymbol{x}_L) = \frac{1}{W_L}\int_{\Gamma_L} \Psi_I(\boldsymbol{x}) n_i(\boldsymbol{x}) d\Gamma$$

The stiffness matrix is then integrated by nodal integration with the smoothed gradient as:

$$\boldsymbol{K} = \sum_{L=1}^{NP} \tilde{\boldsymbol{B}}^{\boldsymbol{T}}(\boldsymbol{x}_L)\tilde{\boldsymbol{B}}(\boldsymbol{x}_L)\, W_L \tag{16}$$

It is noted that Voronoi cells conformed to the material interfaces without confining to the existing pixel points can be constructed, as demonstrated in Figure 5. In such construction, the centers of those Voronoi cells can be viewed as the integration points that are not coincided with the image pixel points. Details of constructing those Voronoi cells near the material interface are given in Appendix A.

# 3 Support Vector Machine (SVM) classification of Micro-CT images and model discretization

## 3.1 Support Vector Machine (SVM) classification algorithm

The Support Vector Machine (SVM) is a class of supervised machine learning algorithms that assigns labels to objects through training [52]. Let a labeled classification dataset containing $l$ sets of data be denote as $\mathbf{D} = \{(\boldsymbol{x}_i, y_i)\}_{i=1}^{l}$, where $\boldsymbol{x}_i \in \mathbb{R}^d$ is the $i^{th}$-dimensional data points, and $y_i$ is the label corresponding to the $i^{th}$ data point. Since the primary focus of this work is bi-material classification, $y_i$ is assumed to be either $-1$ or $+1$, representing the negative (matrix) and positive (inclusion) classes, respectively. If the given dataset is perfectly linearly separable, the SVM classification process can be described as to find a separating hyperplane in the form of a linear discriminant function in $d$-dimension:

$$h(\boldsymbol{x}; \{\boldsymbol{w}, b\}) = \boldsymbol{w}^T\boldsymbol{x} + b \tag{17}$$

where $\boldsymbol{w}$ denotes a $d$-dimensional *weight vector* and $b$ denotes a scalar *bias*. Additionally, $h(\boldsymbol{x}; \{\boldsymbol{w}, b\})$ serves as a linear classifier for class prediction following the decision rule:

$$y = \begin{cases} +1 & \text{if } h(\boldsymbol{x}; \{\boldsymbol{w}, b\}) > 0 \\ -1 & \text{if } h(\boldsymbol{x}; \{\boldsymbol{w}, b\}) < 0 \end{cases} \tag{18}$$

Therefore, the weight vector $\boldsymbol{w}$ is orthogonal to the defined hyperplane, and for each $\boldsymbol{x}_i \in \mathbf{D}$, the relative distance in terms of $\boldsymbol{w}$ to the defined hyperplane can be expressed as:

$$\xi_i = \frac{y_i(\boldsymbol{w}^T\boldsymbol{x}_i + b)}{\|\boldsymbol{w}\|} \tag{19}$$

The *margin* of the linear classifier is identified by selecting a collection of the data points that achieve a minimum distance to $h(\boldsymbol{x};\{\boldsymbol{w},b\})$, which are called the *support vectors* $\{\boldsymbol{x}^{SV}\}$, and are defined as:

$$\{\boldsymbol{x}^{SV}\} = \underset{\mathbf{x}_i \in \boldsymbol{D}}{\operatorname{argmin}} \left\{ \frac{y_i(\boldsymbol{w}^T \boldsymbol{x}_i + b)}{\|\boldsymbol{w}\|} \right\} \tag{20}$$

Note that if the distances of all support vectors to the hyperplane are normalized to be 1, the margin can be defined as $\xi^* = \frac{1}{\|\boldsymbol{w}\|}$. Thus, the goal of training the SVM with a linear classifier as Eq. (17) can be described as to find the optimal hyperplane $h^*(\boldsymbol{x})$ as follows:

$$\begin{gathered} h^*(\boldsymbol{x}) = h(\boldsymbol{x};\{\boldsymbol{w}^*,b^*\}), \\ \boldsymbol{w}^*, b^* = \underset{\boldsymbol{w},b}{\operatorname{argmax}} \left\{ \frac{1}{\|\boldsymbol{w}\|} \right\}, \\ \text{subject to: } y_i(\boldsymbol{w}^T \boldsymbol{x}_i + b) \geq 1, \qquad \forall \boldsymbol{x}_i \in \mathbf{D} \end{gathered} \tag{21}$$

The constrained optimization problem described in Eq. (21) can be formulated as an equivalent convex constrained minimization problem:

$$\begin{gathered} \min_{\boldsymbol{w},b} J(\boldsymbol{w}) = \frac{1}{2}\|\boldsymbol{w}\|^2, \\ \text{subject to: } y_i(\boldsymbol{w}^T \boldsymbol{x}_i + b) \geq 1, \qquad \forall \boldsymbol{x}_i \in \mathbf{D} \end{gathered} \tag{22}$$

which is called the *primal* formulation of SVM with linear classifier [13]. Instead of directly solving the primal convex minimization problem, it is computationally more efficient to solve the dual problem, formulated using the *Lagrange multipliers*. To construct the dual problem, a Lagrange multiplier $\alpha_i$ is introduced for each linear constraint based on the Karush-Kuhn-Tucker (KKT) conditions [53]:

$$\alpha_i(y_i(\boldsymbol{w}^T \boldsymbol{x}_i + b) - 1) = 0, \qquad \alpha_i \geq 0 \tag{23}$$

Then the objective of the dual problem can be formulated as:

$$\min_{\boldsymbol{w},b} L = \frac{1}{2}\|\boldsymbol{w}\|^2 - \sum_{i=1}^{l} \alpha_i(y_i(\boldsymbol{w}^T \boldsymbol{x}_i + b) - 1) \tag{24}$$

By finding the stationary point of the Lagrangian $L$ with respect to $\boldsymbol{w}$, the optimal weight vector $\boldsymbol{w}$ can be expressed in terms of a linear combination of the data points, data label, and Lagrange multipliers:

$$\boldsymbol{w} = \sum_{i=1}^{l} \alpha_i y_i \boldsymbol{x}_i \tag{25}$$

Additionally, a new constrain arises when minimizing $L$ with respect to the bias, which indicates that the sum of the labeled Lagrange multipliers must be equal to zero. Therefore, by substituting Eq. (25) into Eq. (24), the dual problem's training objective can be formulated as:

$$\begin{gathered}\max_{\boldsymbol{\alpha}} L_{dual} = \sum_{i=1}^{l} \alpha_i - \frac{1}{2}\sum_{i=1}^{l}\sum_{j=1}^{l} \alpha_i \alpha_j y_i y_j \boldsymbol{x}_i^T \boldsymbol{x}_j, \\ \text{subject to: } \alpha_i \geq 0, \sum_{i=1}^{l} \alpha_i y_i = 0\,, \text{for } i = 1,2,3,\dots,l\end{gathered} \tag{26}$$

where Eq. (26) forms a well-known convex quadratic programming problem (QPP).

Nevertheless, it is possible that no such hyperplane can be found through Eq. (26) as the real-world data sets are rarely perfectly separable. To deal with cases with overlapping classes, a non-negative slack variable $\nabla\epsilon_i$ is introduced to each data point $\boldsymbol{x}_i \in \mathbf{D}$, such that the linear constraint in Eq. (21) and Eq. (22) is modified as:

$$y_i(\boldsymbol{w}^T\boldsymbol{x}_i + b) \geq 1 - \nabla\epsilon_i, \qquad \forall \boldsymbol{x}_i \in \mathbf{D}, \qquad \nabla\epsilon_i \geq 0, \forall\, i = 1,2,3,\dots,l \tag{27}$$

It is worth noting that the magnitude of $\nabla\epsilon_i$ affects the correctness of the classification of its corresponding data point $\boldsymbol{x}_i$: when $\nabla\epsilon_i \geq 1$, the data point will be misclassified as it appears on the wrong side of the hyperplane. As a result, for non-separable data sets, SVM introduces a "soft margin" concept into the training process, and the new training objective function can be described as:

$$\begin{gathered}\min_{\boldsymbol{w},b,\xi_i} \left\{\frac{1}{2}\|\boldsymbol{w}\|^2 + C\sum_{i=1}^{l} \nabla\epsilon_i\right\}, \\ \text{subject to: } y_i(\boldsymbol{w}^T\boldsymbol{x}_i + b) \geq 1 - \nabla\epsilon_i, \qquad \forall \boldsymbol{x}_i \in \mathbf{D}, \\ \nabla\epsilon_i \geq 0, \forall\, i = 1,2,3,\dots,l\end{gathered} \tag{28}$$

where $C$ is a weight parameter that penalizes the cost of misclassification and $\sum_{i=1}^{l} \nabla\epsilon_i$ gives the loss due to the deviation from the separable cases with the introduction of slack variables. Moreover, the penalty weight parameter $C$ controls the trade-off between maximizing the hyperplane's margin and minimizing the misclassification's loss. Therefore, the selection of $C$ depends on the nature of problems and datasets at hand. Figure 6 presents the hard- and soft-margined linear SVM classifiers trained on a binary dataset. For the hard-margined SVM classifier, the support vectors are located exactly on the maximum lower and upper margins, while the soft-margined SVM classifier relaxes the linear separability constraints, which allows some support vectors to cross over the decision boundary. Note that although the dataset in Figure 6 is linearly separable if the linear separability is strictly enforced, as for the hard-margined case, the resulting margin is significantly smaller than the one for the soft-margined case.

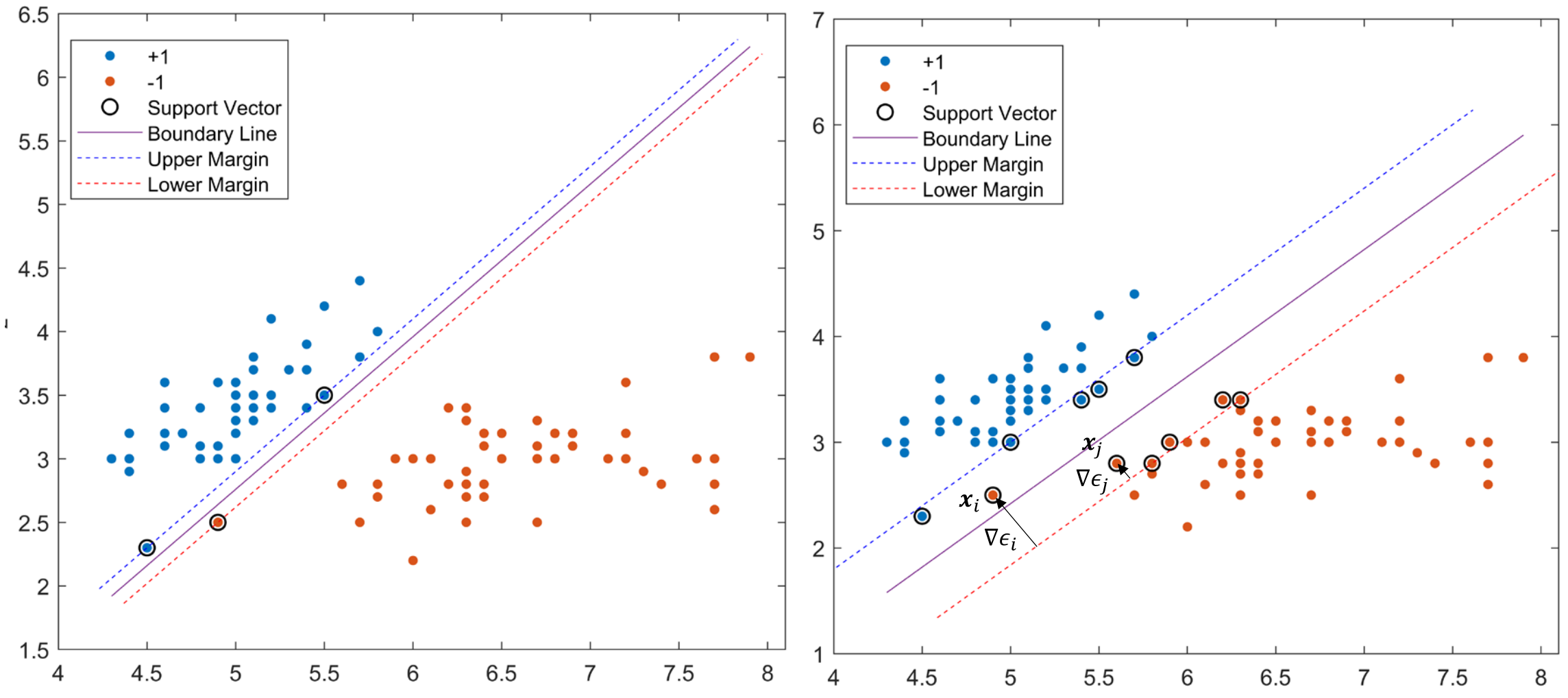

(a) Hard margin: $y_i(\boldsymbol{w}^T\boldsymbol{x}_i + b) \geq 1$ (b) Soft margin: $y_i(\boldsymbol{w}^T\boldsymbol{x}_i + b) \geq 1 - \nabla\epsilon_i$

**Figure 6: Hard- and soft-margined SVM linear classifiers**

By introducing Lagrange multipliers $\alpha_i$ and $\beta_i$, corresponding to each of the constraints in Eq. (28), a Lagrangian of Eq. (28) can be formulated as:

$$L = \frac{1}{2}\|\boldsymbol{w}\|^2 + C\sum_{i=1}^{l}\nabla\epsilon_i - \sum_{i=1}^{l}\alpha_i(y_i(\boldsymbol{w}^T\boldsymbol{x}_i + b) - 1 + \nabla\epsilon_i) - \sum_{i=1}^{l}\beta_i\nabla\epsilon_i \tag{29}$$

By minimizing the $L$ with respect to $\boldsymbol{w}$, $b$, and $\nabla\epsilon_i$, respectively, one additional relationship connecting the penalty weight parameter $C$ to $\alpha_i$ and $\beta_i$ is obtained, in addition to the ones acquired in the linearly separable cases:

$$C = \alpha_i + \beta_i \tag{30}$$

Therefore, the dual objective can be described as:

$$\max_{\boldsymbol{\alpha}} L_{dual} = \sum_{i=1}^{l}\alpha_i - \frac{1}{2}\sum_{i=1}^{l}\sum_{j=1}^{l}\alpha_i\alpha_j y_i y_j \boldsymbol{x}_i^T\boldsymbol{x}_j,$$
$$\text{subject to: } 0 \leq \alpha_i \leq C, \sum_{i=1}^{l}\alpha_i y_i = 0\text{, for } i = 1,2,3,\ldots,l \tag{31}$$

Note that the objective function achieved for the inseparable cases is the same as the one obtained from the linearly separable cases in Eq. (26), except one additional constraint on the Lagrange multiplier $\alpha_i$.

For complicated data sets, the linear classifier is often found inadequate. One strategy is to introduce non-linear transformation function $\phi$, which maps the data points $\boldsymbol{x}$ to a higher dimension so that the projected data points $\phi(\boldsymbol{x})$ are approximately linearly separable in the higher dimensional feature space. However, upscaling the dimensionality usually leads to high and impractical computational costs. Since the Lagrange dual formulation in Eq. (31) only depends on the dot product between two vectors in the feature space, the SVM can utilize the "kernel trick" to include high-degree polynomial features. The idea of the kernel trick is to represent $l$ data point $\boldsymbol{x}$ by a $l$ by $l$ kernel matrix $\boldsymbol{K}$ that contains elements $k_{i,j} = \boldsymbol{K}(\boldsymbol{x}_i, \boldsymbol{x}_j) = \langle \phi(\boldsymbol{x}_i), \phi(\boldsymbol{x}_j) \rangle$, which performs pairwise similarity comparisons between the original low dimensional data points without an explicit definition of the transformation function $\phi$ for mapping data to high dimensions. More detailed introduction to the requirement and existence of kernel function can be found in [14]. Therefore, the dual formulation of the training objective for non-linear SVM can be described by replacing $\boldsymbol{x}_i^T \boldsymbol{x}_j$ in Eq. (31) by a kernel function $\boldsymbol{K}(\boldsymbol{x}_i, \boldsymbol{x}_j)$:

$$\max_{\boldsymbol{\alpha}} L_{dual} = \sum_{i=1}^{l} \alpha_i - \frac{1}{2} \sum_{i=1}^{l} \sum_{j=1}^{l} \alpha_i \alpha_j y_i y_j \boldsymbol{K}(\boldsymbol{x}_i, \boldsymbol{x}_j),$$

$$\text{subject to: } 0 \leq \alpha_i \leq C, \sum_{i=1}^{l} \alpha_i y_i = 0, \text{for } i = 1,2,3, \ldots, l \tag{32}$$

Note that Eq. (32) can be viewed as a generalized Lagrange dual formulation of SVM since for linear cases, the kernel function can be expressed as the data point dot product as:

$$\text{Linear: } \boldsymbol{K}(\boldsymbol{x}_i, \boldsymbol{x}_j) = \boldsymbol{x}_i^T \boldsymbol{x}_j \tag{33}$$

Other widely used kernel functions are polynomial and Gaussian radial basis kernel functions, which are illustrated in Eq. (34) and Eq. (35), respectively.

$$\text{Polynomial: } \boldsymbol{K}(\boldsymbol{x}_i, \boldsymbol{x}_j) = \left(1 + \boldsymbol{x}_i^T \boldsymbol{x}_j\right)^q, q = 2,3, \ldots, \tag{34}$$

$$\text{Gaussian radial basis: } \boldsymbol{K}(\boldsymbol{x}_i, \boldsymbol{x}_j) = e^{-\gamma \|x_i - x_j\|^2}, \gamma > 0 \tag{35}$$

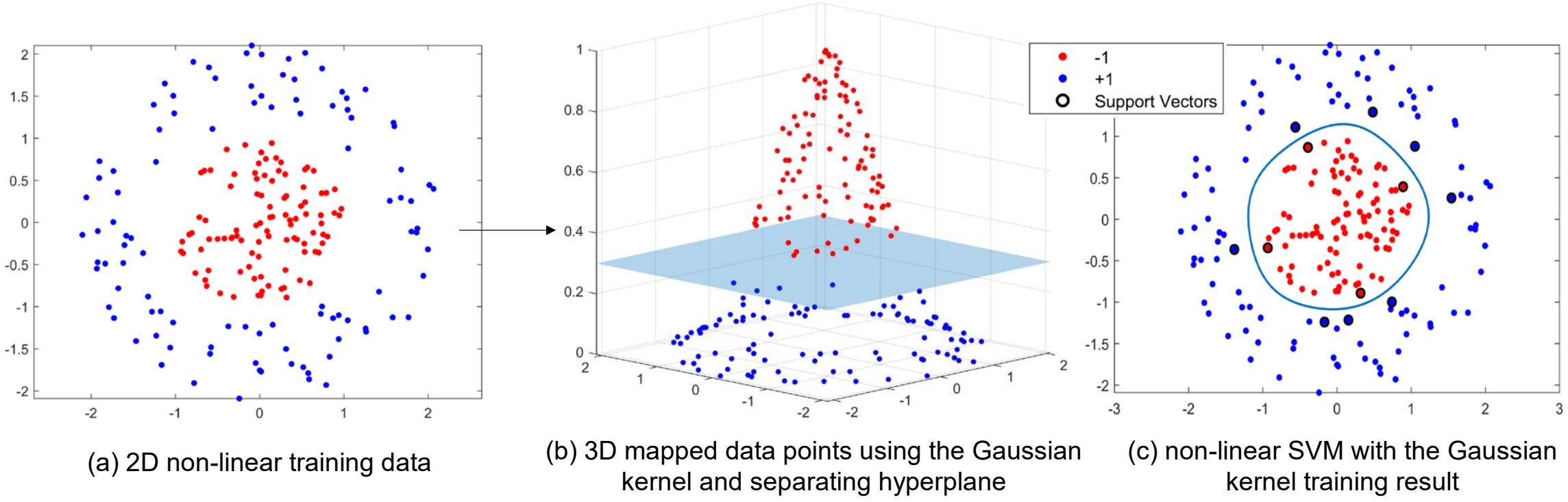

**Figure 7: An example of training the non-linear SVM with the Gaussian kernel** (The circled data points are the support vectors)

Figure 7 demonstrates the transformation of the 2-dimensional training data to 3-dimension using a Gaussian kernel. It is clear that the 3-dimensional data points become linearly separable by a 2-dimensional hyperplane, and the resulting separating hyperplane can be mapped back to the 2-dimensional space, which becomes a nonlinear curve.

Moreover, the SVM produces a classification score by predicting new datasets that provide information about the material class and reveal the location of material interfaces. The classification score is a signed distance measure for an observation point $\boldsymbol{x}$ to its nearest decision boundary, with a score of zero denoting $\boldsymbol{x}$ is precisely on the decision boundary. Therefore, the classification score acts as a guide in identifying material boundaries in the image, facilitating more accurate numerical model generation. The classification score for predictions at $\boldsymbol{x}$ to the positive class is defined as:

$$S(\boldsymbol{x}) = \sum_{j=1}^{n} \alpha_j y_j \boldsymbol{K}(\boldsymbol{x}_j, \boldsymbol{x}) + b \tag{36}$$

where $n$ is the total number of support vectors and $(\alpha_1, \alpha_2, \dots, \alpha_n, b)$ are the trained SVM parameters.

In summary, the SVM algorithms have shown promising performances in many application fields [54]. During the hyperplane selection process, SVMs utilize different kernel functions to transform the low-dimensioned, non-linear, and possibly non-separable training data to higher-dimensional feature spaces, which allows the data to be linearly separated. In addition, the selected high-dimensional hyperplane can be projected back down to the original space where the training data belong, providing non-linear decision boundaries between the separated classes [55]. As a result, SVMs not only aid in classifying different material pixels from Micro-CT images of heterogeneous materials but also inherently identify material interfaces. Here, we use SVMs to guide numerical model discretization for the proposed Interface-modified Reproducing Kernel Particle Method, which will be introduced in the later sections.

## 3.2 From Micro-CT images to numerical models

In this work, the sample images are taken from Micro-Computed Tomography (Micro-CT). Micro-CT is an imaging technique that generates three-dimensional images of an object's microstructure with (sub)micron resolution using an X-ray tube with cone-beam geometry as a source and a rotating sample holder [56].

### 3.2.1 Training data preparation and training the SVM

The training data points are located at the centroid of each pixel cell in the sample image, and the physical coordinates of those data points are assigned as the training data for the SVM. To supervise the SVM's training, response labels $y$, are created by segmenting the sample image using Otsu's method [57]. Otsu's method selects a global threshold that maximizes inter-class intensity variance from the zeroth- and the first-order cumulative moments of the sample image's intensity-level histogram. Figure 8 illustrates the alumina-epoxy composite sample images, where the white areas in the sample image indicate the alumina inclusion material and the grey areas represent the epoxy material in the matrix. Thus, the supervised training for the SVM aims to establish the correlation between physical locations in the image and their respective material types.

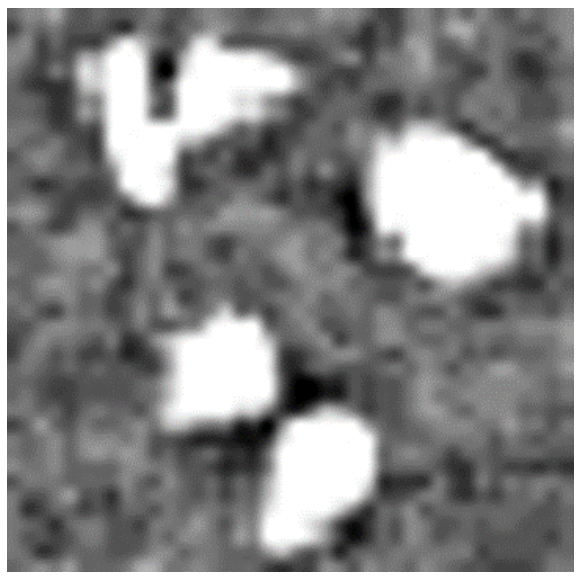

**Figure 8: Sample alumina-epoxy image**

Specific hyperparameters of the training must be determined beforehand to facilitate SVM classification, which are summarized in Table 1.

| SVM training hyperparameters | Values |
|---|---|
| Kernel function | Gaussian radial basis function |
| Kernel scale ($\gamma$ in Eq. (35)) | 0.25 |
| Penalty weight parameter ( $C$ in Eq. (28)) | 500 |

**Table 1: SVM hyperparameters selected for training the sample image**

The kernel scale determines the extent to which each data point affects the shape of the decision boundaries, which is selected as 0.25. Furthermore, a penalty weight parameter $C$ of $500$ is chosen to ensure that the resulting separation hyperplane resembles the material interface. Additionally, a standardization in which the training data points are normalized to have a zero mean and a standard deviation of one is performed. The standardization of the training data is critical because SVM training is based on the relative distances between the training data points, and without standardization, larger-scale training data may dominate in distance determination, leading to a biased model.

### 3.2.2 RK interface nodes

In the Interface-modified RK approximation to be proposed in Section 4, a set of interface nodes are included to introduce proper weak discontinuity across material interfaces. In this section, we present an approach to generate interface nodes, utilizing the score $S(\boldsymbol{x})$ (Eq. (36)) that is produced during the SVM classification and can be interpreted as a scaled signed distance function.

Let $\mathbb{S}^0 \equiv \{\boldsymbol{x}_I\}_{I=1}^{NP_0}$ be the set of training data points located at the image pixel centroids in the image domain $\bar{\Omega}$, and define $\mathbb{S}^+ \equiv \{\boldsymbol{x} \in \mathbb{S}^0 \mid S(\boldsymbol{x}) \geq 0\}$ and $\mathbb{S}^- \equiv \{\boldsymbol{x} \in \mathbb{S}^0 \mid S(\boldsymbol{x}) < 0\}$. Consequently, defining a set of interface-searching node pairs $\{\boldsymbol{x}_K^+, \boldsymbol{x}_K^-\}_{K=1}^{NP_{IF}}$ in which $\boldsymbol{x}_K^\pm \in \mathbb{S}^\pm$ is a near-interface master/slave node (see Remark 3.1 for details). The search of an interface node $\boldsymbol{x}_K^* \equiv \boldsymbol{x}_K^+ + d_K^* \boldsymbol{R}_K$ can be defined as follows:

Find $d_K^* \in \mathbb{R}$ such that:

$$S(\boldsymbol{x}_K^+ + d_K^* \boldsymbol{R}_K) = 0, \;\; \forall \, K = 1 \cdots NP^* \tag{37}$$

where $\boldsymbol{R}_K = (\boldsymbol{x}_K^- - \boldsymbol{x}_K^+)/\|\boldsymbol{x}_K^- - \boldsymbol{x}_K^+\|$ is the line search direction. The resulting RK node set is then $\mathbb{S}^{RK} \equiv \mathbb{S}^0 \cup \mathbb{S}^{IF}$ with $\mathbb{S}^{IF} \equiv \{\boldsymbol{x}_K^*\}_{K=1}^{NP^*}$, which serves as the SVM-RK discretized model.

**Remark 3.1.** *The interface-searching node pairs* $\{\boldsymbol{x}_K^+, \boldsymbol{x}_K^-\}_{K=1}^{NP_{IF}}$ *can be determined in various ways. In this work, the following approach is taken: given the set of support vectors* $\{\boldsymbol{x}_L^{SV}\}_{L=1}^{N_{SV}}$,

*define the master candidate node set* $\mathbb{S}^{C+} \equiv \{\boldsymbol{x}_I \in \mathbb{S}^+ \mid \|\boldsymbol{x}_I - \boldsymbol{x}_L^{SV}\| \leq \xi\ell,\ \forall L = 1 \cdots N_{SV}\}$, *in which* $\ell$ *and* $\xi$ *denote the image voxel size and a scaling factor, respectively. In this work* $\xi = 1.5$ *is used. The corresponding nearest slave nodes* $\boldsymbol{x}_K^-$ *are found such that, for* $\boldsymbol{x}_K^+ \in \mathbb{S}^{C+}$,

$$\boldsymbol{x}_K^- = \underset{\boldsymbol{x}_I \in \mathbb{S}^{C-}}{\operatorname{argmin}} \|\boldsymbol{x}_I - \boldsymbol{x}_K^+\| \tag{38}$$

*with the slave candidate node set* $\mathbb{S}^{C-} \equiv \{\boldsymbol{x}_I \in \mathbb{S}^- \mid \|\boldsymbol{x}_I - \boldsymbol{x}_L^{SV}\| \leq \xi\ell,\ \forall L = 1 \cdots N_{SV}\}$. *Note that Eq. (38) can result in multiple* $\boldsymbol{x}_K^-$ *for one* $\boldsymbol{x}_K^+$ *and lead to the master-slave pairs. Figure 9 (a) illustrates an example of the master and slave candidate nodes plotted along with the support vectors. The corresponding candidate node pairs are shown in Figure 9 (b).*

**Remark 3.2.** *The solution of* $d_K^*$ *in Eq. (37) can be determined iteratively by the Newton-Raphson method. For the* $(\nu + 1)^{th}$ *iteration, the increment* $\Delta d^{\nu+1}$ *for* $d_K^{*\nu+1} = \Delta d_K^{*\nu+1} + d_K^{*\nu}$ *is obtained as follows:*

$$\Delta d_K^{*\nu+1} = -S(\boldsymbol{x}_K^{*\nu}) / \left( \frac{\partial S(\boldsymbol{x}_K^{*\nu})}{\partial \boldsymbol{x}_K^{*\nu}} \cdot \boldsymbol{R}_K \right) \tag{39}$$

**Remark 3.3.** *One may consider interpolating the score function with SVM predicted nodal score values without constructing Eq.* (36) *as follows:*

$$\tilde{S}(\boldsymbol{x}) = \sum_{I=1}^{NP_0} \Psi_I^0(\boldsymbol{x}) s_I \tag{40}$$

*where* $\Psi_I^0(\boldsymbol{x})$ *and* $s_I$ *are the RK shape function constructed on* $\mathbb{S}^0$ *and SVM score value (signed distance) for node* $\boldsymbol{x}_I$, *respectively. The RK shape function can serve as a filter for potentially noisy predicted score values. Eq. (40) is used for numerical implementation in current work.*

**Remark 3.4.** *To ensure relative even distribution of nodes around interfaces, a MATLAB in-built function "uniqetol" with a relative tolerance 0.01 is applied to the interface node set* $\mathbb{S}^{IF}$. *In addition,* $\boldsymbol{x}_I \in \mathbb{S}^0$ *is removed if* $\|\boldsymbol{x}_I - \boldsymbol{x}_K^*\| < \zeta\ell$ *for all* $\boldsymbol{x}_K^* \in \mathbb{S}^{IF}$, *and* $\zeta = 1/3$ *is selected in this work. An example of rearranged RK nodes is illustrated in Figure 10. The material interfaces are represented by a simple line connection in Figure 10 for visualization purposes; the interface-modified RK approximation to be discussed next requires only the interface point locations and the signed distance of each discrete point obtained from SVM. Note that discretization far from the material interfaces can be made coarser to improve computational efficiency.*

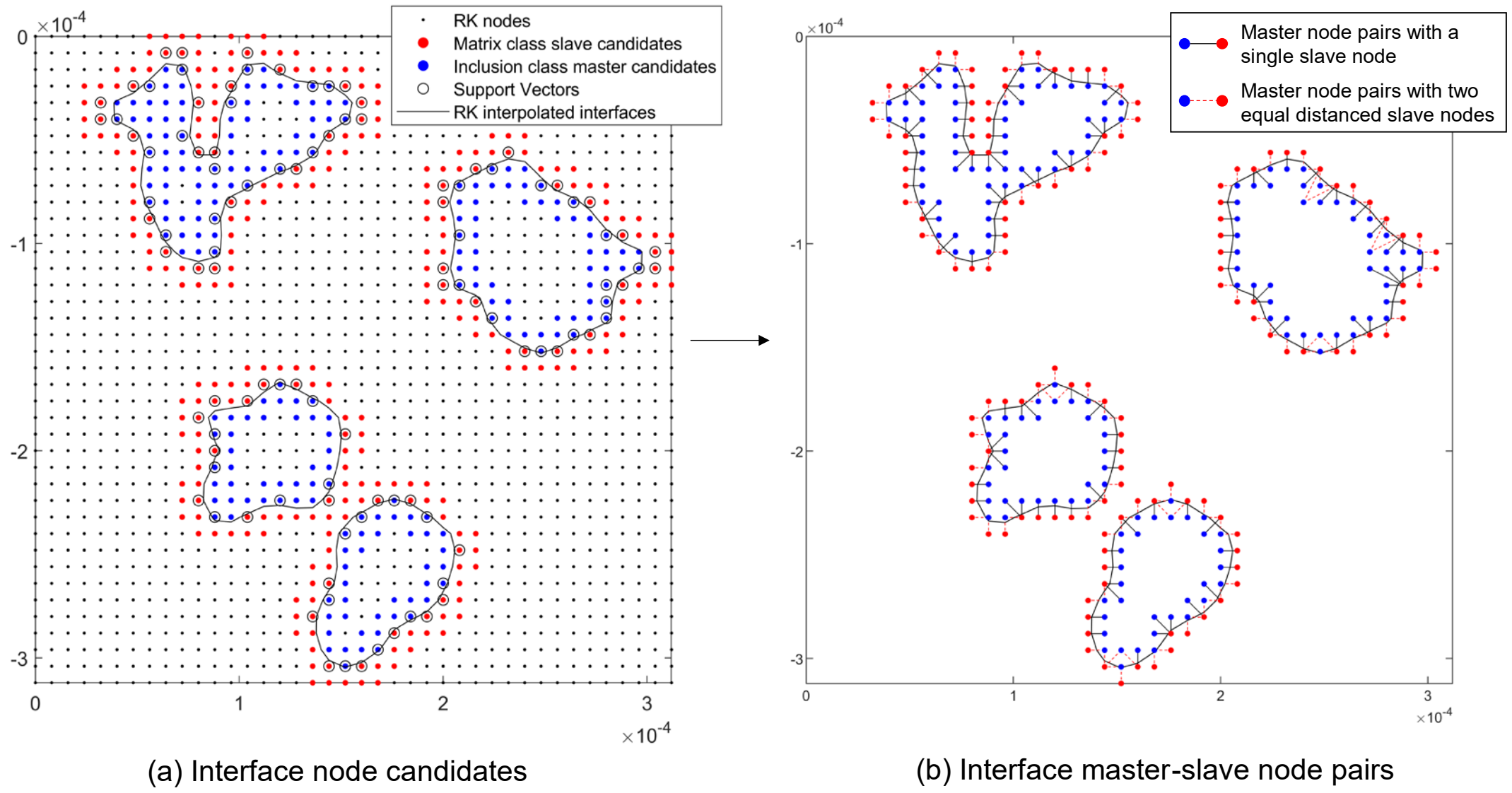

**Figure 9: Interface node candidates and resulting master-slave interface node search pairs**

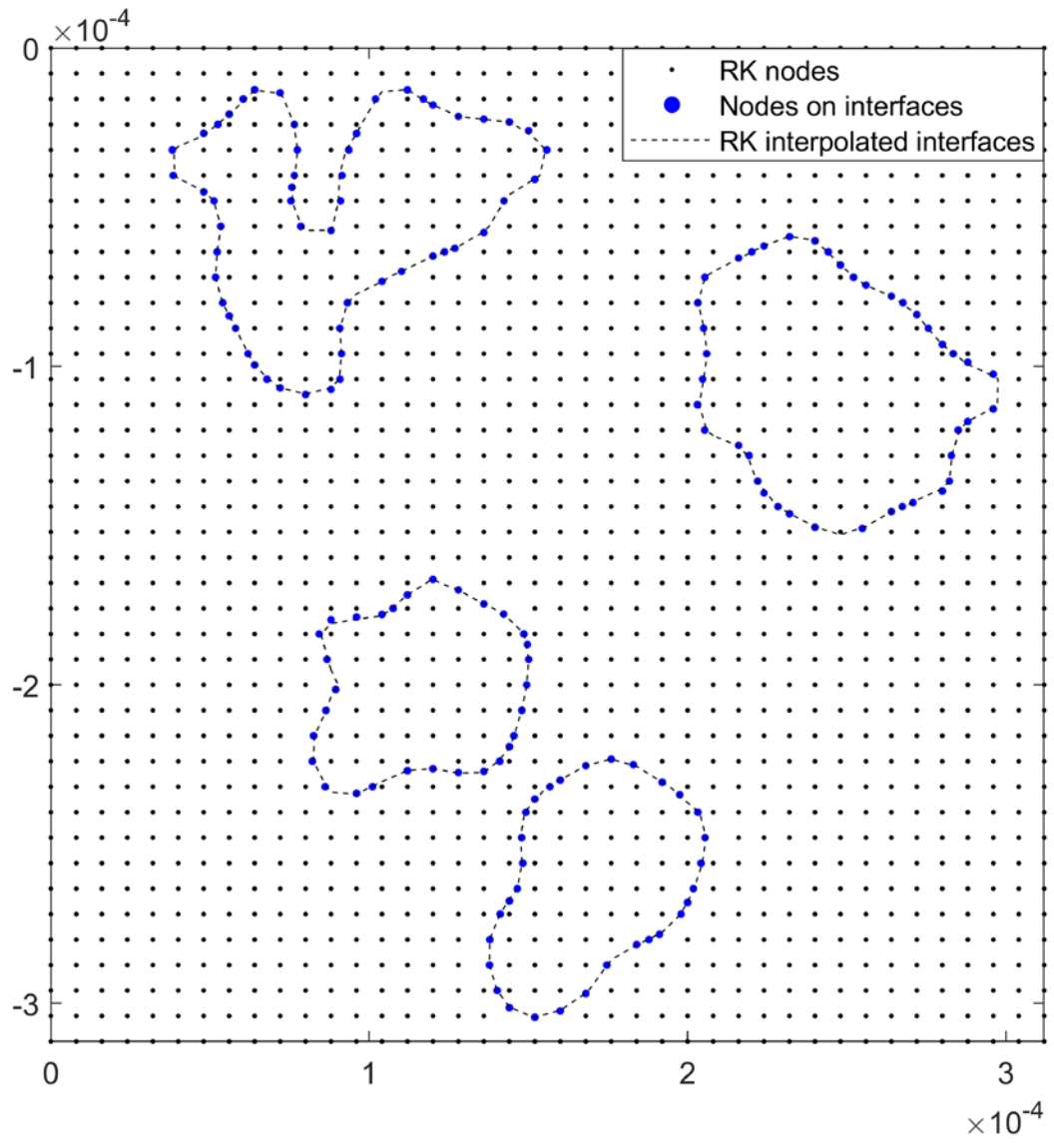

**Figure 10: RK numerical model for the test image**

## 3.3 Image-based SVM-RK model validation

### 3.3.1 Validation with a synthetic image

A synthetic two-phase image containing the known locations of inclusions is generated, as illustrated in Figure 11. The synthetic image has a dimension of 10 mm × 10 mm and a resolution of 224 × 224 pixels. To account for uncertainties in the imaging process, Gaussian noise is added to the original image, and the manufactured testing image is scaled down to 100 × 100 pixels to lower the resolution, especially around the material interfaces. Figure 12 demonstrates the manufactured noisy testing image, which will serve as the input image for the image-based SVM-RK model generation. The accuracy of the obtained interface nodes is determined by a normalized mean square error of the discretized material interfaces as:

$$MSE = \frac{1}{NC \cdot L}\left(\sum_{j=1}^{NC}\sum_{K=1}^{NP^*}\left(\|\boldsymbol{x}_K^* - \boldsymbol{c}_j\| - R_j\right)^2\right)^{\frac{1}{2}} \tag{41}$$

where $\boldsymbol{c}_j$ and $R_j$ represent the center coordinates and radius of the inclusion to which the interface node $\boldsymbol{x}_K^* \in \mathbb{S}^{IF}$ belongs, $NP^*$ is the total number of generated interface nodes in the set $\mathbb{S}^{IF}$, and $NC$ and $L$ denote the total number of inclusions in the synthetic image and the x-dimension of the synthetic image, respectively.

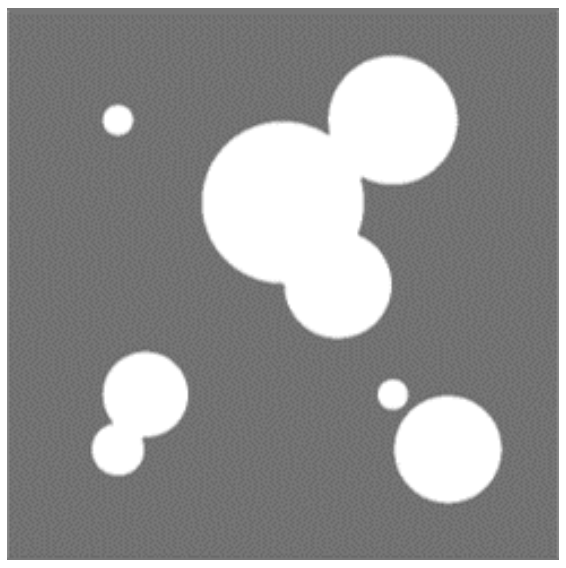

**Figure 11: Synthetic testing image for validating the SVM-RK interface node generation**

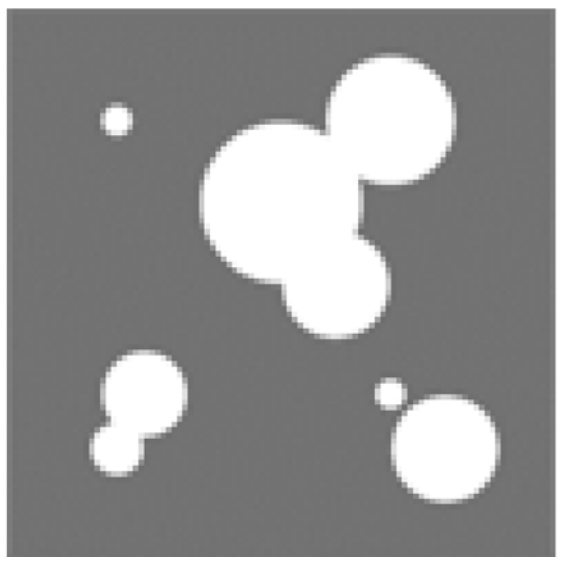

**Figure 12: Manufactured noisy testing image for validating the SVM-RK interface node generation**

As previously discussed, for implementation the score function is interpolated using the RK shape function in Eq. (40), and the locality and smoothness of the RK shape function may differ depending on the size and order of continuity of the kernel function chosen for its construction. Therefore, various RK kernel support sizes and kernel functions with different orders of continuity are employed to study their effects on the accuracy of the SVM-RK interface node generation algorithm. Figure 13 illustrates the obtained interface nodes overlaid with the synthetic image using a cubic B spline RK kernel (B3, $C^2$ continuity) with a normalized support size of 2 and linear bases. In addition, results of the interface node generated with various support sizes and kernel function continuities can be found in Table 2 and Table 3, respectively. Upon comparison of results in Table 2 and Table 3, it can be observed that the proposed SVM-RK interface node search algorithm converges less than an average of 5 iterations for all instances, and the resulting interface nodes achieve average scores (Eq. (40)) to the order of $10^{-12}$. Additionally, the normalized mean square error of the proposed image-based RK discretization model generation process is approximated 0.65% for all scenarios. Moreover, the results show that the generated interface nodes are not sensitive to the choices of the kernel support size and kernel continuity (tent function with $C^0$ continuity, quadratic B spline function (B2) with $C^1$continuity, and B3 with $C^2$ continuity) used in constructing the score function (Eq. (40)).

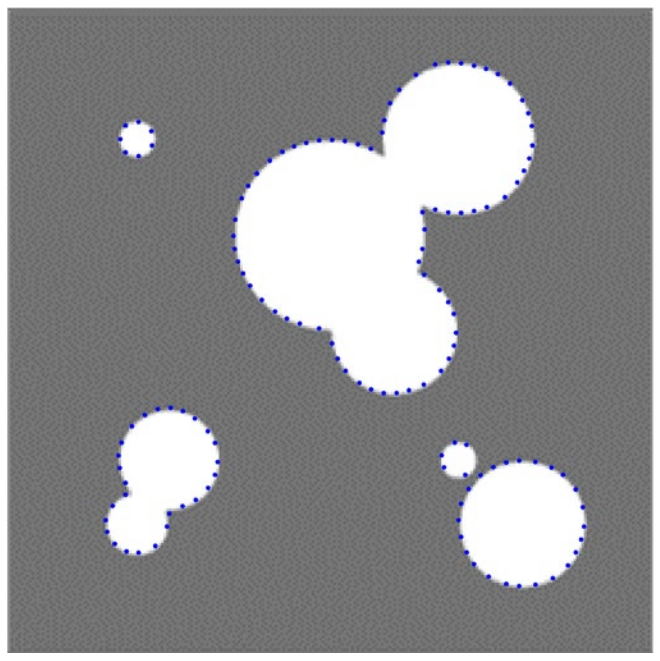

**Figure 13: Interface nodes overlapped with the manufactured test image**

| Support size | Number of constructed interface nodes | Mean iteration number | Mean score values | Edge detection mean squared error |
|---|---|---|---|---|
| 1.10 | 134 | 3.72 | 1.63E-12 | 0.0065 |
| 1.50 | 134 | 4.61 | 5.91E-12 | 0.0065 |
| 2.00 | 134 | 4.60 | 4.40E-12 | 0.0065 |
| 2.50 | 133 | 4.62 | 2.99E-12 | 0.0066 |
| 3.00 | 131 | 4.97 | 4.15E-12 | 0.0067 |

**Table 2: Results of interface node search algorithm with various kernel support sizes**

| Kernel Function | Number of constructed interface nodes | Mean iteration number | Mean score values | Edge detection mean squared error |
|---|---|---|---|---|
| $C^0$ (Tent) | 133 | 3.62 | 5.92E-13 | 0.0065 |
| $C^1$ (B2) | 132 | 4.64 | 4.78E-12 | 0.0066 |
| $C^2$ (B3) | 134 | 4.60 | 4.40E-12 | 0.0065 |

**Table 3: Results of interface node search algorithm with various RK kernel functions**

### 3.3.2 Validation of the image-based SVM-RK discrete model with Scanning Electron Microscopy (SEM) images

To analyze the quality of the proposed image-based RK discretization model generation procedure, a comparison is made between the constructed digital surface model from Micro-CT and a surface image obtained from Scanning Electron Microscopy (SEM) with a spatial resolution of 1.5 $\mu m$ for the same specimen as a comparison reference. SEM uses an electron beam to scan the surface of a material, producing a high-resolution image that reveals details such as surface topography, crystalline structure, chemical composition, and electrical behavior of the top 1 $\mu$m portion of a specimen [58]. The inclusion materials in SEM are identified based on the Energy Dispersive X-ray Spectroscopy (EDS), a chemical analysis technique that detects X-rays emitted by the material in response to the electron beam to form an elemental mapping of the SEM-scanned specimen surface [59]. The Micro-CT input image for constructing the numerical model is selected accordingly near the surface of the same specimen, and a Region of Interest (ROI) around 2.82 mm by 2.37 mm is chosen to match the SEM scanned area, which is highlighted in the red box in Figure 14.

An SVM-RK discretization model is created from the input 2D slice of the Micro-CT image using the proposed method, as illustrated in Figure 15, containing discretized nodes in the epoxy matrix, alumina inclusions, and on the identified material interfaces.

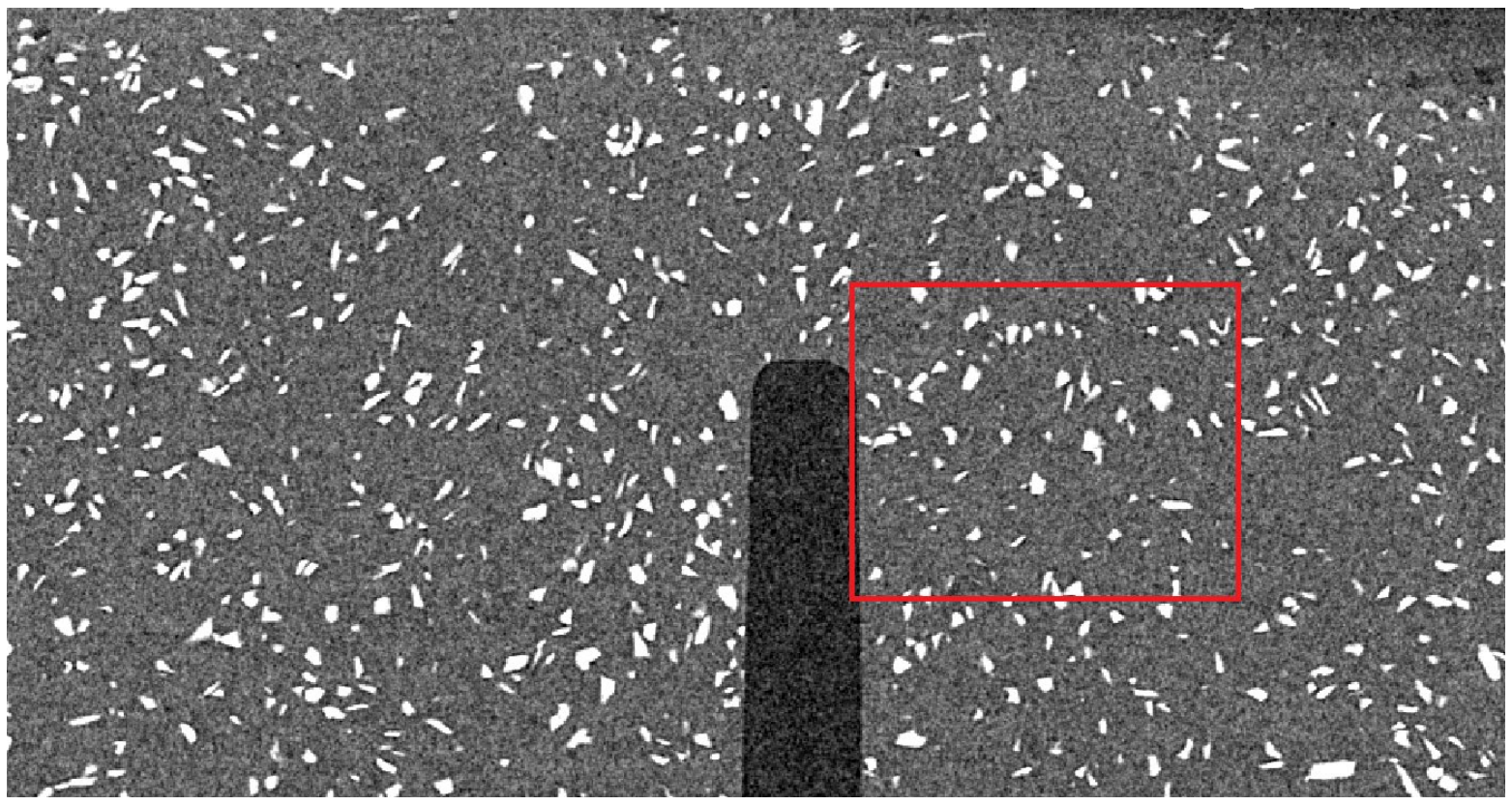

**Figure 14: Micro-CT input image selection for quantitative RK discretization model validation**

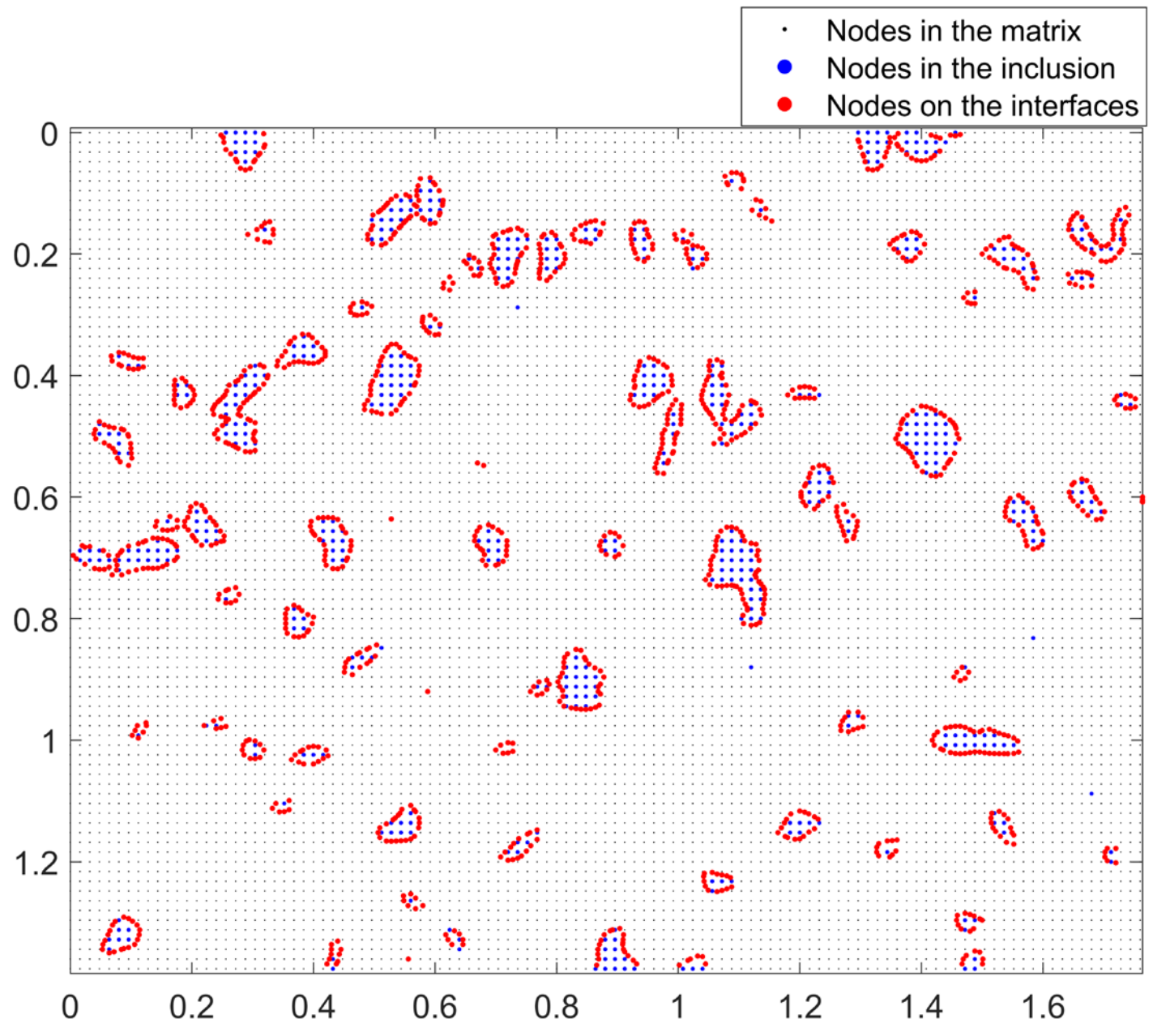

**Figure 15: Constructed RK discretization model for the quantitative validation**

The constructed SVM-RK discretization model is superimposed over the original Micro-CT image, and the result is shown in Figure 16. Furthermore, the alumina inclusions enclosed by the identified interface nodes in the constructed SVM-RK discretization model are highlighted and overlaid onto the SEM surface image, which is contrasted with the EDS-layered SEM image shown in Figure 17. As can be seen, the obtained image-based RK discretization model agrees well with the input Micro-CT image in detail.

**Remark 3.5**: *It is worth noting that the surface of the specimen was polished to enhance imaging quality for SEM, which may cause slight alterations to the distribution of surface particles. Figure 18 illustrates the minor discrepancies between the SEM and Micro-CT images. As the blue boxes indicate, misalignments can be observed for certain inclusion particles. Additionally, smaller particles were not captured in the SEM scan, as highlighted in the red boxes. Consequently, the inclusion particles identified by the SVM-RK discretization model exhibit slight variances compared to the EDS elemental mapping result, as illustrated in Figure 17. Nevertheless, they are mostly consistent, particularly for the larger and more distinctive inclusion particles.*

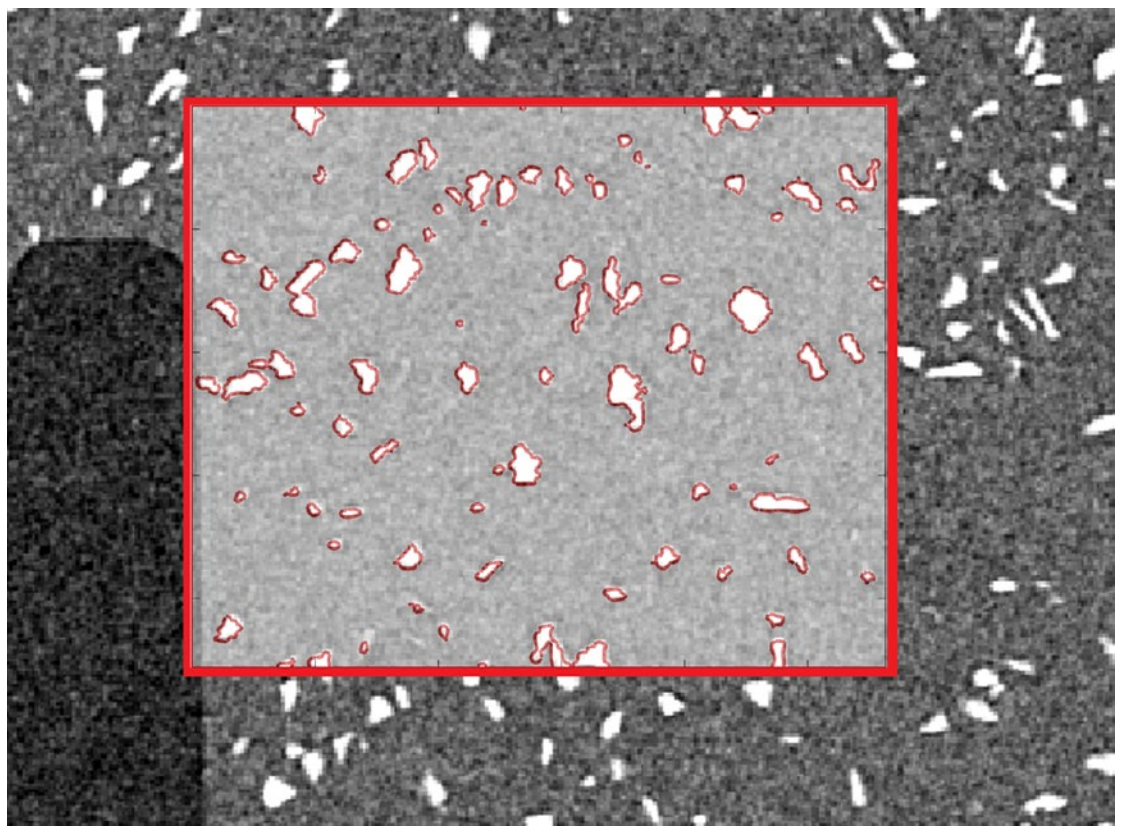

**Figure 16: Constructed SVM-RK discretization model superimposed with the micro-CT input image** (only showing the interface nodes)

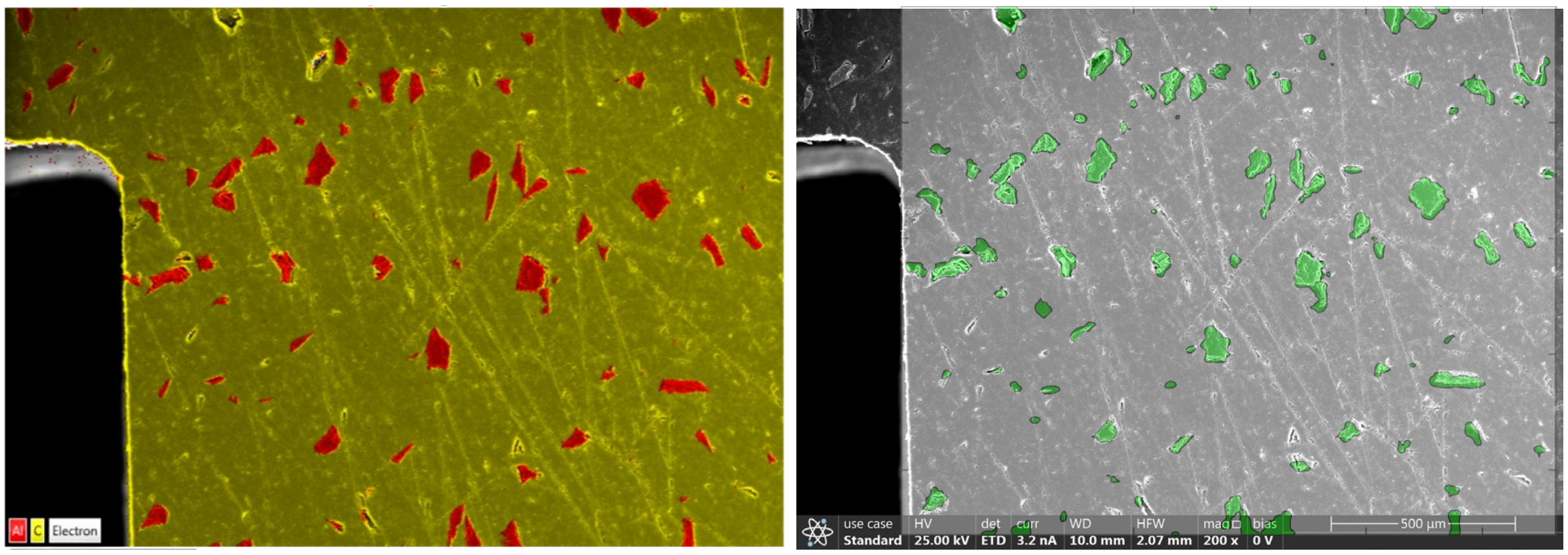

(a) SEM image overlaid with EDS mapping (b) SEM image overlaid with SVM-RK discretization model

**Figure 17: Comparison of EDS overlaid and SVM-RK discretization model overlaid SEM surface images**

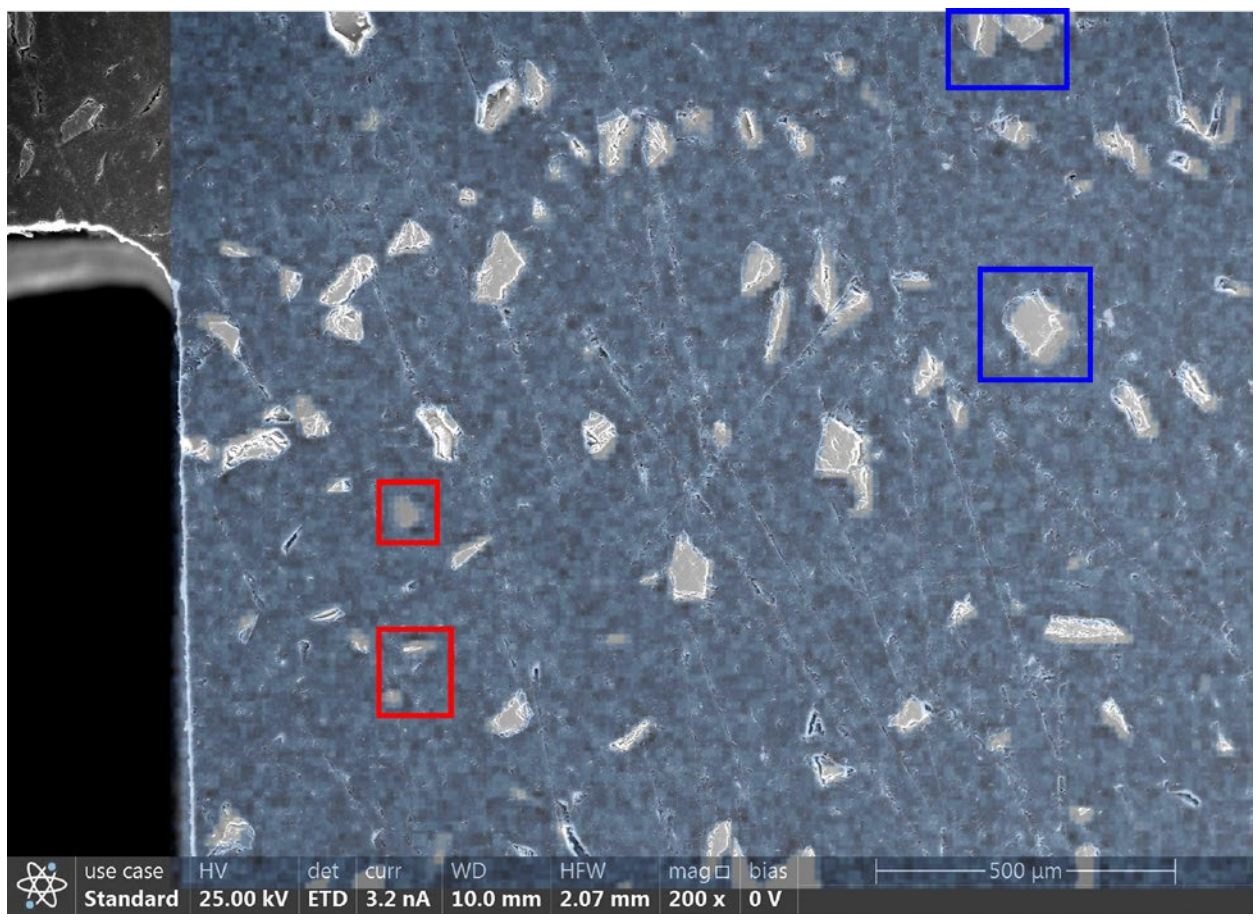

**Figure 18: Comparison between the SEM surface image and Micro-CT image**

# 4 Interface-Modified Reproducing Kernel Approximation Guided by Support Vector Machine

## 4.1 Interface-Modified kernel functions

With the material interface segmented by the SVM, the weak discontinuities across the material interfaces are to be introduced by modifying the regular RK kernel function with a regularized heavy-side function $\widetilde{H}$ as follows:

$$\bar{\phi}_a(\boldsymbol{x}-\boldsymbol{x}_I)=\phi_a(\boldsymbol{x}-\boldsymbol{x}_I)\widetilde{H}\left(\bar{\xi}_I(\boldsymbol{x})\right) \tag{42}$$

where $\bar{\phi}_a(\boldsymbol{x}-\boldsymbol{x}_I)$ is a modified kernel function, and $\widetilde{H}(\cdot)$ and $\bar{\xi}_I(\boldsymbol{x})$ in Eq. (42) are defined as:

$$\widetilde{H}(\cdot)=\max(0,\tanh(\cdot)) \tag{43}$$

and

$$\bar{\xi}_I(\boldsymbol{x})=\begin{cases}-\dfrac{S(\boldsymbol{x})}{c}, & S(\boldsymbol{x}_I)<0\\ +\dfrac{S(\boldsymbol{x})}{c}, & S(\boldsymbol{x}_I)>0\end{cases} \tag{44}$$

where $S(\boldsymbol{x})$ is the score function, and $c$ denotes a scaling factor that has a length of the order of nodal spacing. Note that $S(\boldsymbol{x})$ is a signed distance of an evaluation point to its nearby interface, which is given from the output of the SVM-RK image segmentation and is readily available for evaluation of regularized heavy-side function $\widetilde{H}$. This normalized distance measure $\bar{\xi}(\boldsymbol{x})$ is applicable to general n-dimensional image data. Since the kernel functions associated with nodes away from the interfaces have been scaled to zero at the material interfaces by the regularized heavy-side function, the kernel functions associated with the interface nodes are not scaled. As will be discussed in the next section, the "reproduced" RK shape functions via the reproducing conditions given in Eqs. (9)-(10) reveals a weak ($C^0$) continuity of the approximated function at the material interface due to the heavy-side scaling in Eq. (42) regardless of the continuity of kernel function associated with nodes at the material interfaces. Hence, cubic B spline (B3) kernel functions with $C^2$ continuity or power kernel (PK) function with $C^0$ continuity [60] can be considered for the kernel function associated with the interface nodes:

$$\phi_a^{B3}(z)=\begin{cases}\dfrac{2}{3}-4|z|^2+4|z|^3 & \text{for } 0\le|z|\le\dfrac{1}{2}\\ \dfrac{4}{3}-4|z|+4|z|^2-\dfrac{4}{3}|z|^3 & \text{for } \dfrac{1}{2}\le|z|\le 1\\ 0 & \text{otherwise}\end{cases}\qquad z=\frac{\boldsymbol{x}-\boldsymbol{x}_I}{a} \tag{45}$$

$$\phi_a^{PK}(z) = \begin{cases} (1-z)^\alpha & \text{for } 0 \le z \le 1 \\ 0 & \text{otherwise} \end{cases} \qquad z = \frac{\boldsymbol{x} - \boldsymbol{x}_I}{a} \tag{46}$$

Figure 19 shows the un-modified kernel functions $\phi_a(\boldsymbol{x} - \boldsymbol{x}_I)$, regularized Heavy-side function $\widetilde{H}(\bar{\xi}_I(\boldsymbol{x}))$, the interface-modified kernel functions $\bar{\phi}_a(\boldsymbol{x} - \boldsymbol{x}_I)$ and their derivatives $\bar{\phi}_{a,x}(\boldsymbol{x} - \boldsymbol{x}_I)$ in 1D with different choices of interface kernels. The blue, red, and black kernel functions are associated with the interface node, nodes within the support of interface nodes, and nodes away from the interface, respectively.

**Remarks 4.1.**

*1. One can observe that after the interface modification, the influence domains of all nodes, except the interface node, terminate at the interface location, which naturally introduces a weak discontinuity to interface-modified kernel functions, even for the case of the smooth B-spline kernel at the interface.*

*2. The added computational cost to perform the proposed kernel modifications is marginal because only a scaling is applied to the original kernel functions to construct kernel modifications, and this can be done effortlessly for arbitrary spatial dimensions.*

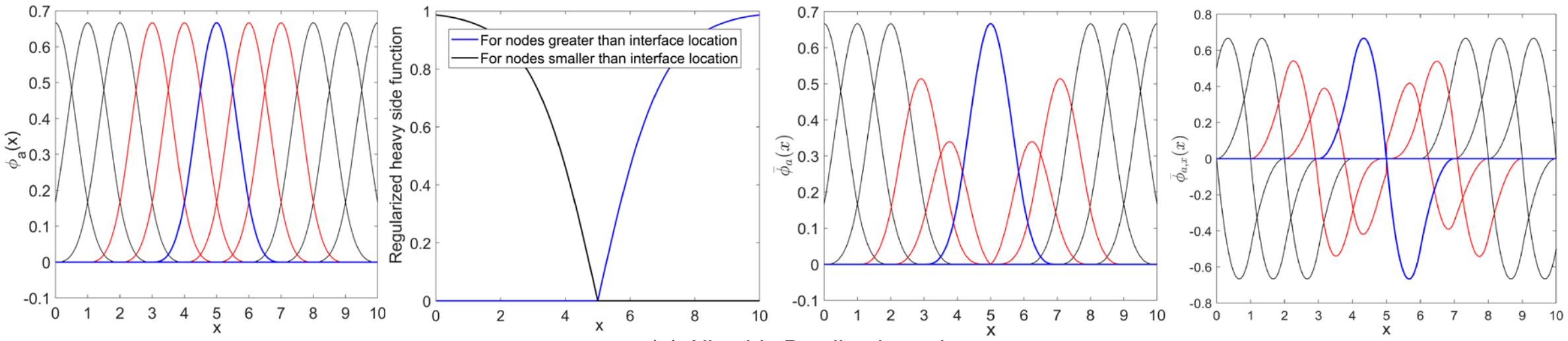

(a) All cubic B spline kernels

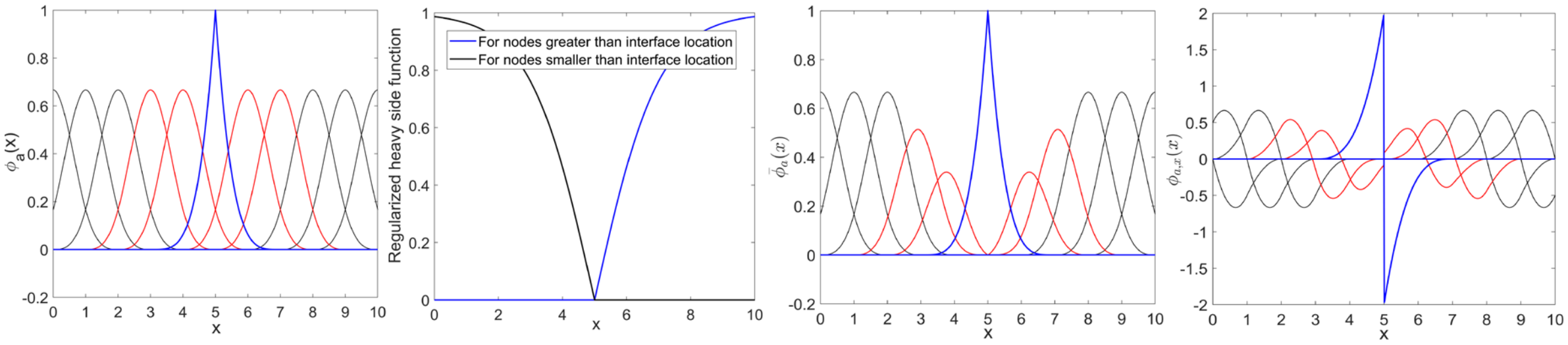

(b) Order 4 power function interface kernel, cubic B spline kernels for the rest

**Figure 19: Plots of 1D interface-modified kernels with different interface kernels** (Left to right: original kernel function, regularized heavy-side scaling function, modified kernel function, modified kernel function's derivative)

## 4.2 Interface-Modified RK (IM-RK) approximation

Let us consider the RK discretization node set $\mathbb{S}^{RK}$ in the SVM-RK discretized numerical model. Recall that interface nodes in $\mathbb{S}^{RK}$are contained in the set $\mathbb{S}^{IF}$. Let $\mathbb{S}^{RK}\backslash\mathbb{S}^{IF}$ denotes the set of all RK discrete points excluding those on the interfaces, then the RK shape function can be written as follows:

$$\begin{cases}\Psi_I(\boldsymbol{x}) = C(\boldsymbol{x};\boldsymbol{x}-\boldsymbol{x}_I)\phi_a(\boldsymbol{x}-\boldsymbol{x}_I) = (\boldsymbol{H}^{\mathrm{T}}(\boldsymbol{x}-\boldsymbol{x}_{\mathbf{I}})\boldsymbol{b}(\boldsymbol{x}))\phi_a(\boldsymbol{x}-\boldsymbol{x}_I), \qquad \forall I \in \mathbb{S}^{IF} \\ \Psi_I(\boldsymbol{x}) = C(\boldsymbol{x};\boldsymbol{x}-\boldsymbol{x}_I)\bar{\phi}_a(\boldsymbol{x}-\boldsymbol{x}_I) = (\boldsymbol{H}^{\mathrm{T}}(\boldsymbol{x}-\boldsymbol{x}_{\mathbf{I}})\boldsymbol{b}(\boldsymbol{x}))\bar{\phi}_a(\boldsymbol{x}-\boldsymbol{x}_I), \forall I \in \mathbb{S}^{RK}\backslash\mathbb{S}^{IF}\end{cases} \tag{47}$$

where $\phi_a(\boldsymbol{x}-\boldsymbol{x}_I)$ is the regular kernel functions without interface modification and $\bar{\phi}_a(\boldsymbol{x}-\boldsymbol{x}_I)$ is the interface-modified kernel function defined Eq. (42). The unknown coefficient vector $\boldsymbol{b}(\boldsymbol{x})$ is obtained by imposing the $n^{th}$ order reproducing conditions, as shown in Eq. (10), which can also be expressed in terms of the basis vector as:

$$\sum_{I=1}^{NP}\Psi_I(\boldsymbol{x})\boldsymbol{H}(\boldsymbol{x}-\boldsymbol{x}_I) = \boldsymbol{H}(\mathbf{0}) \tag{48}$$

Substituting Eq. (47) into Eq. (48) yields:

$$\boldsymbol{b}(\boldsymbol{x}) = \bar{\boldsymbol{M}}^{-1}(\boldsymbol{x})\boldsymbol{H}(\mathbf{0}) \tag{49}$$

where $\bar{\boldsymbol{M}}(\boldsymbol{x})$ is the modified moment matrix:

$$\begin{aligned}\bar{\boldsymbol{M}}(\boldsymbol{x}) = &\sum_{I\in\mathbb{S}^{IF}}\boldsymbol{H}^{\mathrm{T}}(\boldsymbol{x}-\boldsymbol{x}_I)\boldsymbol{H}(\boldsymbol{x}-\boldsymbol{x}_I)\phi_a(\boldsymbol{x}-\boldsymbol{x}_I) \\ &+ \sum_{I\in\mathbb{S}^{RK}\backslash\mathbb{S}^{IF}}\boldsymbol{H}^{\mathrm{T}}(\boldsymbol{x}-\boldsymbol{x}_I)\boldsymbol{H}(\boldsymbol{x}-\boldsymbol{x}_I)\bar{\phi}_a(\boldsymbol{x}-\boldsymbol{x}_I)\end{aligned} \tag{50}$$

Finally, by substituting Eq. (49) into Eq. (47), the interface modified reproducing kernel shape function is obtained as:

$$\bar{\Psi}_I(\boldsymbol{x}) = \begin{cases}\boldsymbol{H}^T(\mathbf{0})\bar{\boldsymbol{M}}^{-1}(\boldsymbol{x})\boldsymbol{H}(\boldsymbol{x}-\boldsymbol{x}_I)\phi_a(\boldsymbol{x}-\boldsymbol{x}_I), \qquad \forall I \in \mathbb{S}^{IF} \\ \boldsymbol{H}^T(\mathbf{0})\bar{\boldsymbol{M}}^{-1}(\boldsymbol{x})\boldsymbol{H}(\boldsymbol{x}-\boldsymbol{x}_I)\bar{\phi}_a(\boldsymbol{x}-\boldsymbol{x}_I), \qquad \forall I \in \mathbb{S}^{RK}\backslash\mathbb{S}^{IF}\end{cases} \tag{51}$$

Finally, the IM-RK approximation of the displacement field $\boldsymbol{u}^h(\boldsymbol{x})$ is expressed as:

$$u_i^h(\boldsymbol{x}) = \sum_{I=1}^{NP}\bar{\Psi}_I(\boldsymbol{x})d_{iI} \tag{52}$$

As shown in Eq. (52), no duplicated degrees of freedom associated with the interface nodes are added (such as interface enrichments) when using the IM-RK to approximate the displacement field. Figure 20 compares the 1D traditional RK and IM-RK shape functions and their derivatives in a domain [0.0, 10] with a material interface at $x = 5$. The red-colored node in Figure 20 is the interface node, and the shape functions colored black, blue, and red are associated with nodes outside the support of the interface node, nodes within the influence of the interface node, and the interface node, respectively.

**Remarks 4.2.**

*1. Owing to the regularization function* $\widetilde{H}(\bar{\xi}_I(\boldsymbol{x}))$ *introduced in Eqs. (43)-(44), nodes on different sides of the interface lose communication, leading to weak discontinuities in the IM-RK shape functions. This is true even when a smooth cubic B-spline kernel is used for all nodes, including the interface nodes, as shown in Figure 20. Same weak discontinuity properties exist in high dimensions, as shown in Figure 21, regardless of the smoothness of interface kernel functions. Figure 22 illustrates the IM-RK interpolation of a function:*

$$f(\boldsymbol{x}) = \begin{cases} e^{c_1\|\boldsymbol{x}\|^2}, & if\ \|\boldsymbol{x}\| \leq R \\ e^{c_2\|\boldsymbol{x}\|^2} + (e^{c_1R^2} - e^{c_2R^2}), & if\ \|\boldsymbol{x}\| > R \end{cases}$$

*where* $c_1 = 0.5$, $c_2 = 0.1$, *and the interface is a circular arc with* $R = 0.8$. *IM-RK shape functions with smooth cubic B-spline kernels can effectively capture weak discontinuities along the interface in the interpolated function and represent its discontinuous x-directional derivative field, while both the interpolated function and its x-directional derivative fields are smooth across the interface when using the standard RK shape functions with B3 kernels.*

*2. Since all kernel functions vanish at the interface, except for the kernel functions defined on the interface nodes, the resulting IM-RK approximation functions possess weak Kronecker delta properties, as have been discussed in [38]. These weak Kronecker delta properties, however, do not exist in high dimensions because the supports of interface nodes overlap except for interface nodes located on domain boundaries.*

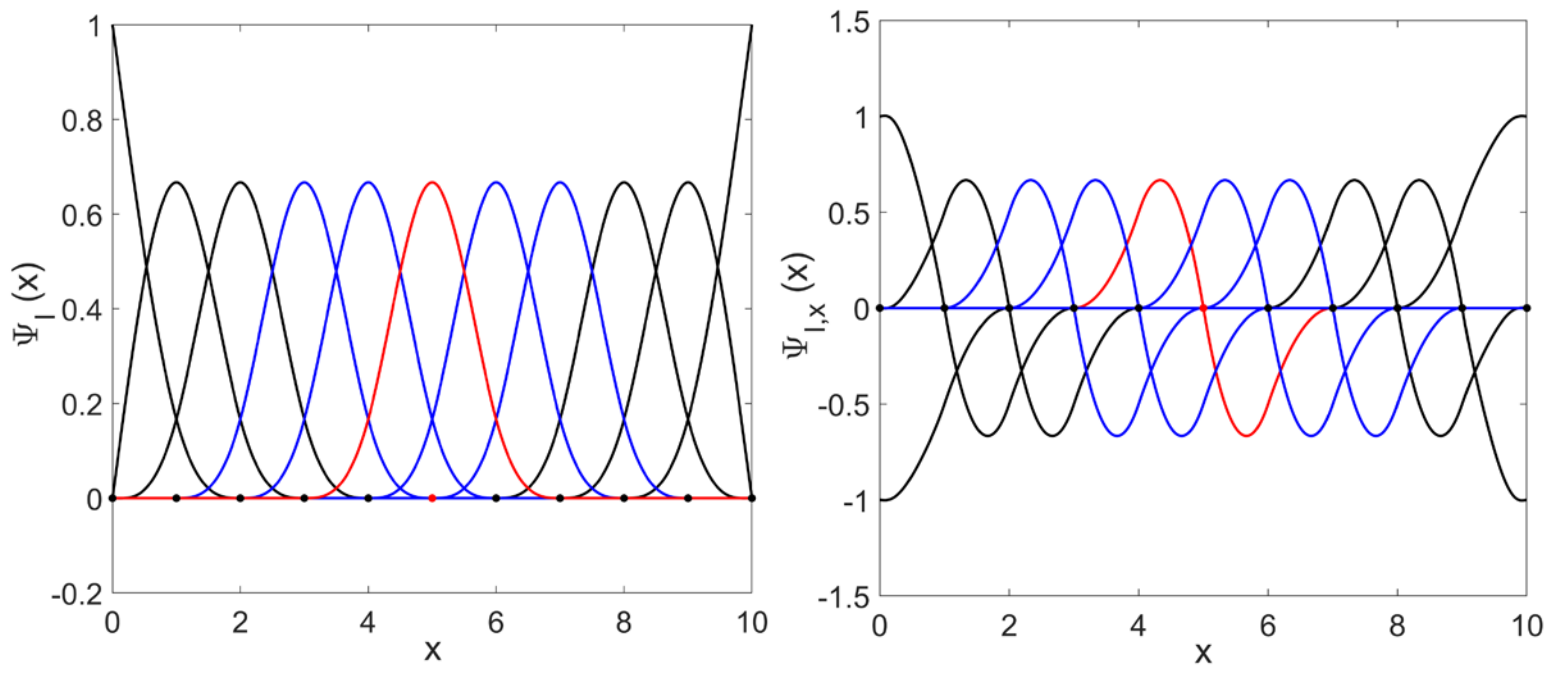

(a) RKPM shape function and derivative with cubic B spline kernels

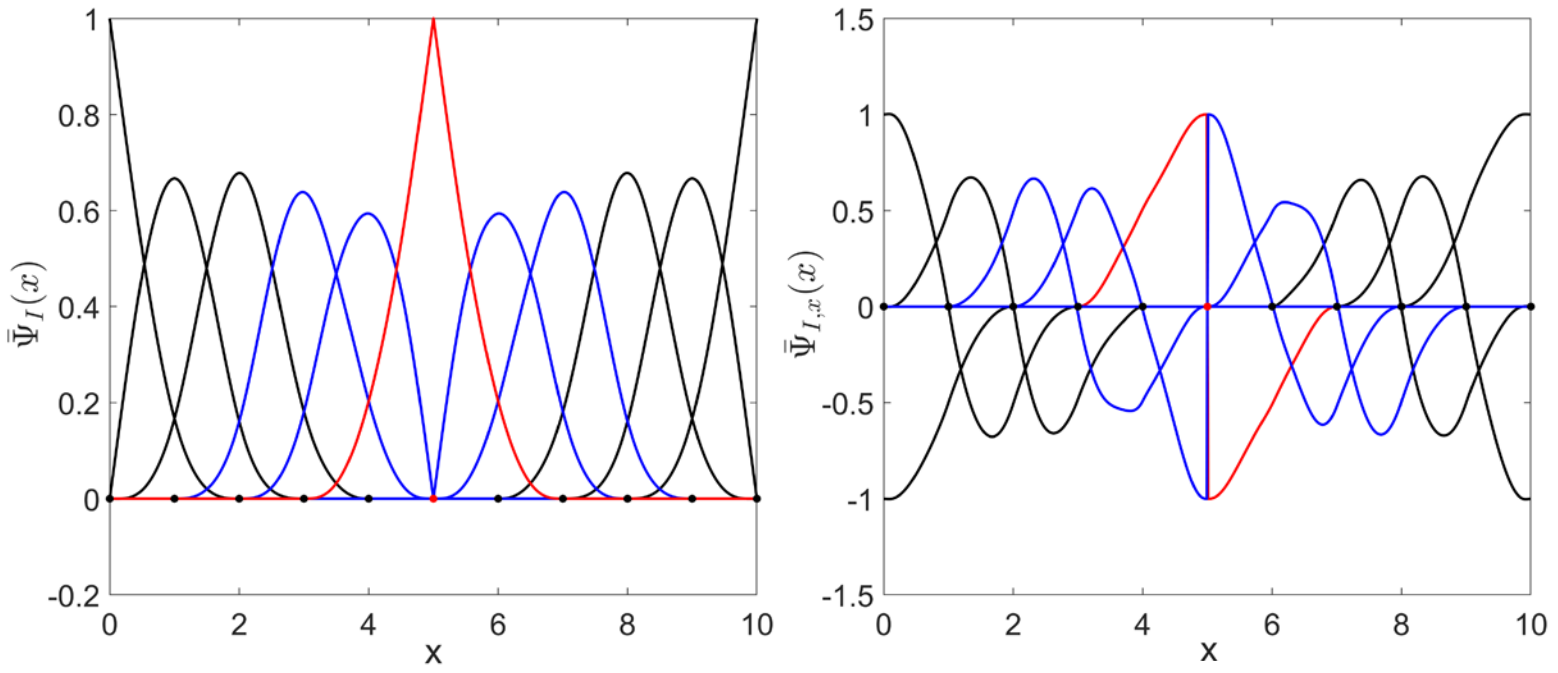

(b) IM-RKPM shape function and derivative with cubic B spline interface kernel

**Figure 20: 1D traditional RK and IM-RK shape functions and derivatives**

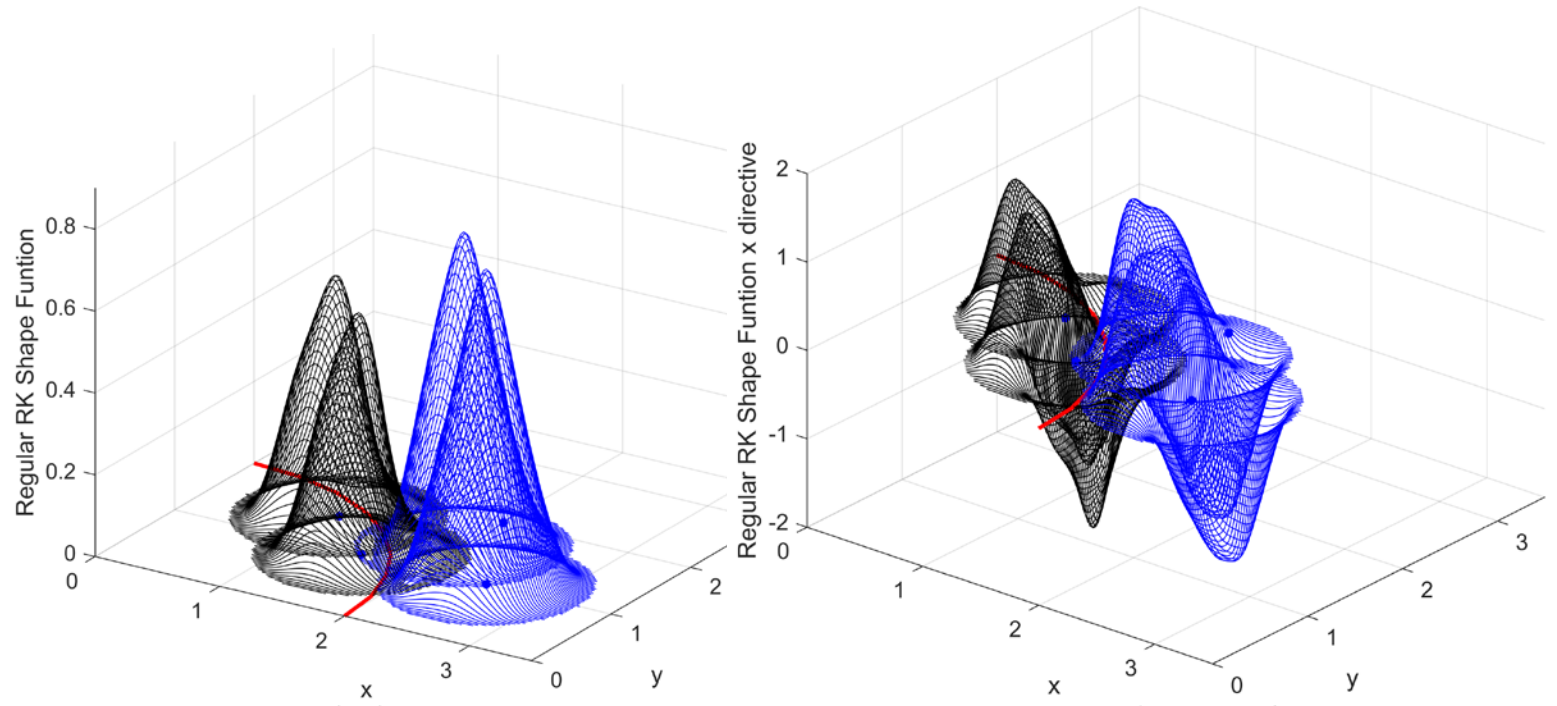

(a) 2D RKPM shape function and x derivative with cubic B spline kernels

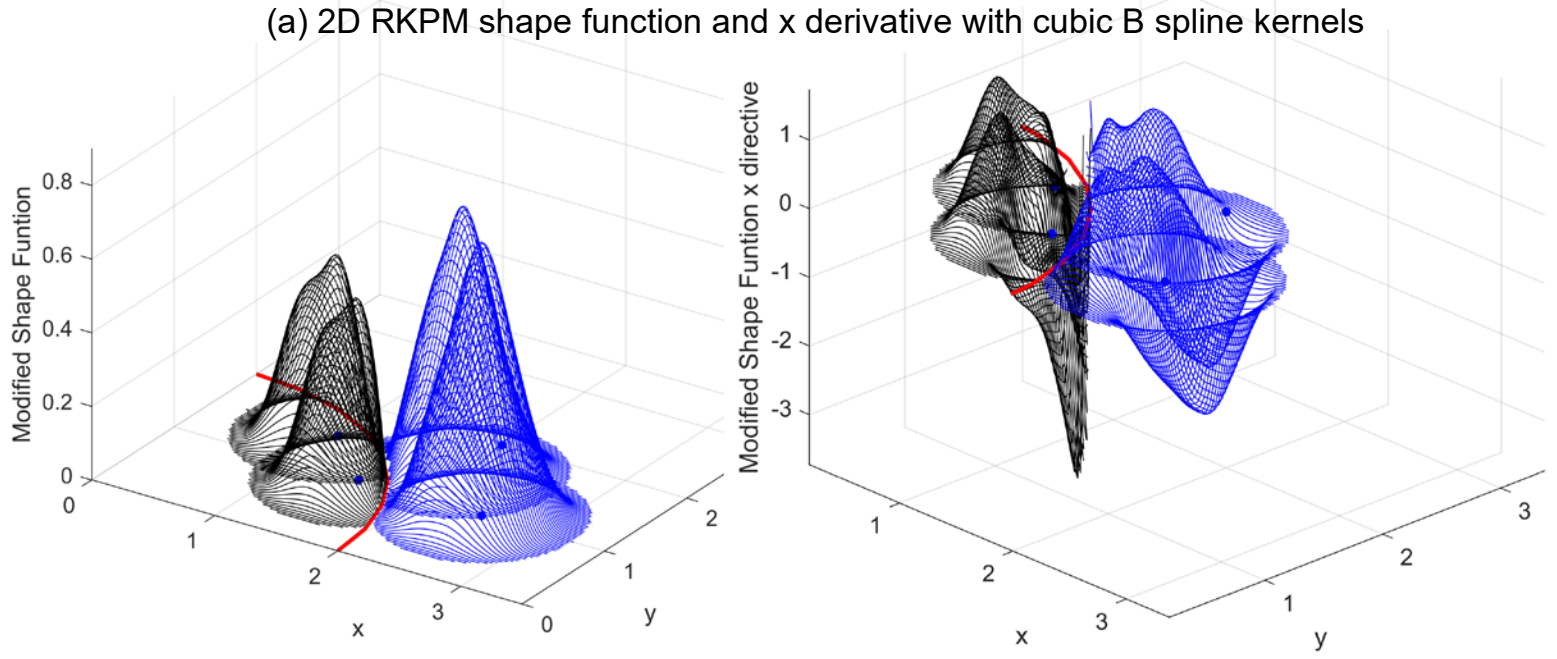

(b) 2D M-RKPM shape function and x derivative with cubic B spline interface kernel

**Figure 21: 2D IM-RK shape functions constructed with different interface kernels using (a) cubic B-spline interface kernels and (b) 4th order power interface kernels** (the blue and the black functions on the left represent IM-RK shape function on the opposite sides of the material interface)

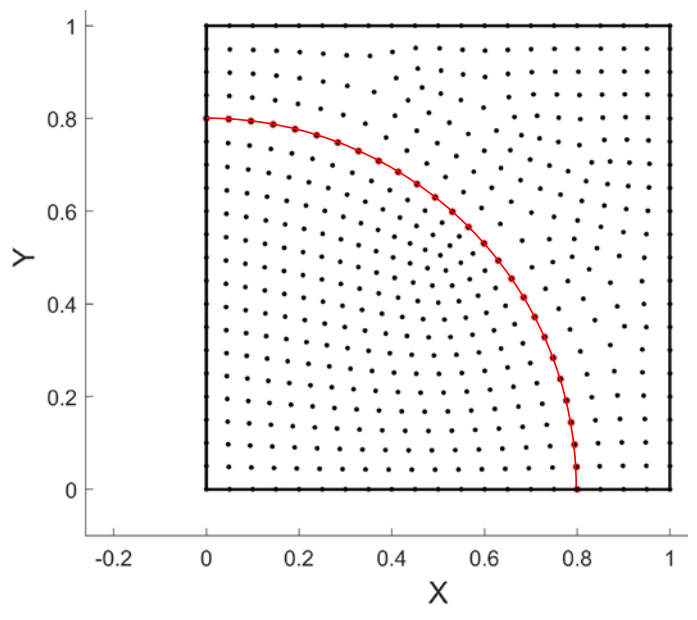

(a) RK node distribution (red arc indicates the interface)

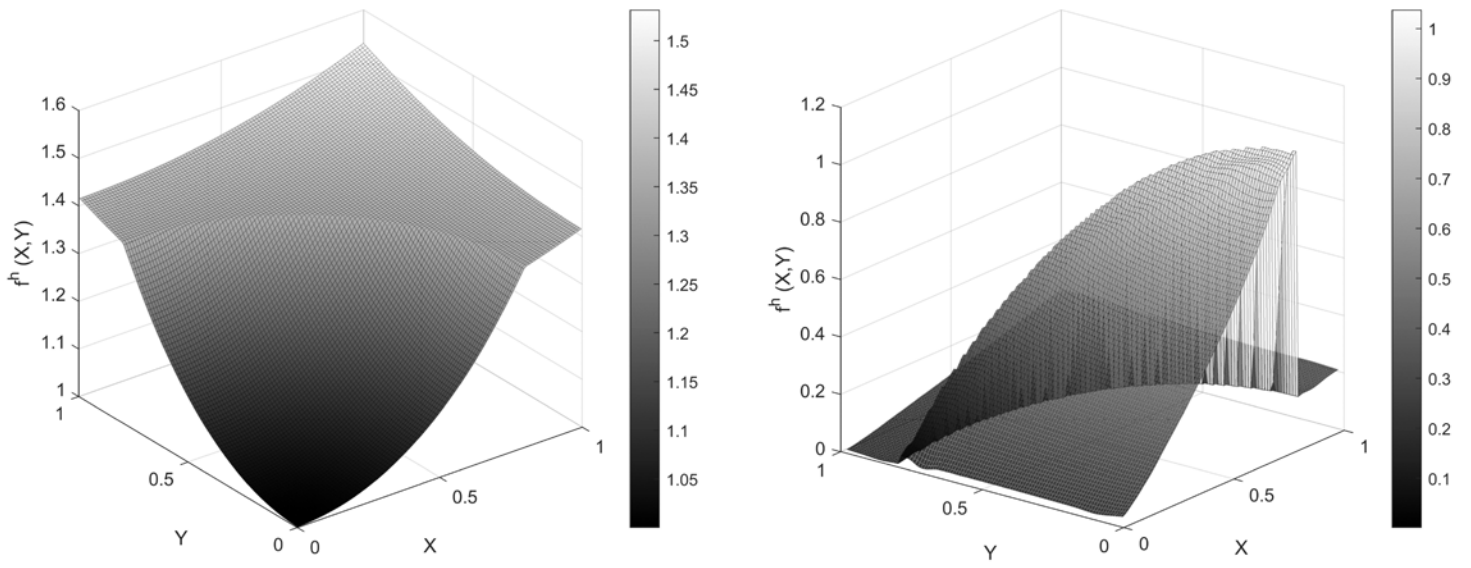

(b) IM-RK function interpolation with B3 kernel: interpolated function (left) and x-directional derivative (right)

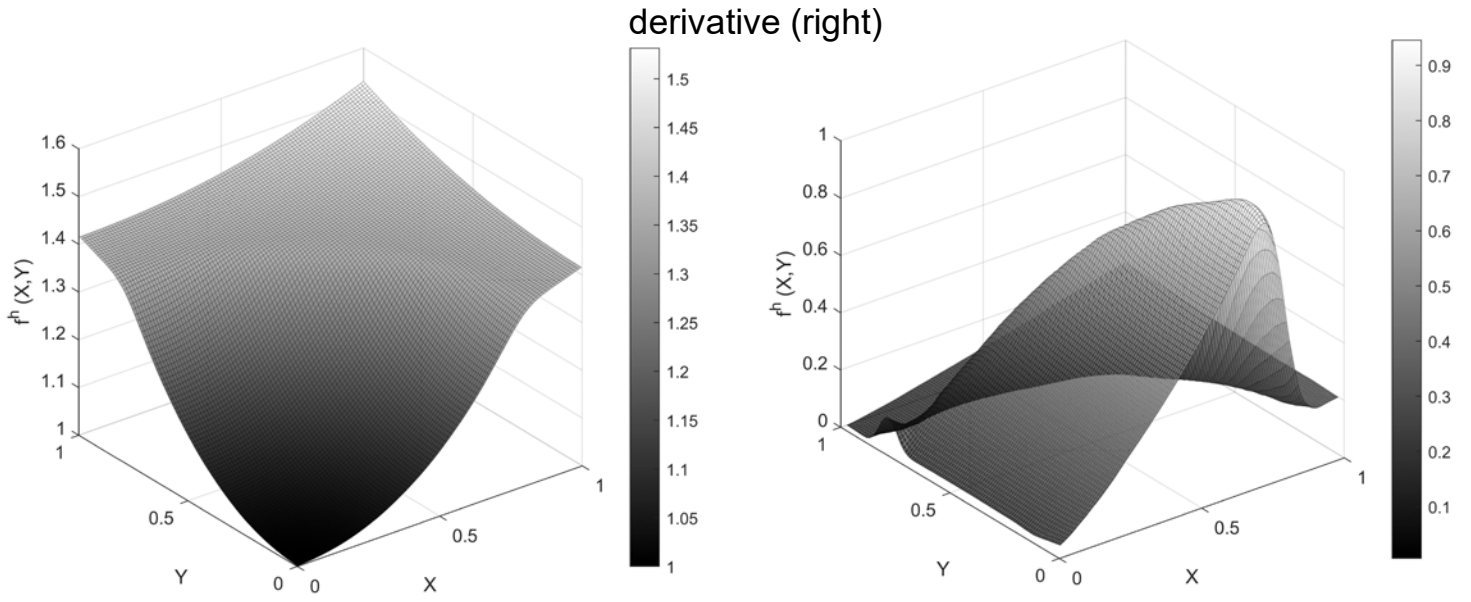

(c) Standard RK function interpolation with B3 kernel: interpolated function (left) and x-directional derivative (right)

**Figure 22: IM-RK interpolation of a 2D function**

In this work, the meshfree method using the above IM-RK approximation as the approximation function for the test and trial functions under the Galerkin framework is named the Interface-Modified Reproducing Kernel Particle Method (IM-RKPM).

## 4.3 IM-RK shape functions for the image-based numerical model

Figure 23 and Figure 24 respectively show the IM-RK shape functions of non-interface nodes near the interfaces and the IM-RK shape functions of the interface nodes, constructed on the image-based SVM-RK discretized model shown in Figure 10.

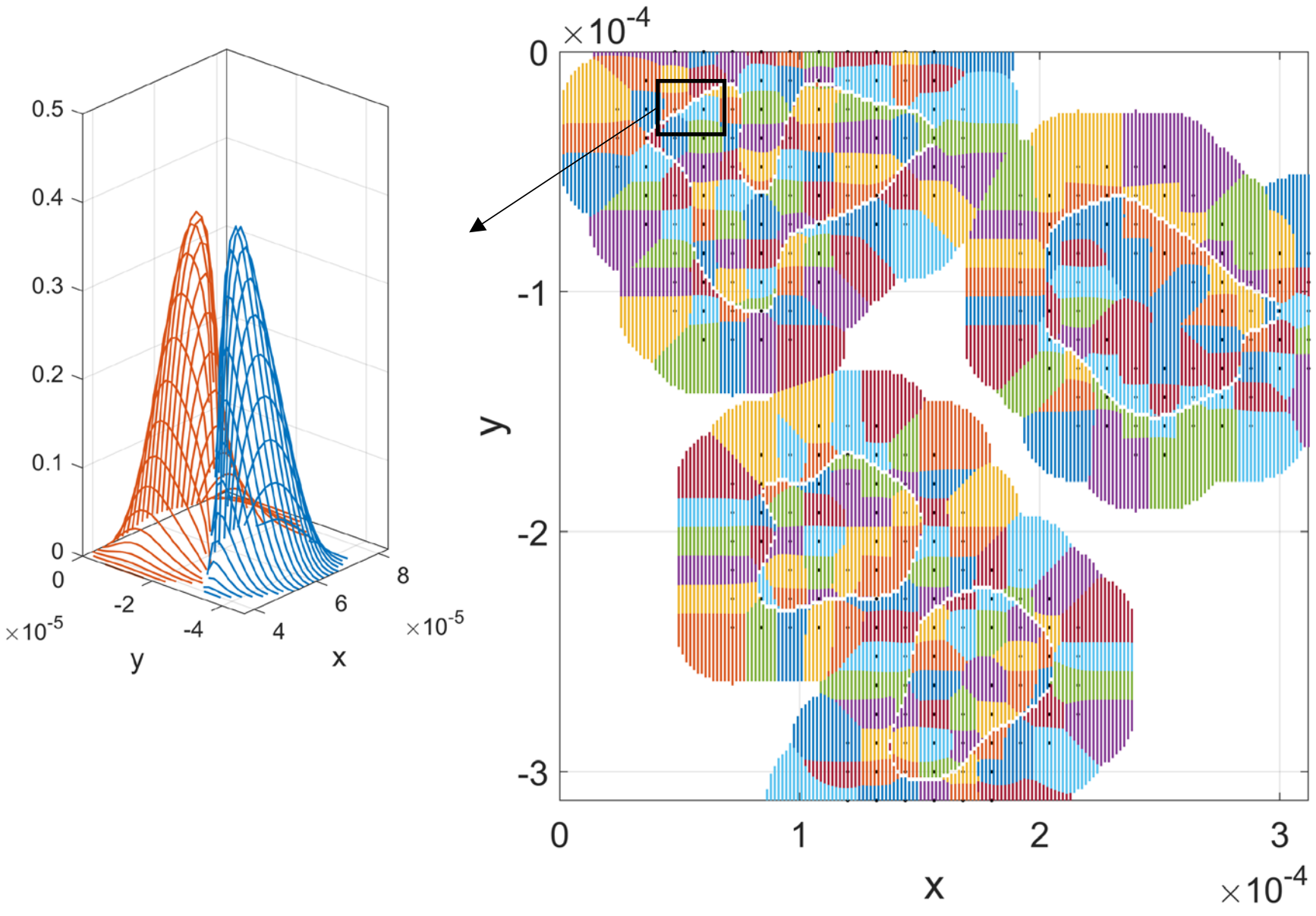

**Figure 23: IM-RK shape functions for nodes around the interfaces: top view (right) and the zoom-in plot of two shape functions in the black box (left)**

In Figure 23 and Figure 24, the non-zero IM-RK shape functions are color-coded by different color blocks, and the maximum shape function is shown at each plotting point in the top view. By observing the results in Figure 23, the shape functions are truncated across arbitrarily shaped interfaces. The interface nodes' shape functions, however, provide support coverage to the nodes located on both sides of the interface with $C^0$ continuity along the interfaces' normal direction for embedding weak discontinuities normal to the interface.

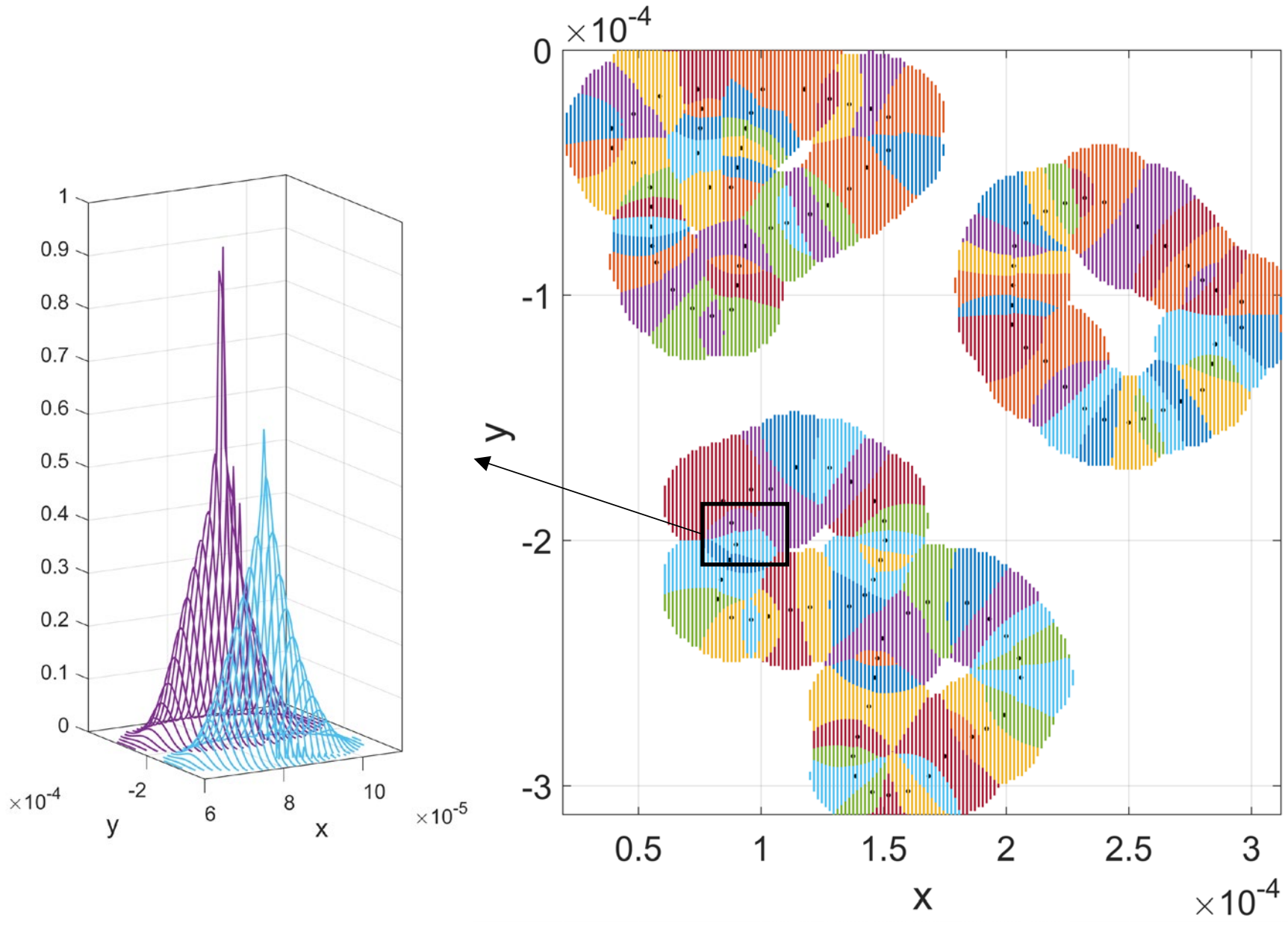

**Figure 24: IM-RK shape function for interface nodes: top view (right) and the zoom-in plot of two shape functions in the black box (left)**

## 4.4 Verification of the IM-RKPM

### 4.4.1 One-dimensional composite rod problem

A one-dimensional composite rod with a centered material interface is fixed at the left and is subjected to a displacement of 1 on the right end, as demonstrated in Figure 25. The rod is also subjected to a polynomial body force up to the third order $b(x) = a_0 + a_1 x + a_2 x^2 + a_3 x^3$. The Young's modulus of the two sections of the rod are set as $E_1 = 10000, \text{for } x \in [0,5]$ and $E_2 = 1000, \text{for } x \in ]5,10]$. The exact solution to this problem is provided in [25].

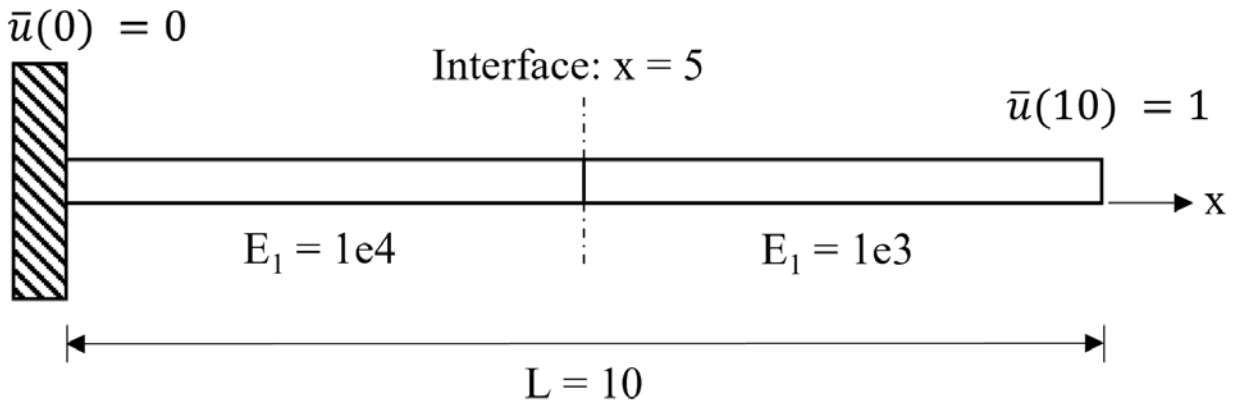

**Figure 25: Schematic of the 1D bi-material rod problem**

The example is analyzed with two body force cases: (1) $b = 0$; (2) $b(x) = 25x - 7.5x^2 + 0.5x^3$. The 1D problem domain is discretized with 11 uniformly distributed nodes, and the problem is approximated using a linear basis in both standard RKPM and IM-RKPM with a constant normalized nodal support size of 2. SCNI and 5-point Gauss integration are selected as numerical integration methods for case (1) and case (2), respectively. Figure 26 and Figure 27 demonstrate the approximated displacement and strain solutions using RKPM and IM-RKPM for case (1), respectively. The results show that RKPM strain solution exhibits Gibb's-like oscillations and fails to reproduce the exact weak discontinuity at the material interface. On the other hand, IM-RKPM with SCNI can precisely capture the displacement and strain field to the machine's precision.

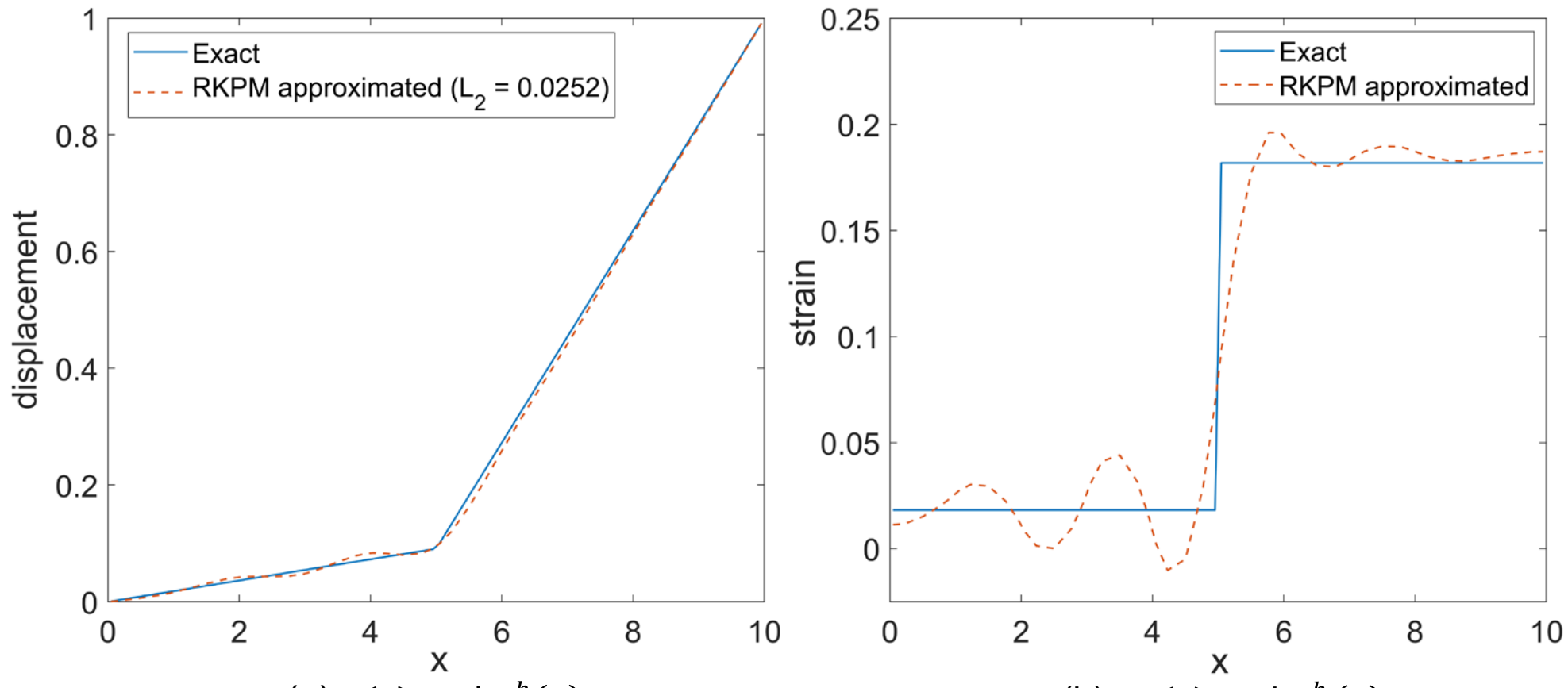

(a) $u(x)$ and $u^h(x)$ (b) $u_{,x}(x)$ and $u^h_{,x}(x)$

**Figure 26: Case (1) RKPM approximated solutions compared to the analytical solutions**

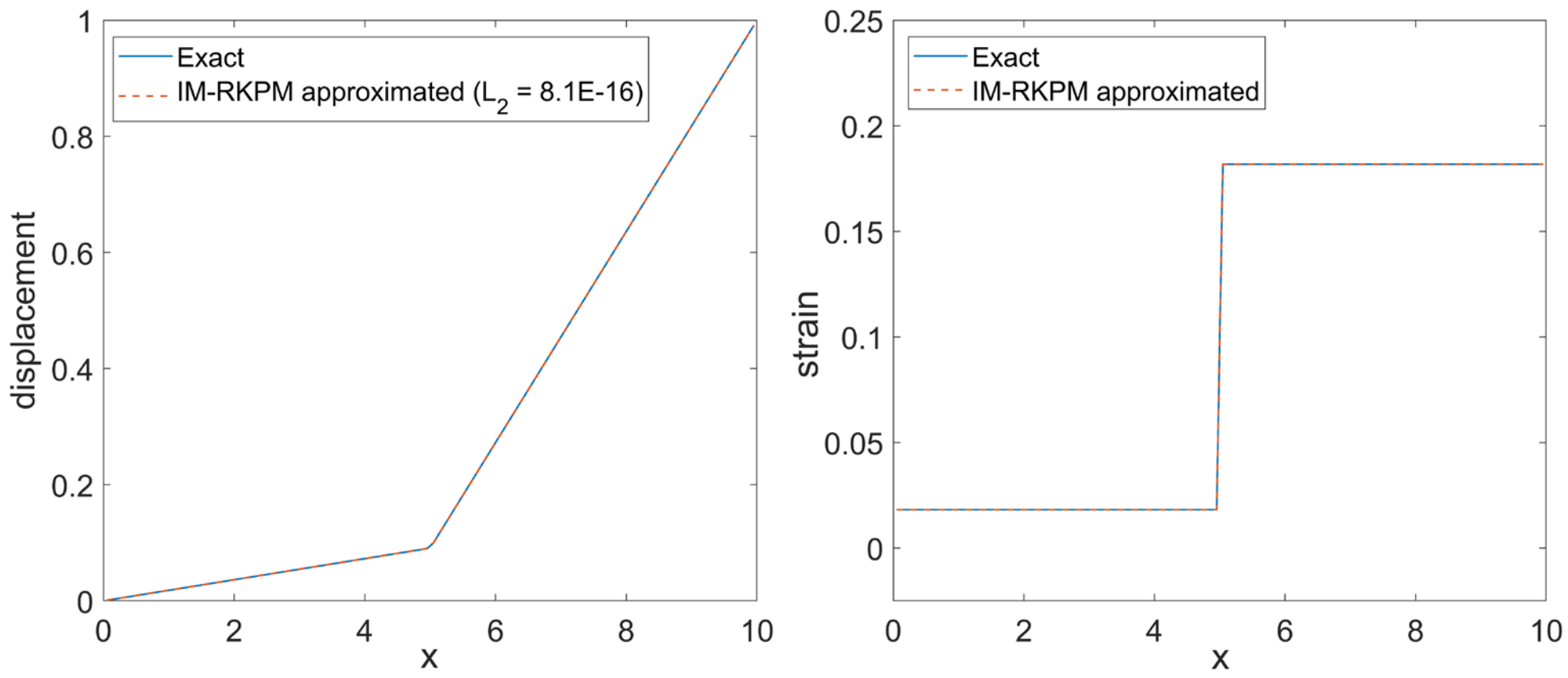

(a) $u(x)$ and $u^h(x)$ (b) $u_{,x}(x)$ and $u^h_{,x}(x)$

**Figure 27: Case (1) IM-RKPM approximated solutions compared to the analytical solutions**

Similar behaviors are observed for case (2), as illustrated in Figure 28 and Figure 29, IM-RKPM significantly reduces the oscillations of the strain solution and can accurately capture the weak discontinuity across the material interface.

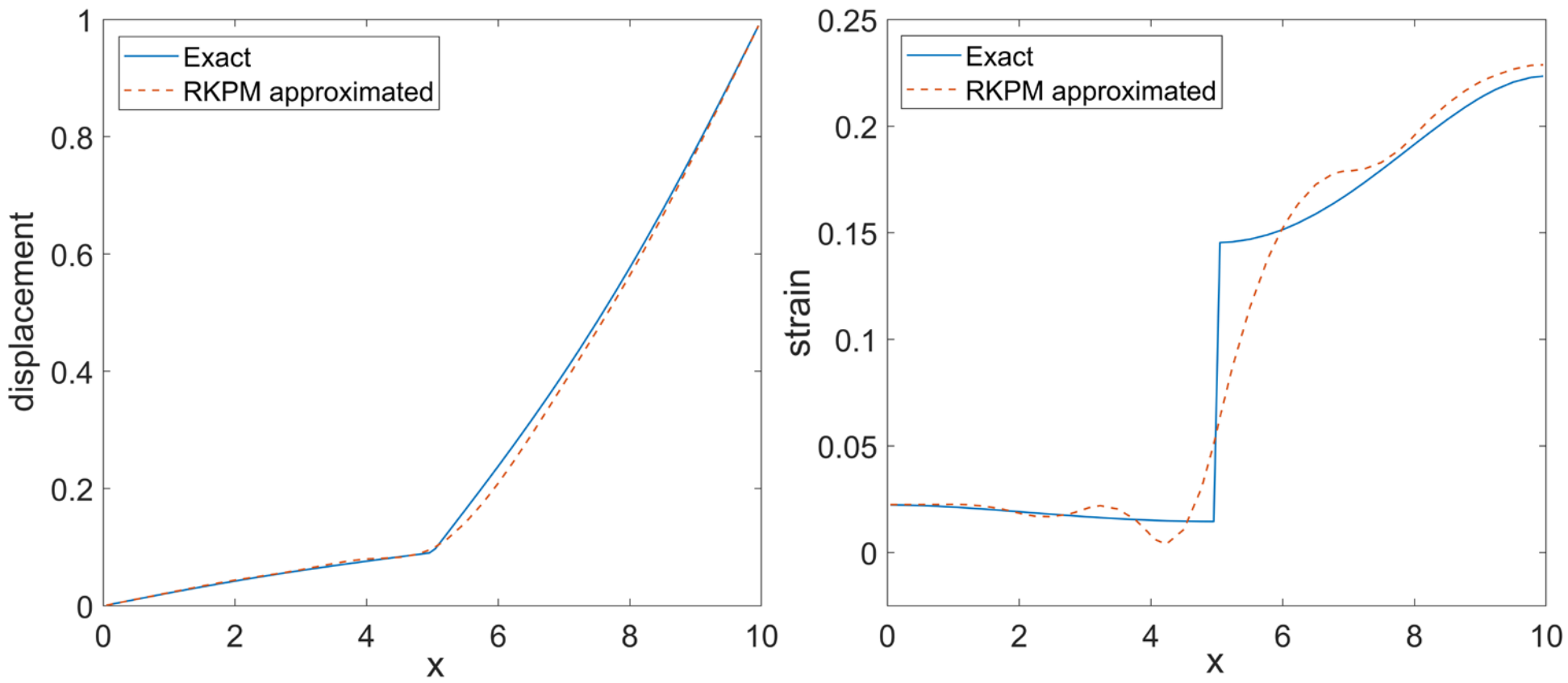

(a) $u(x)$ and $u^h(x)$ (b) $u_{,x}(x)$ and $u^h_{,x}(x)$

**Figure 28: Case (2) RKPM approximated solutions compared to the analytical solutions**

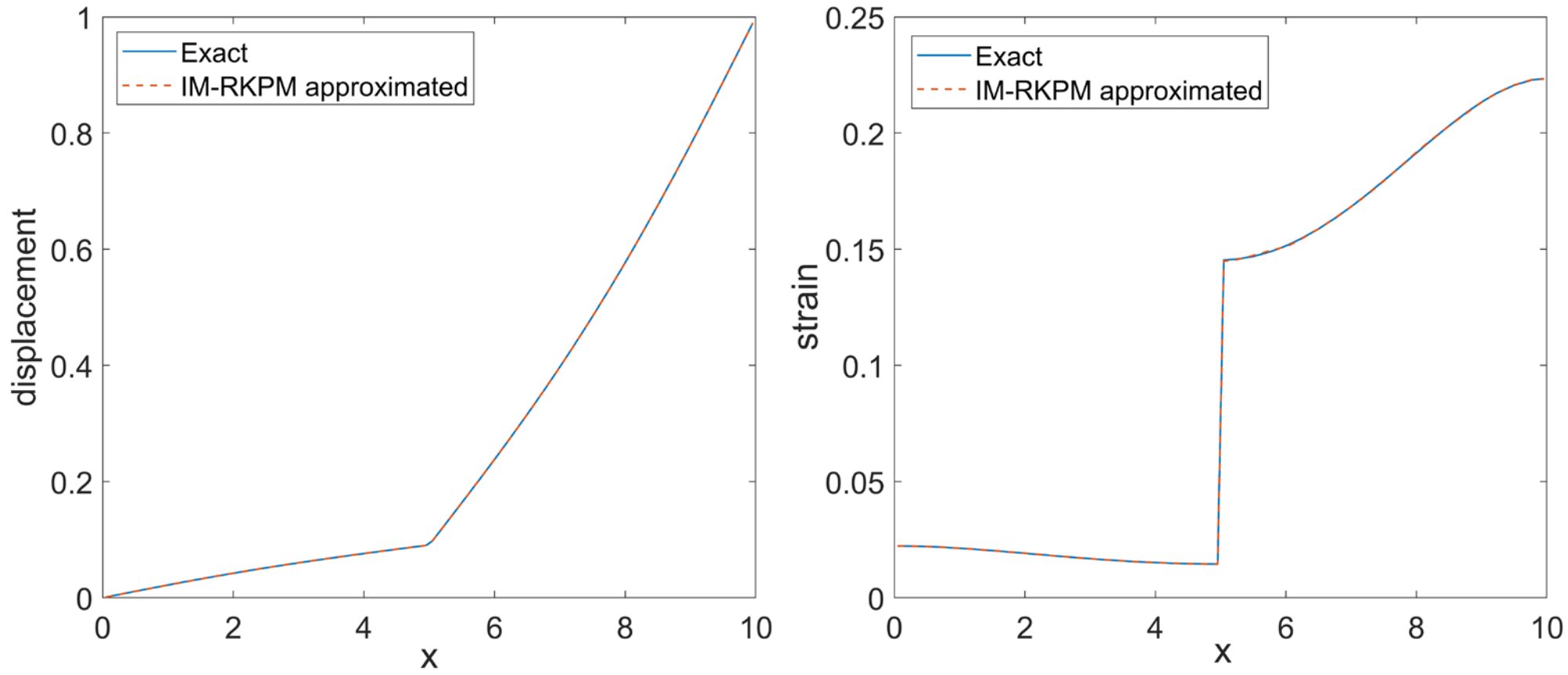

(a) $u(x)$ and $u^h(x)$ (b) $u_{,x}(x)$ and $u^h_{,x}(x)$

**Figure 29: Case (2) IM-RKPM approximated solutions compared to the analytical solutions**

In addition, the convergence behaviors of IM-RKPM with the cubic B spline and the power interface kernels and standard RKPM are investigated in terms of the normalized displacement and energy error norms as follows with high-order Gauss quadrature rule:

$$\left\|\boldsymbol{u}-\boldsymbol{u}^h\right\|_0 = \sqrt{\frac{\int_\Omega \left(\boldsymbol{u}^{\text{exact}}(\boldsymbol{x})-\boldsymbol{u}^h(\boldsymbol{x})\right)^2 d\Omega}{\int_\Omega \left(\boldsymbol{u}^{\text{exact}}(\boldsymbol{x})\right)^2 d\Omega}} \tag{53}$$

$$\left\|\boldsymbol{u}-\boldsymbol{u}^h\right\|_E = \sqrt{\frac{\int_\Omega \left(\boldsymbol{\varepsilon}^{\text{exact}}(\boldsymbol{x})-\boldsymbol{\varepsilon}^h(\boldsymbol{x})\right)\cdot\left(\boldsymbol{\sigma}^{\text{exact}}(\boldsymbol{x})-\boldsymbol{\sigma}^h(\boldsymbol{x})\right) d\Omega}{\int_\Omega \boldsymbol{\varepsilon}^{\text{exact}}(\boldsymbol{x})\cdot\boldsymbol{\sigma}^{\text{exact}}(\boldsymbol{x}) d\Omega}} \tag{54}$$

As illustrated in Figure 30, standard RKPM exhibits a suboptimal convergence rate of 1 for the displacement norms and 0.5 for the energy norm, while the accuracy and convergence rates are substantially improved in IM-RKPM, restoring the optimal convergence rates of 2 and 1, independent to the continuity of the interface kernel function.

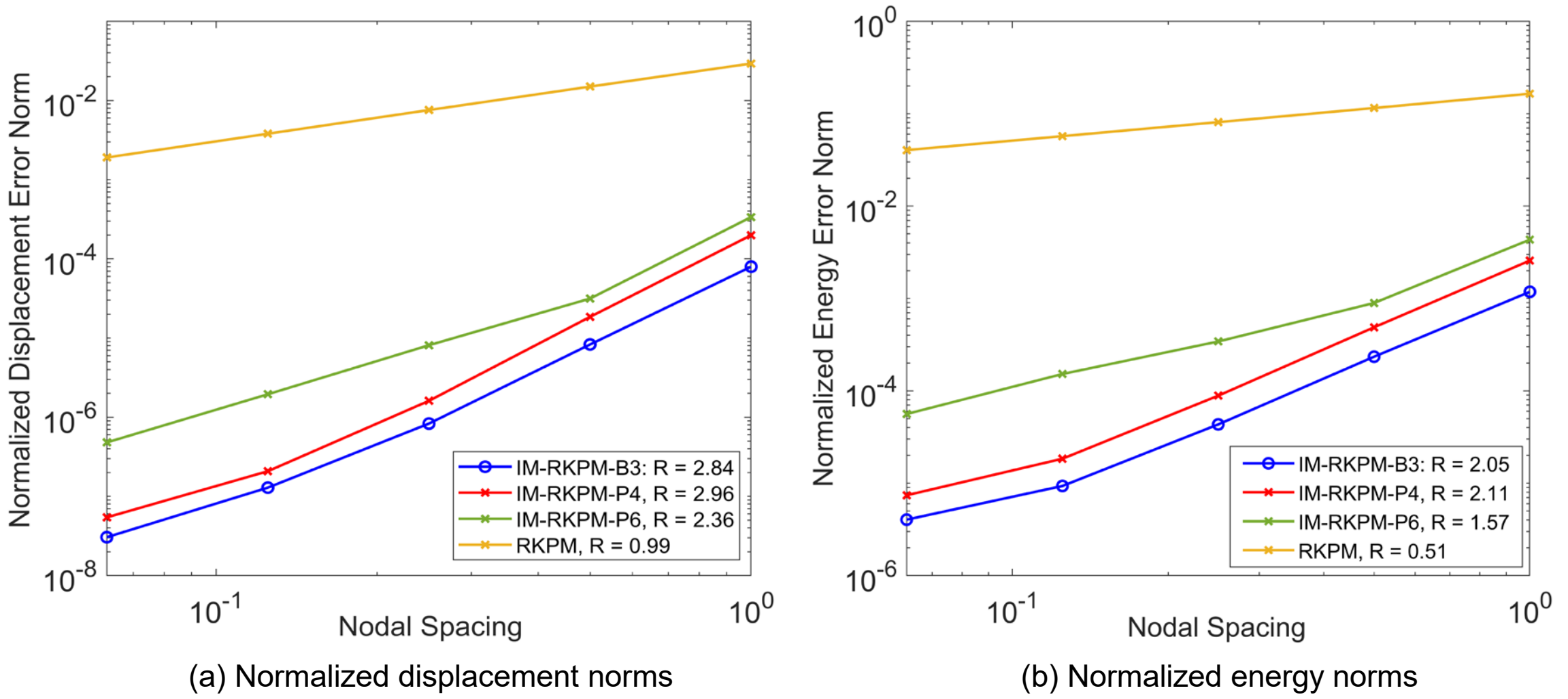

**Figure 30: Accuracy of RKPM and IM-RKPM with different interface kernels** (R: rate of convergence)

### 4.4.2 2D circular inclusion in an infinite plate

An infinite plate with a circular inclusion subjected to a constant dilatational eigenstrain $\varepsilon^* = 0.01$, as shown in Figure 31, is analyzed.

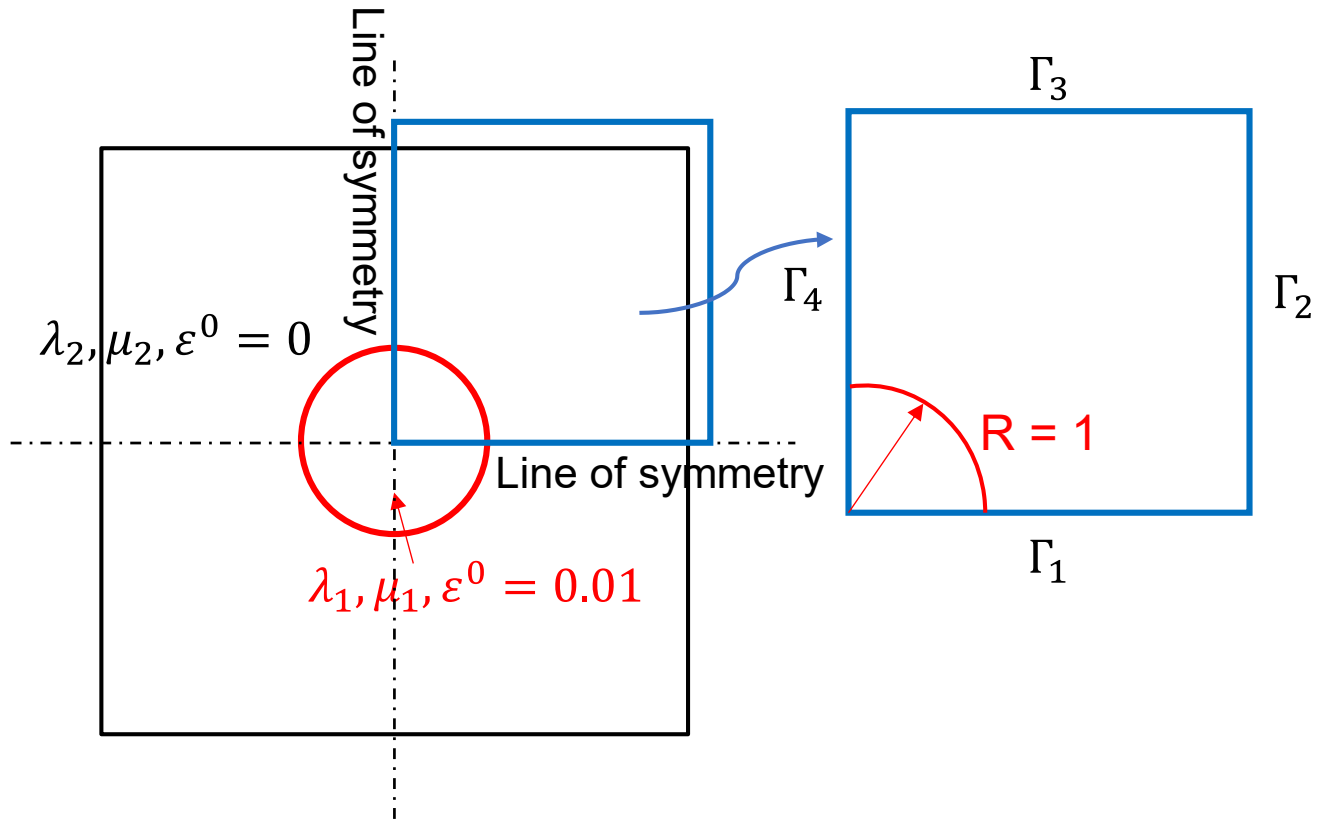

**Figure 31: Schematic of the 2D infinite plate with circular inclusion problem**

The material constants selected for the inclusion material are: $\lambda_1 = 497.16, \mu_1 = 390.63$, and matrix material are: $\lambda_2 = 656.79, \mu_2 = 338.35$, where $\lambda$ and $\mu$ are Lamé parameters. Due to the symmetry of the domain and loading conditions, only the upper right quadrant of the domain is modeled. The length of each side of the finite quarter domain is 5, the radius of the circular inclusion is $R = 1$, and an analytical displacement field is prescribed on the boundaries. The analytical solutions in cylindrical coordinates can be found in [61].

The example is modeled as a plane strain axisymmetric problem. Both $5 \times 5$ Gauss integration and SCNI method are employed as the numerical integration schemes, and RK approximation with linear basis and a normalized support size of 2 are utilized throughout the numerical analysis. Figure 32 demonstrates an example of domain discretization and background integration cell arrangement for the Gauss integration.

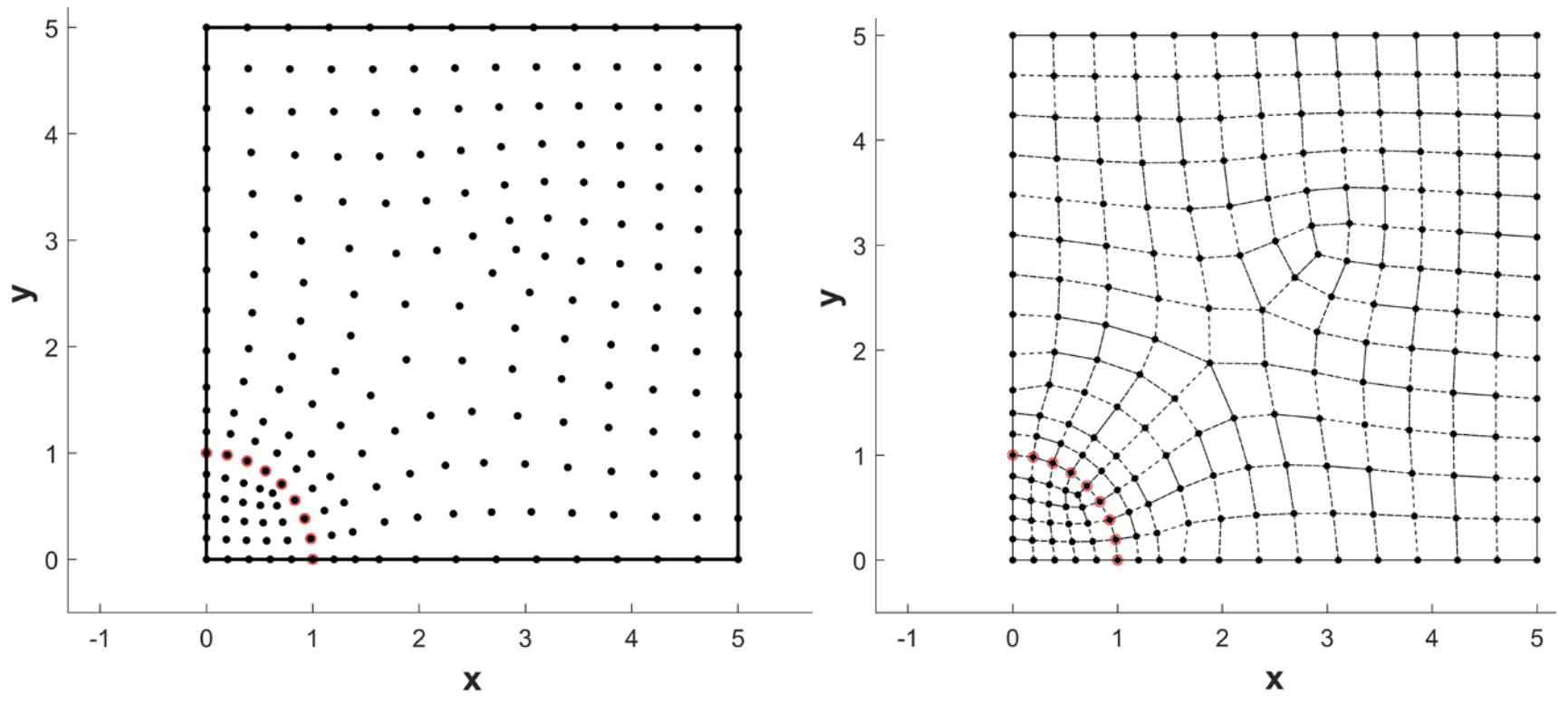

(a) Nodal arrangement

(b) Background integration cells

**Figure 32: Nodal arrangement and background integration cells for GI**

The approximated radial displacement, radial strain, and hoop strain solutions using RKPM, IM-RKPM with cubic B spline interface kernels, and IM-RKPM with fourth-order power interface kernels, accompanied with 224 non-uniform nodes and $5 \times 5$ GI, are plotted along the line $y = x$ in Figure 33. Like the 1D composite rod example, the RKPM solution of the radial strain and hoop strain are both oscillatory near the interface. IM-RKPM, on the other hand, effectively alleviates the oscillations in the strain solutions. In addition, a convergence study is performed for both RKPM and IM-RKPM with different interface kernels, and the results are shown in Figure 34. The IM-RKPM recovers the optimal convergence rates with the Gauss domain integration for both smooth and $C^0$ interface kernels.

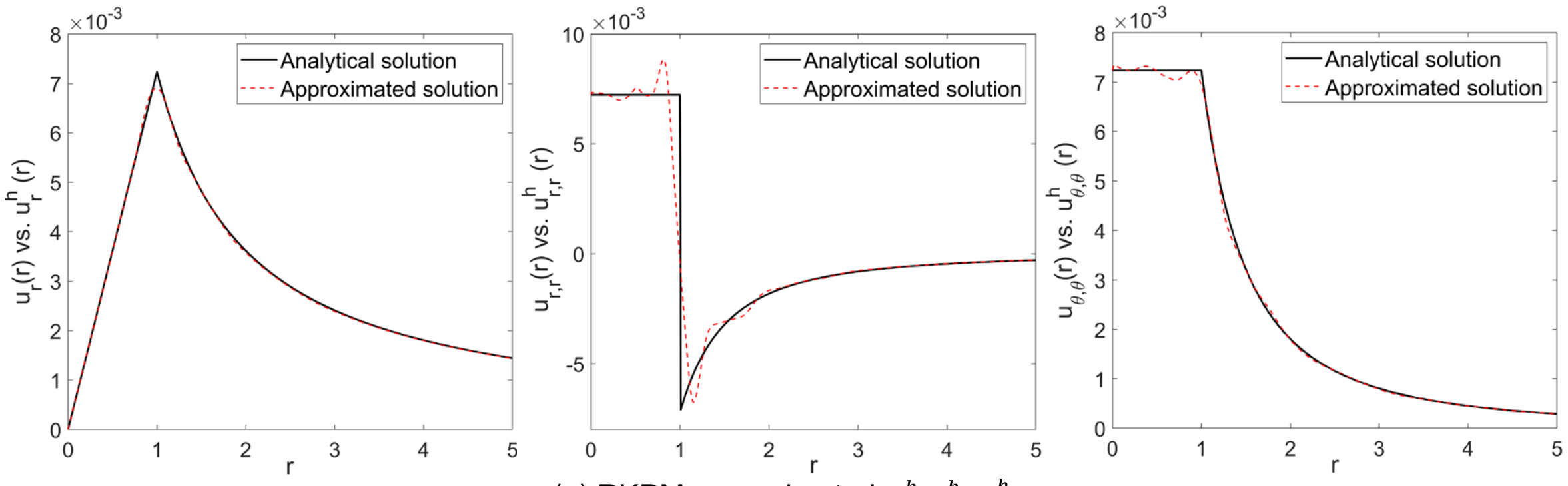

(a) RKPM approximated $u_r^h, \varepsilon_{rr}^h, \varepsilon_{\theta\theta}^h$

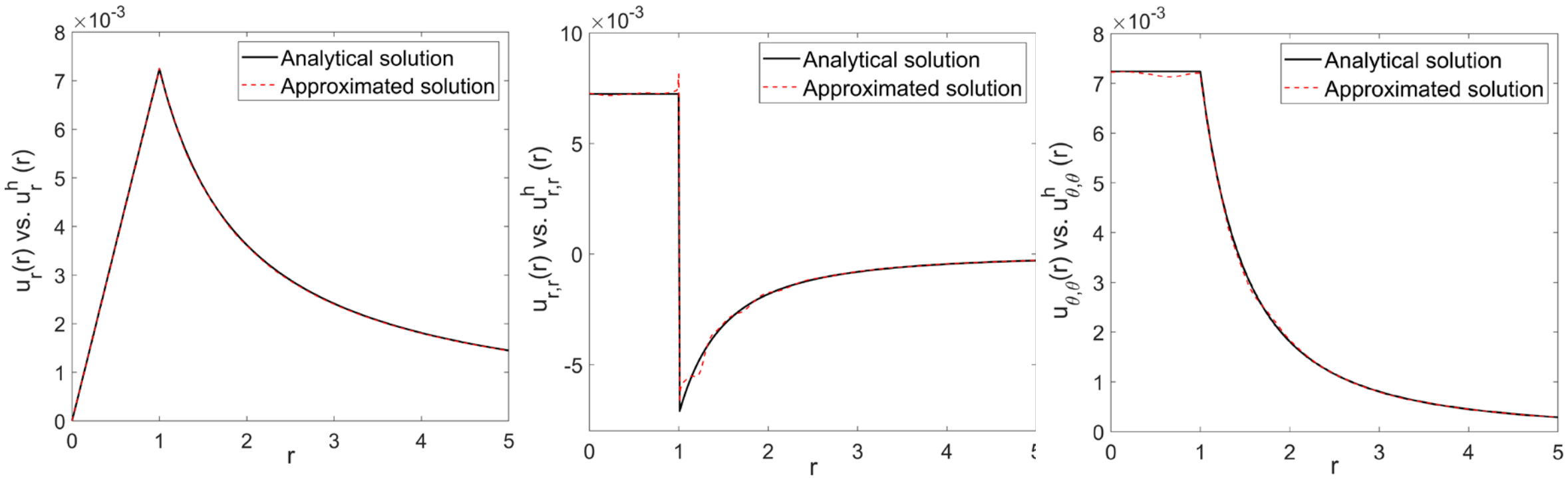

(b) IM-RKPM with B3 approximated $u_r^h, \varepsilon_{rr}^h, \varepsilon_{\theta\theta}^h$

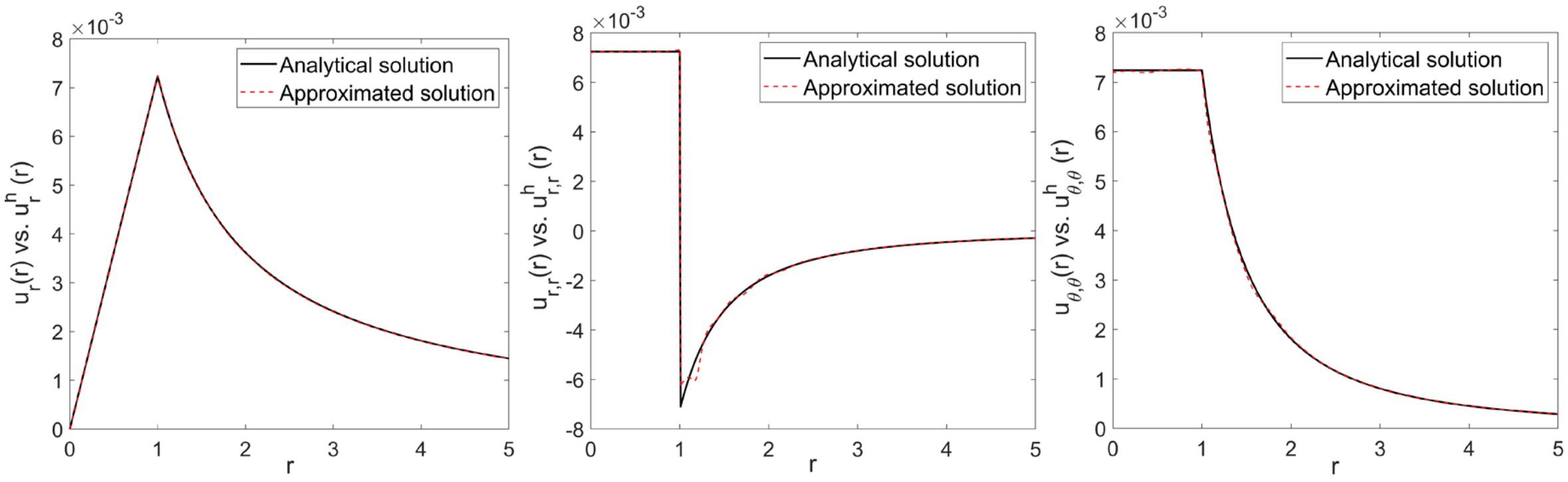

(c) IM-RKPM with PK4 approximated $u_r^h, \varepsilon_{rr}^h, \varepsilon_{\theta\theta}^h$

**Figure 33: $u_r^h, \varepsilon_{rr}^h, \varepsilon_{\theta\theta}^h$ approximated using RKPM and IM-RKPM with GI**

Next, the same problem is solved using the computationally efficient SCNI. Figure 35 demonstrates an arrangement of 211 non-uniformly distributed nodes and conforming strain smoothing cells for SCNI. The numerical solutions obtained by RKPM and IM-RKPM with different interface kernels are plotted in Figure 36.

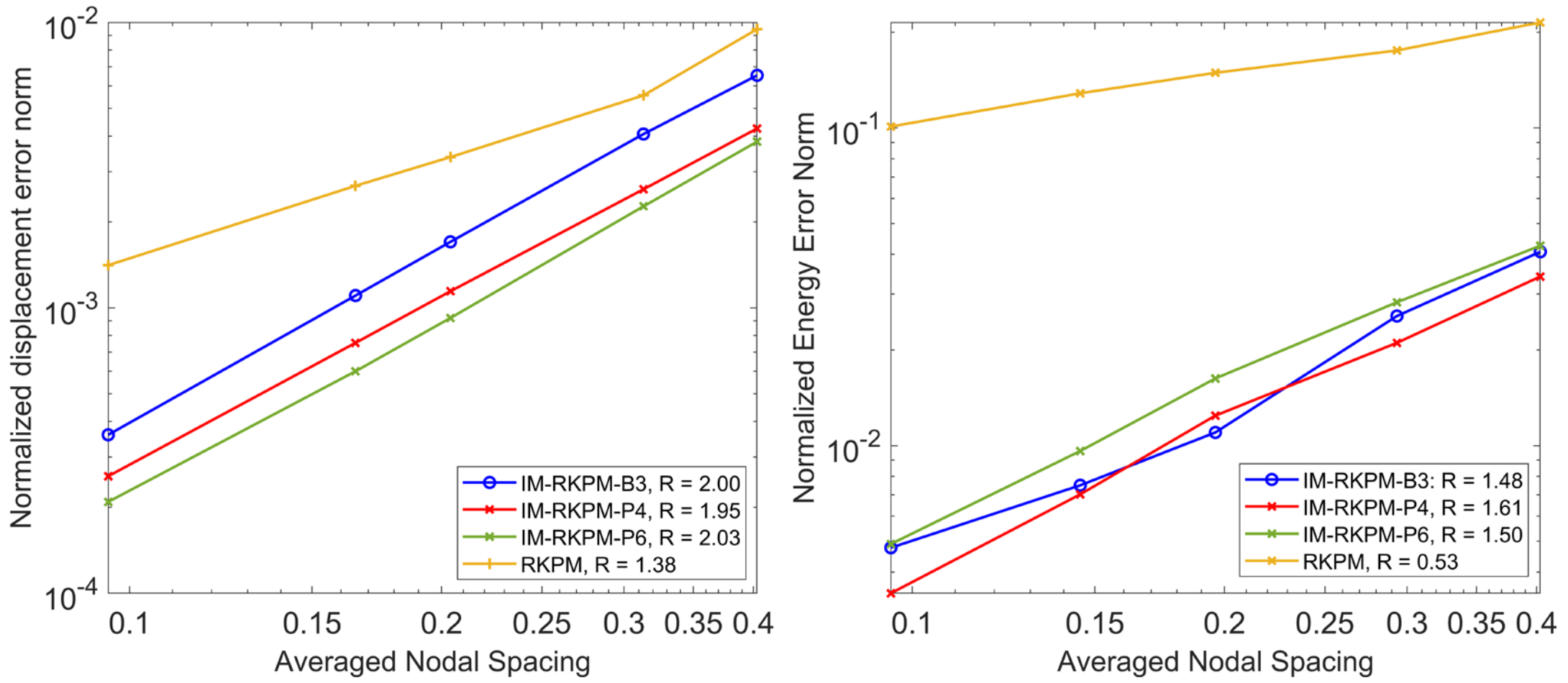

(a) Normalized displacement norms (b) Normalized energy norms

**Figure 34: Accuracy of RKPM and IM-RKPM with different interface kernels with GI** (R: rate of convergence)

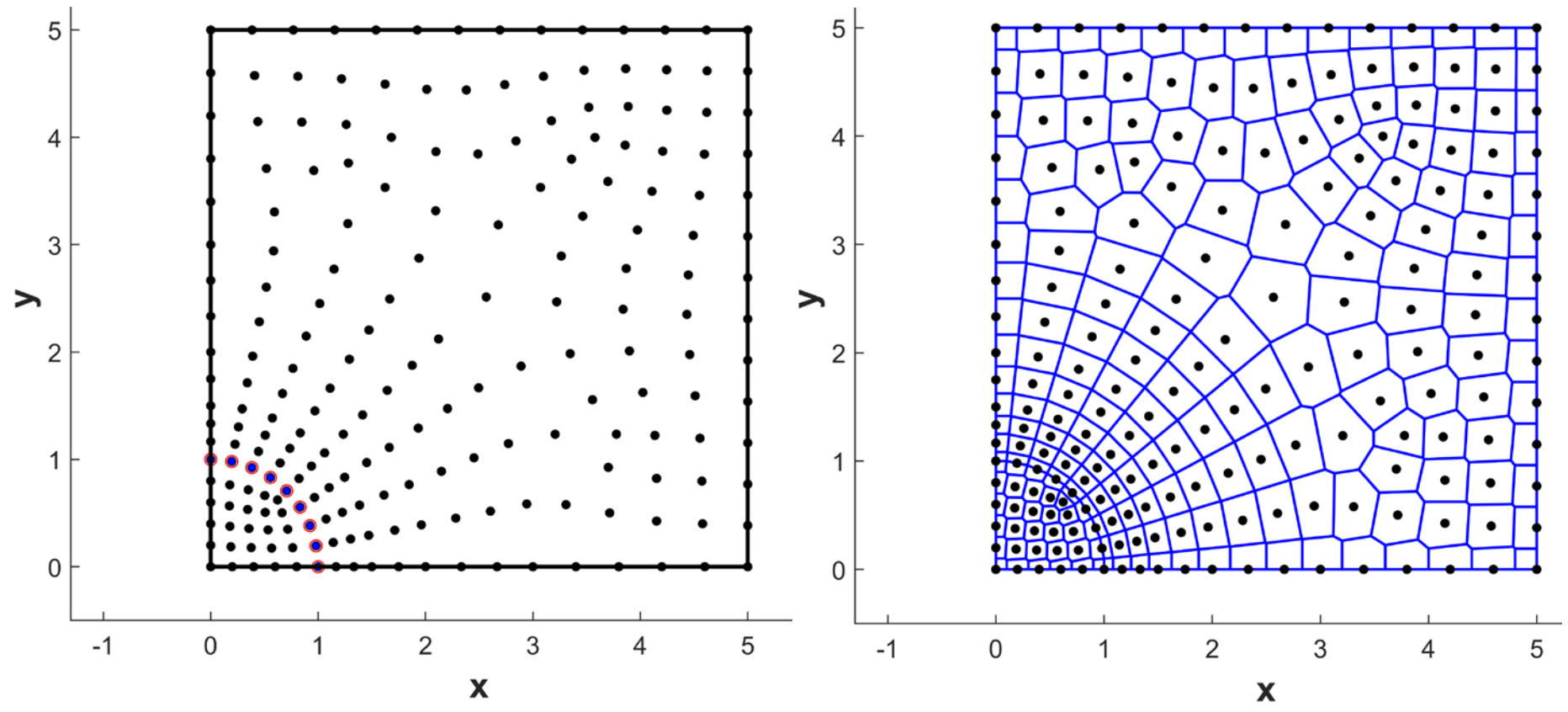

(a) Nodal arrangement (b) Conforming gradient smoothing cells

**Figure 35: Nodal arrangement and conforming strain smoothing cells for SCNI**

Similar to the Gauss integration, the standard RKPM with SCNI again experiences strain oscillations near the interface. However, solutions obtained by the IM-RKPM with different interface kernel functions are consistent with the ones obtained with Gauss integration in Figure 33. By observing results in the convergence plots shown in Figure 37, IM-RKPM with SCNI has displacement and strain solutions to converge optimally. These results show that the proposed IM-RKPM performs well with different selections of numerical domain integration techniques.

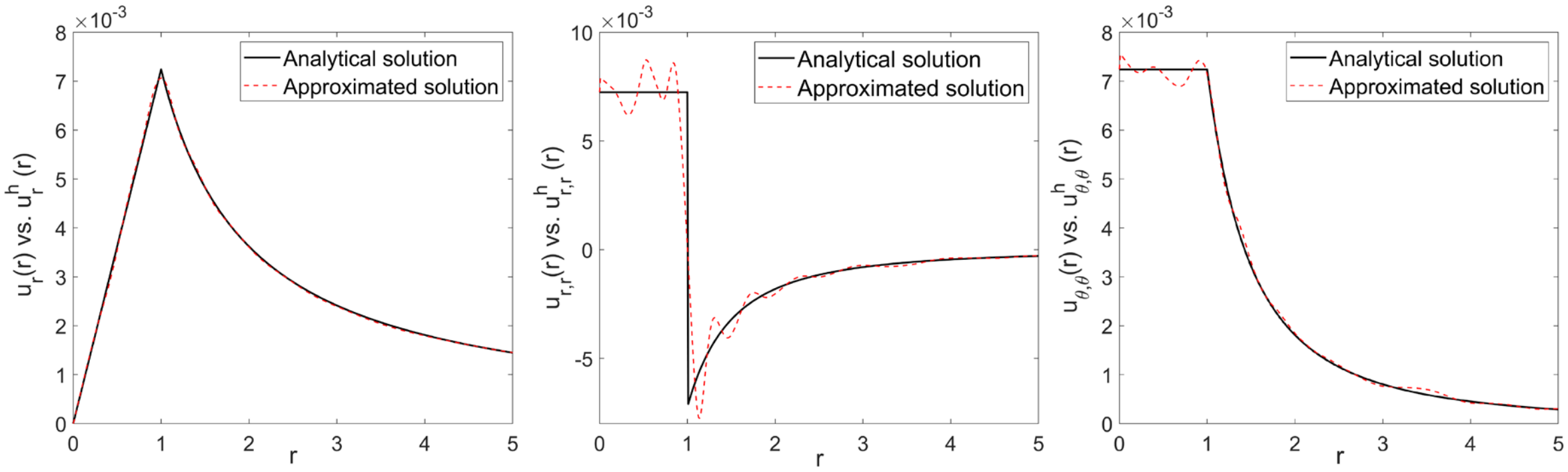

(a) RKPM approximated $u_r^h, \varepsilon_{rr}^h, \varepsilon_{\theta\theta}^h$

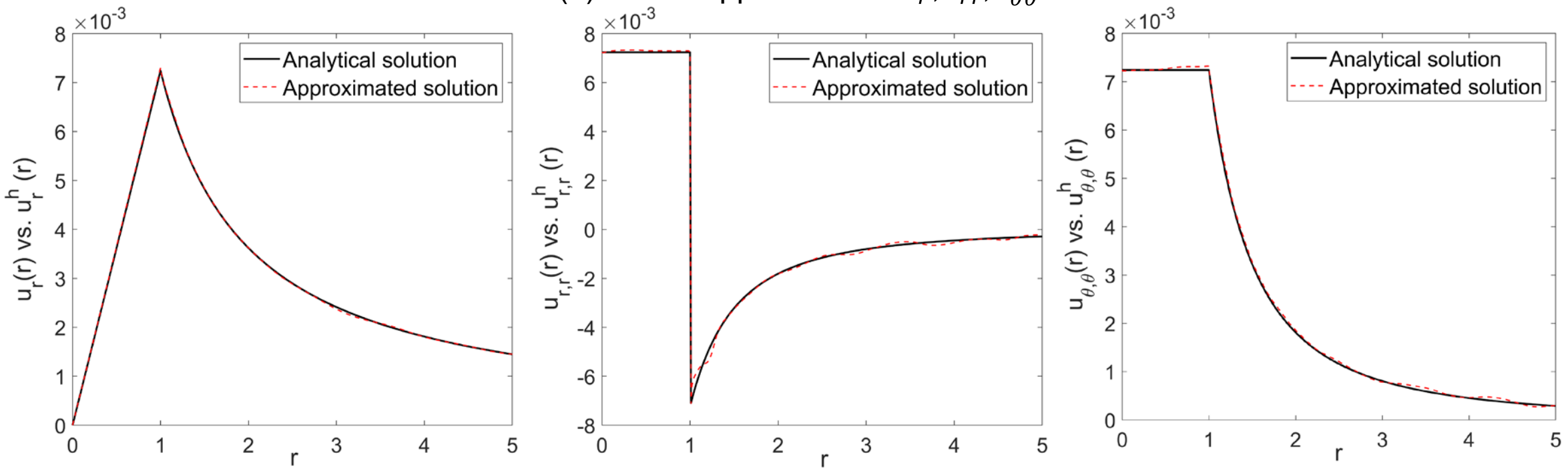

(b) IM-RKPM with B3 approximated $u_r^h, \varepsilon_{rr}^h, \varepsilon_{\theta\theta}^h$

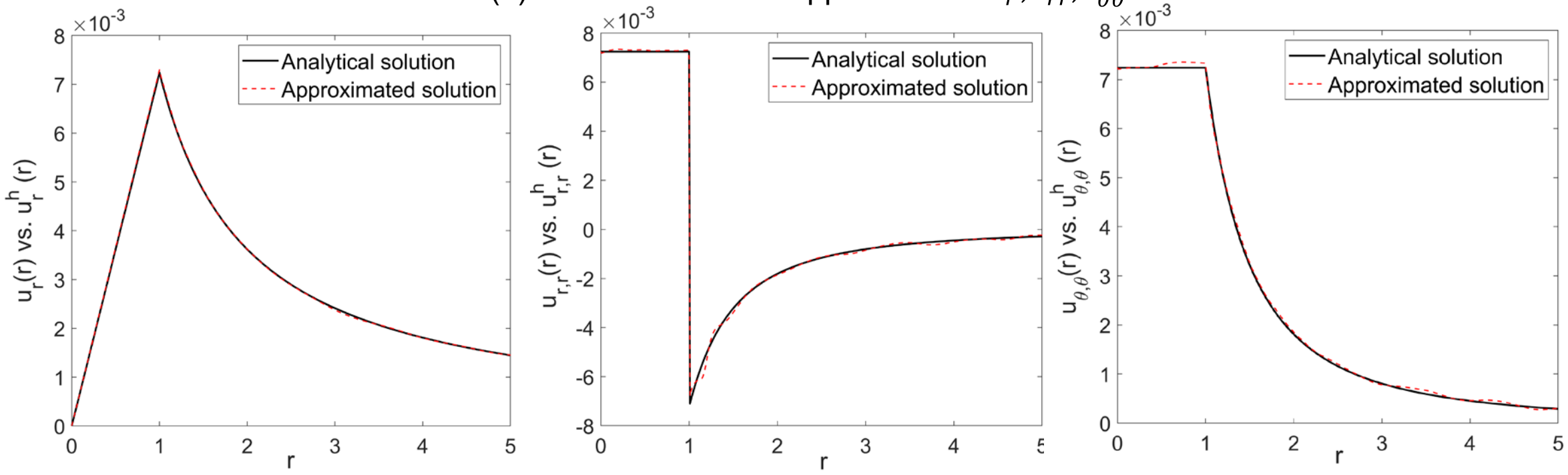

(c) IM-RKPM with PK4 approximated $u_r^h, \varepsilon_{rr}^h, \varepsilon_{\theta\theta}^h$

**Figure 36: $u_r^h, \varepsilon_{rr}^h, \varepsilon_{\theta\theta}^h$ approximated using RKPM and IM-RKPM with SCNI**

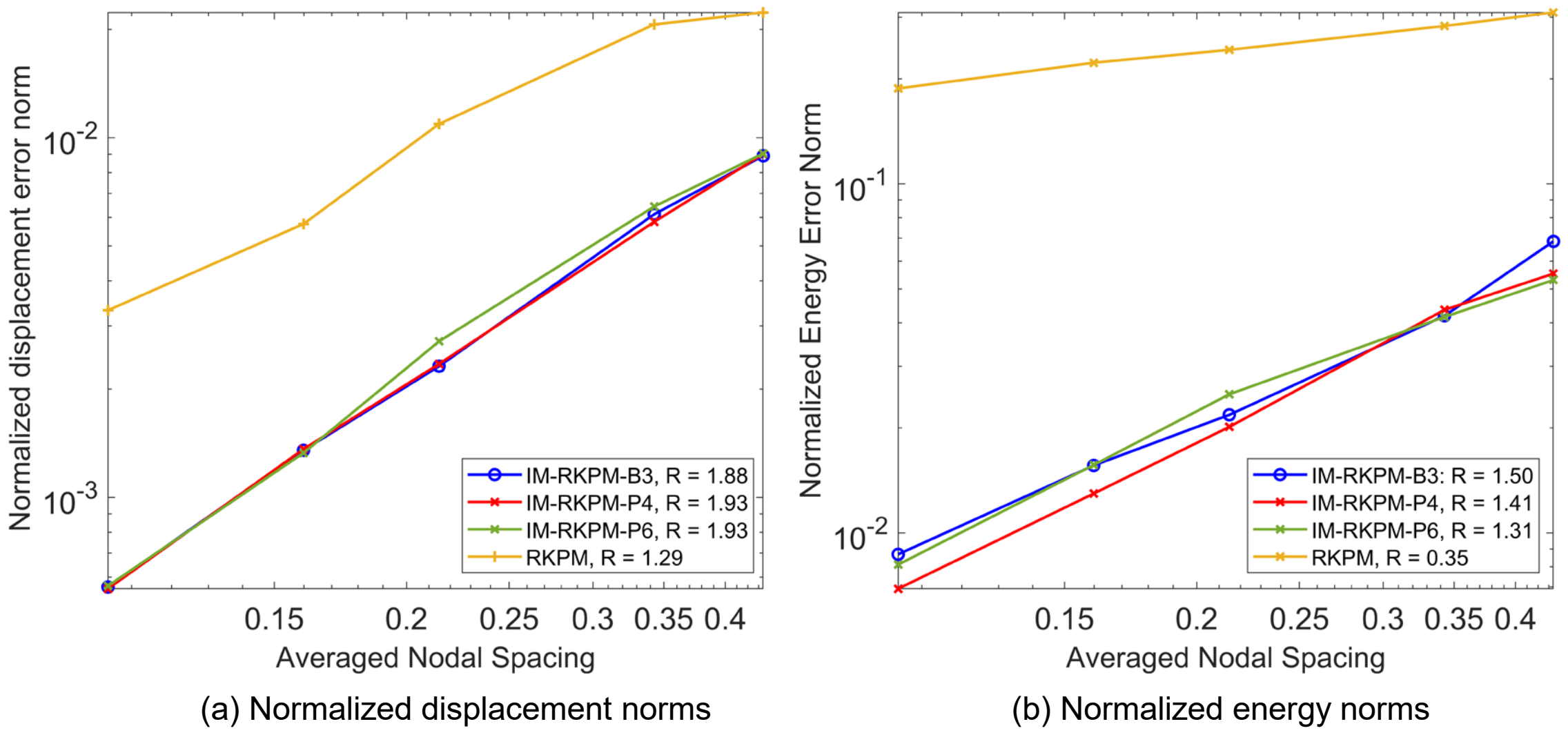

**Figure 37: Accuracy of RKPM and IM-RKPM with different interface kernels with SCNI**

(R: rate of convergence)

# 5 Image-Based Numerical Results

## 5.1 Compression-shear test on 2D composite microstructure

In this numerical example, a compression-shear test is conducted on a composite constructed based on the image shown in Figure 38. The image consists of $200 \times 200$ pixels with a pixel size of 8 μm. The physical dimensions of the specimen are 1.6 mm in width and height. The bottom edge of the specimen is fixed in both x- and y- directions, while the top edge is prescribed with a total displacement of $-0.01$ mm in both x- and y- directions. In addition, two vertical edges of the specimen are assigned as traction-free. The material constants for the alumina inclusion materials are: $E_1 = 320$ GPa, $\nu_1 = 0.23$, while the epoxy material is with material constants: $E_1 = 3.66$ GPa, $\nu_1 = 0.358$.

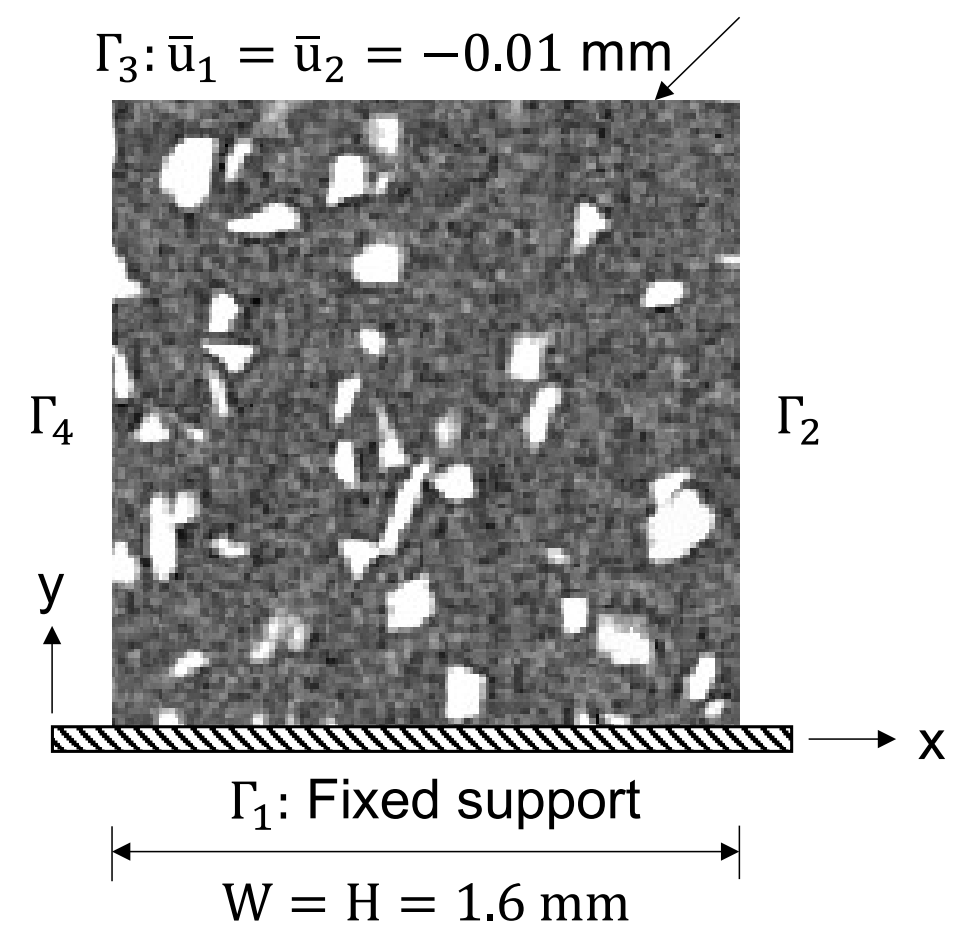

**Figure 38: Schematic of compression-shear test on a polymer-ceramic composite**

The problem at hand is examined using the proposed IM-RKPM and compared with the results produced using ANSYS [62], a commercially available FEM software with a much refined body-fitted mesh. All numerical analyses are performed under the 2D plane strain condition. The numerical model of the test image is constructed following Section 3.2, as shown in Figure 39, and the IM-RKPM approximation functions presented in Section 4 using cubic B-spline kernel with normalized support size of 2 and linear bases. The model employed for FEM analysis is manually traced from the inclusion geometries of the test image, resulting in a slight variation between the FEM and IM-RKPM discretization near interfaces. The FEM approximation involves a fine body-fitted mesh that comprises of 37,454 elements and 112,538 nodes, as illustrated in Figure 40. On the other hand, the IM-RKPM approximation uses only 11,316 nodes to discretize the image domain, which is approximately one-tenth of the number of nodes used in the FEM model. Figure 41 demonstrates the IM-RKPM and FEM approximated displacement solutions, respectively, and it is observed that both IM-RKPM and FEM predict similar displacements.

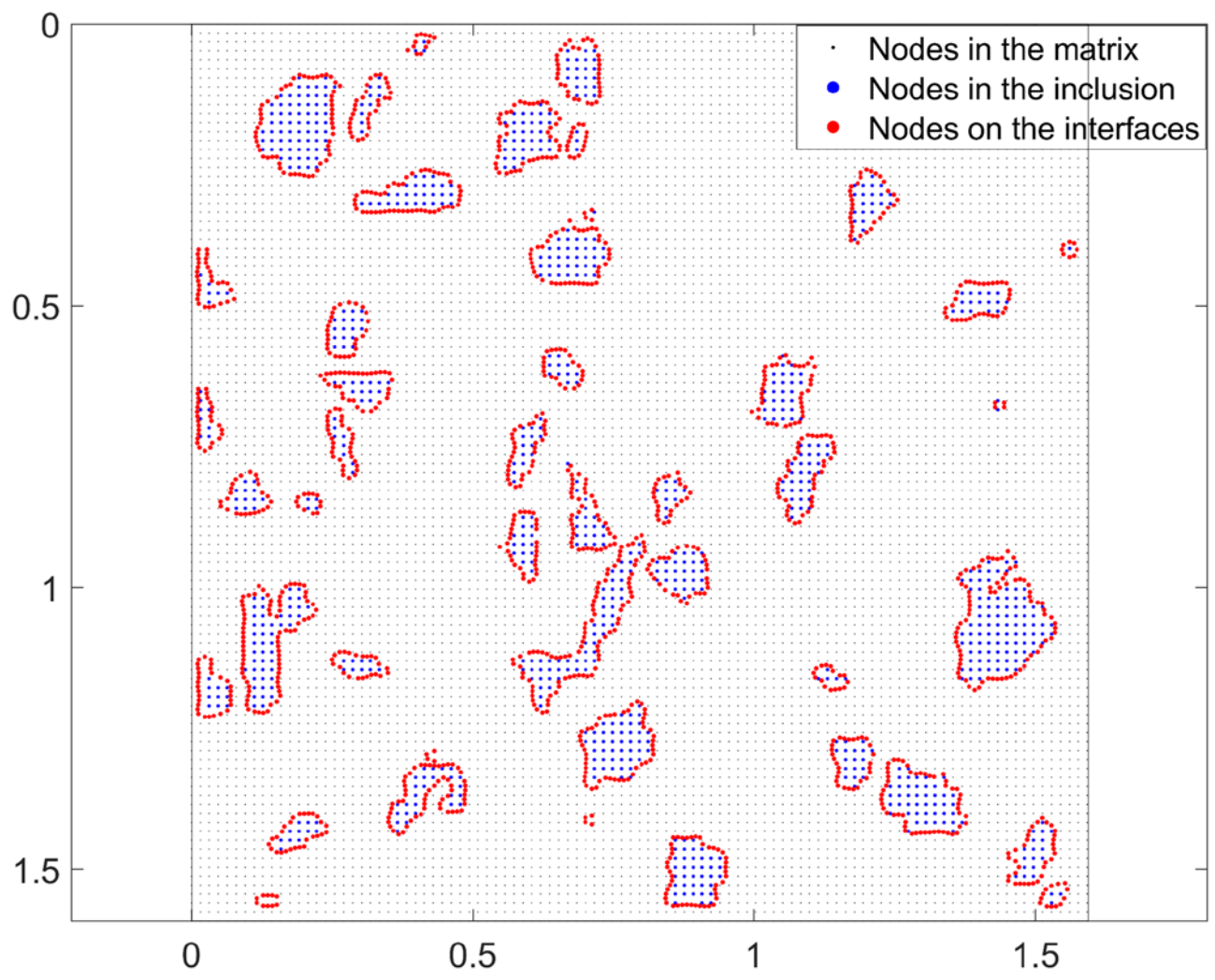

**Figure 39: Discretized RK numerical model for IM-RKPM simulation (Unit: mm)**

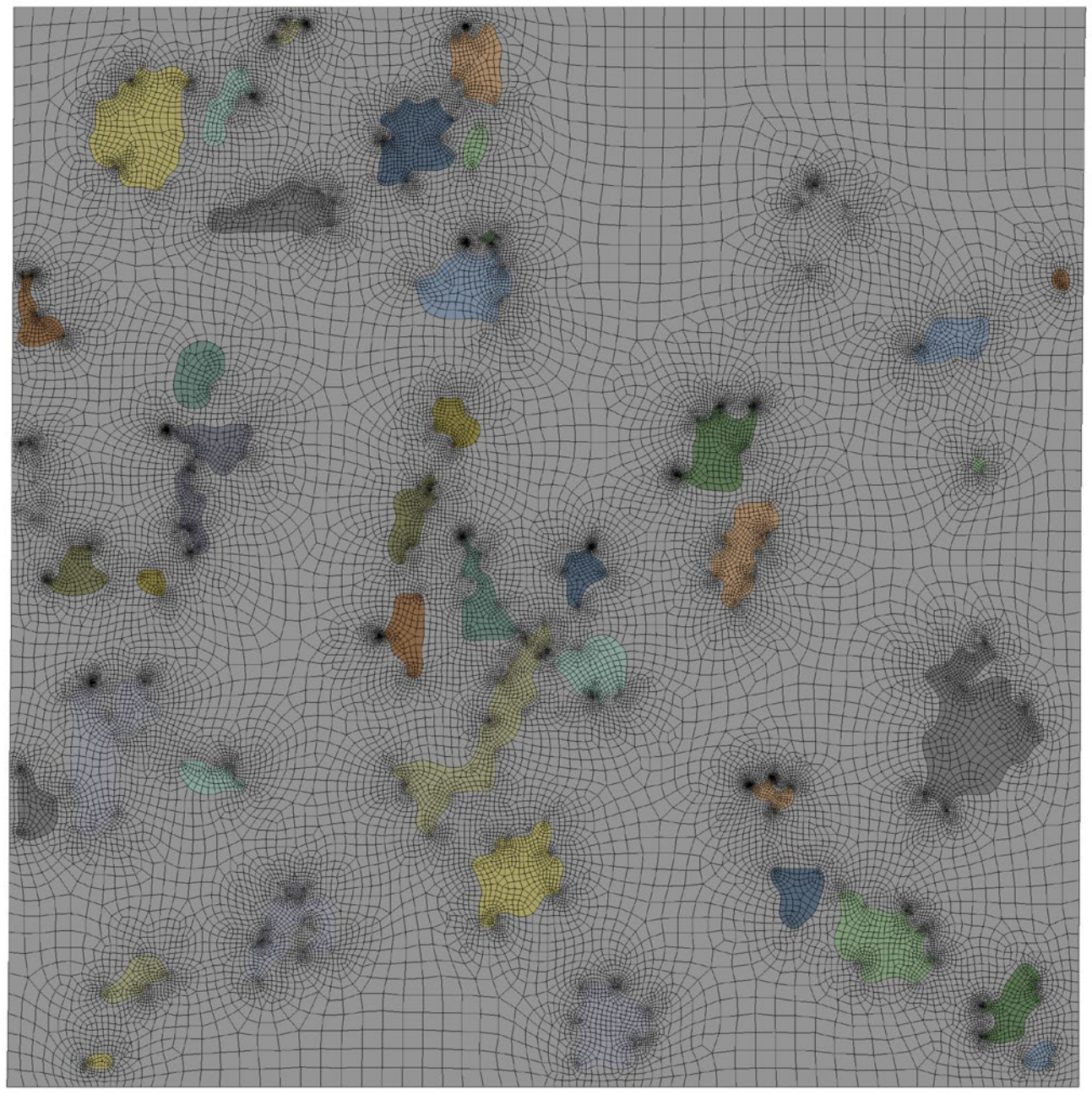

**Figure 40: FEM body-fitted mesh**

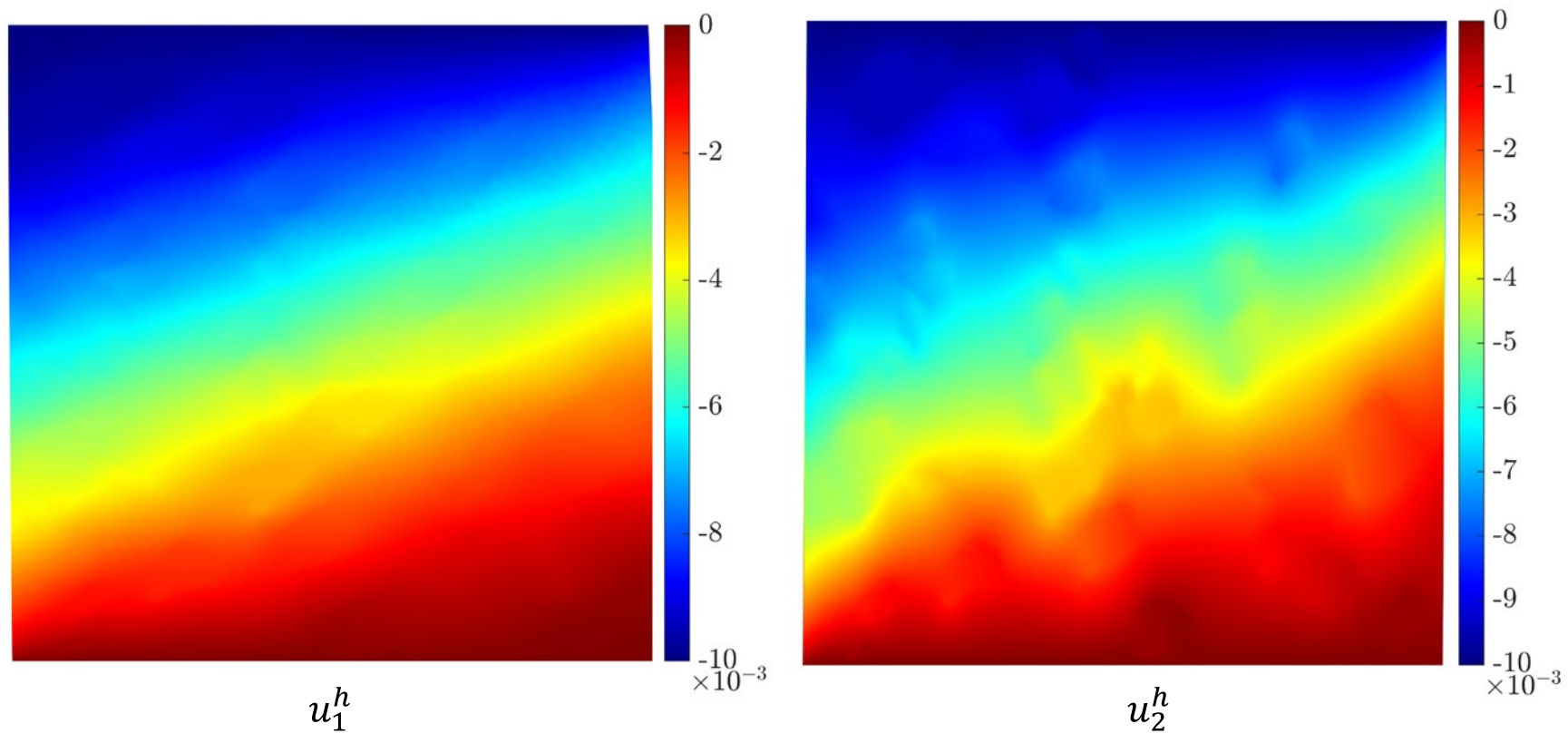

(a) Approximated displacement solutions using IM-RKPM

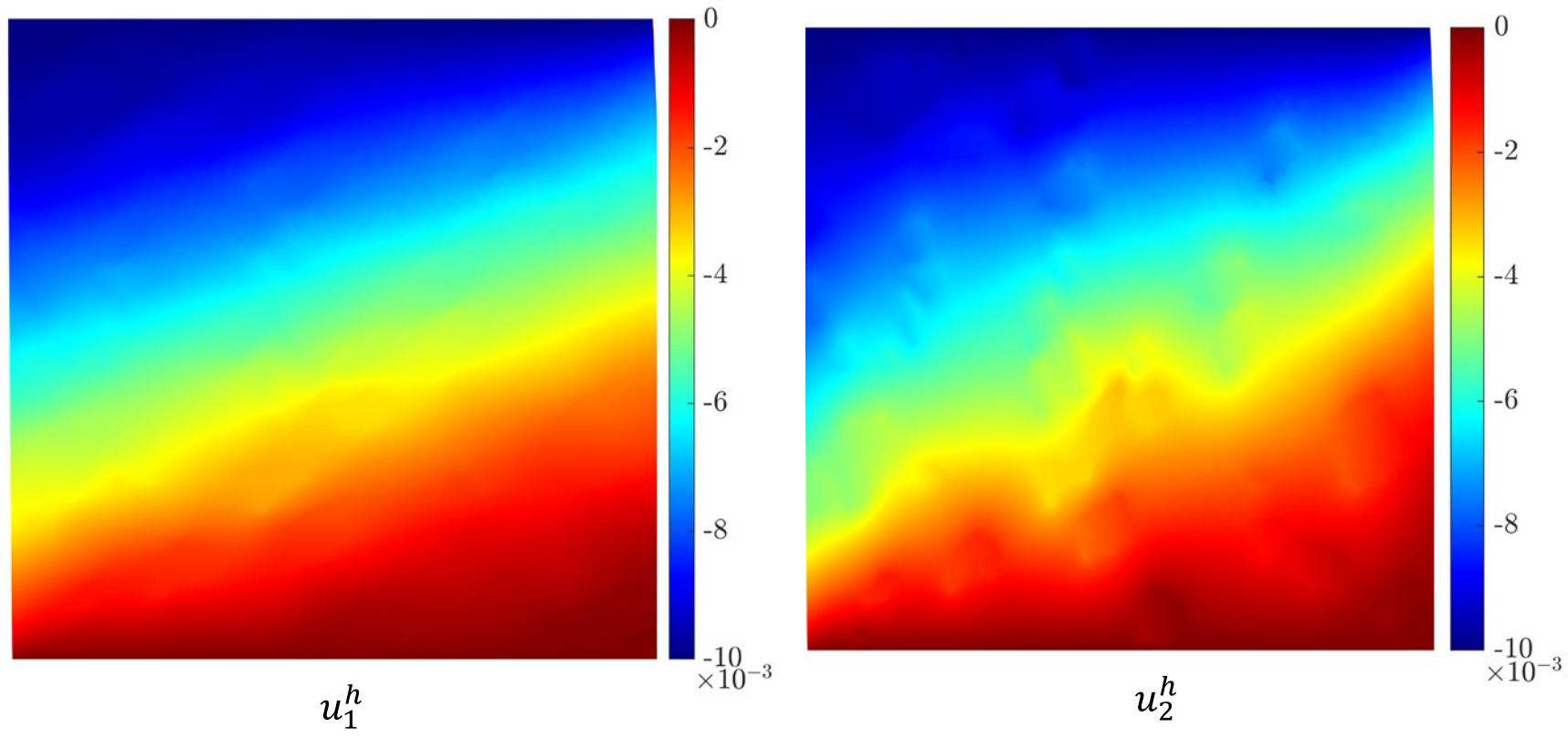

(b) Approximated displacement solutions using FEM

**Figure 41: IM-RKPM and FEM approximated displacement solution in both x- and y-directions (Unit: mm)**

Figure 42 shows the strains predicted by IM-RKPM and FEM, respectively. As shown in Figure 42, the strains of IM-RKPM display sharp transitions across the material interfaces and concentrated strains around the corners of the material interfaces, comparable to the results obtained using the FEM approximation. Furthermore, both the IM-RKPM and FEM approximations of strain solutions show some coalescence of the strain concentration around some closely positioned inclusions, as indicated by the boxed areas in Figure 42. The results show that the proposed IM-RKPM, accompanied by the SVM-based RK discretization with simple interface modified RK approximation functions, is capable of modeling composite materials with complicated microstructure and arbitrarily shaped inclusions with accuracy comparable to that obtained from a much refined and laborious FEM model.

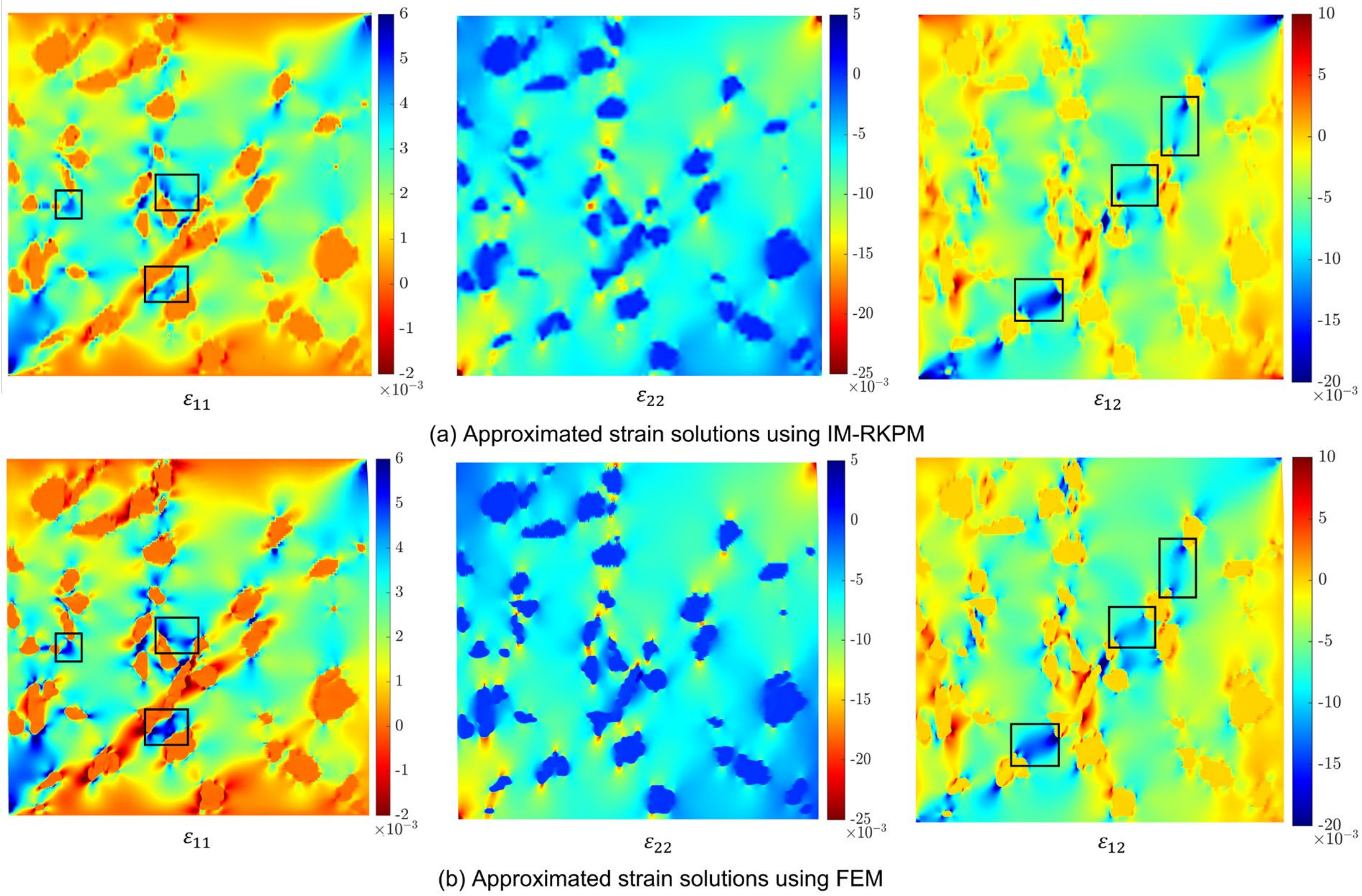

**Figure 42: IM-RKPM and FEM approximated strain solutions in plane view**

## 5.2 Uniaxial tensile test on 3D composite microstructure

In this example, a three-dimensional image-based SVM-RK model is constructed, and a uniaxial tensile test is conducted on the specimen's numerical model with the same material properties as those used in Example 5.1. The input of imaged-based 3D numerical model generation is performed by stacking 30 slices of ROI of 30 by 30 pixels extracted from reconstructed Micro-CT 2D images of a specimen's internal microstructure along the z-direction into a volumetric data matrix, as illustrated in Figure 43. The size of the test volume is $0.24\text{ mm } \times 0.24\text{ mm} \times 0.24\text{ mm}$, corresponding to an input image voxel size of $8\,\mu\text{m}$. The uniaxial tension is applied to the two surfaces with surface normal in the z-direction under prescribed displacements in the z-direction while without constraints in the x-y displacements.

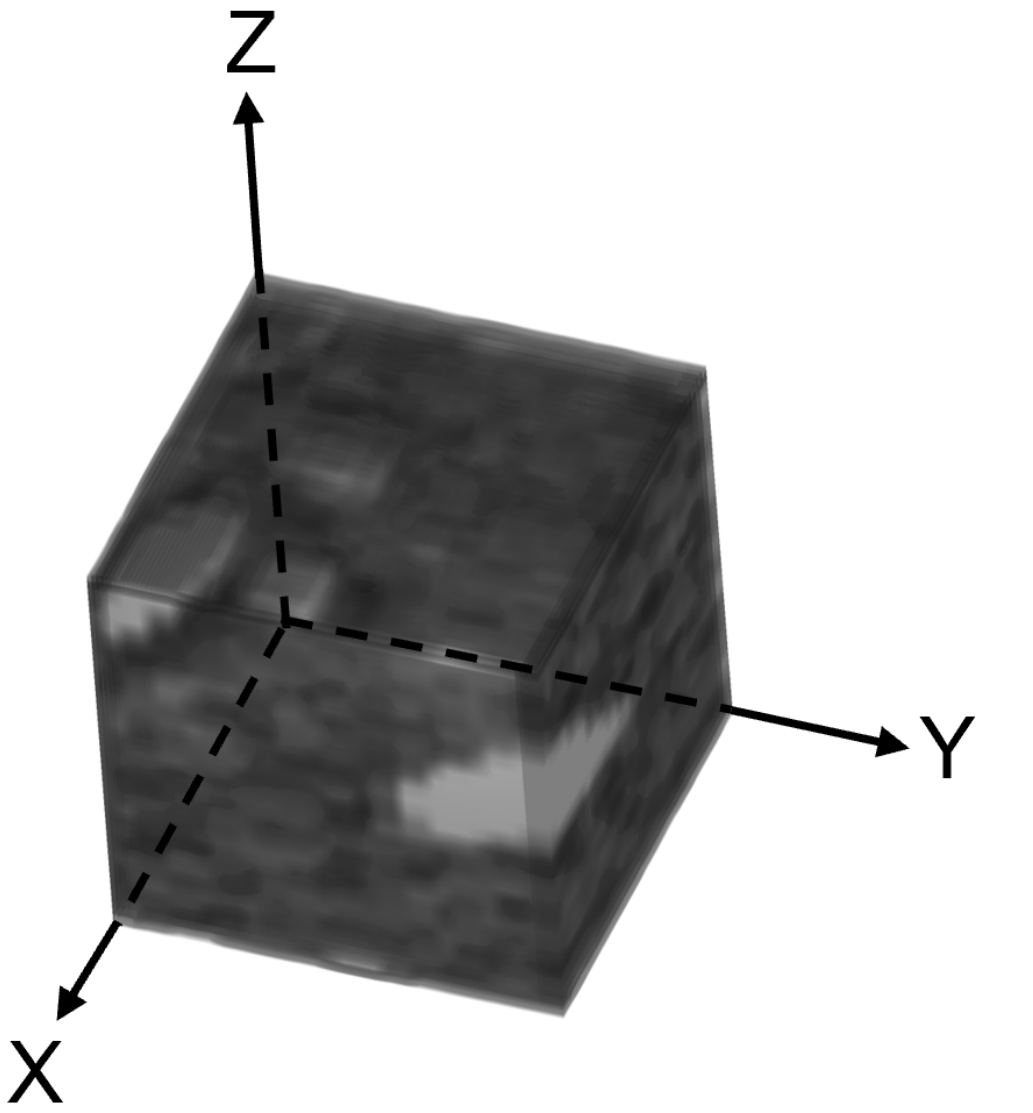

**Figure 43: 3D view of the input image volumetric data matrix**

Although the training data points are now in $\mathbb{R}^3$, the discrete model generation procedures remain the same as those for 2D, which corresponds to the physical coordinates of voxel centroids, as detailed in Section 3.2. The training response labels are obtained by stacking the segmented ROIs using Otsu's method into a binary volumetric data label matrix. It is important to note that both matrices that contain the training data points and training response labels are concatenated before being fed into the SVM, which means that the combined training data set can be represented as $\mathbf{D} = \{\boldsymbol{x}_i, y_i\}_{i=1}^{l}$, where $\boldsymbol{x}_i \in \mathbb{R}^3, y_i \in \{-1,1\}$, and $l = 27{,}000$ for the present case. For the SVM training, the same hyperparameters specified in Section 3.2.1 are utilized. Figure 44 demonstrates the SVM material classification results, the support vectors resulting from training, and the RK interpolated decision boundaries. It is worth mentioning that the resulting RK interpolated separating hyperplane, which is the material interface determined by the SVM training and RK interpolation, exhibits a smoother appearance in contrast to the inclusion geometries represented by binary label data volumetric matrix. This outcome is not surprising, given that SVM considers all three dimensions of the training data points to identify an appropriate separating hyperplane, which is more realistic and resembles a specimen image stack with higher resolution (i.e., with a smaller voxel size).

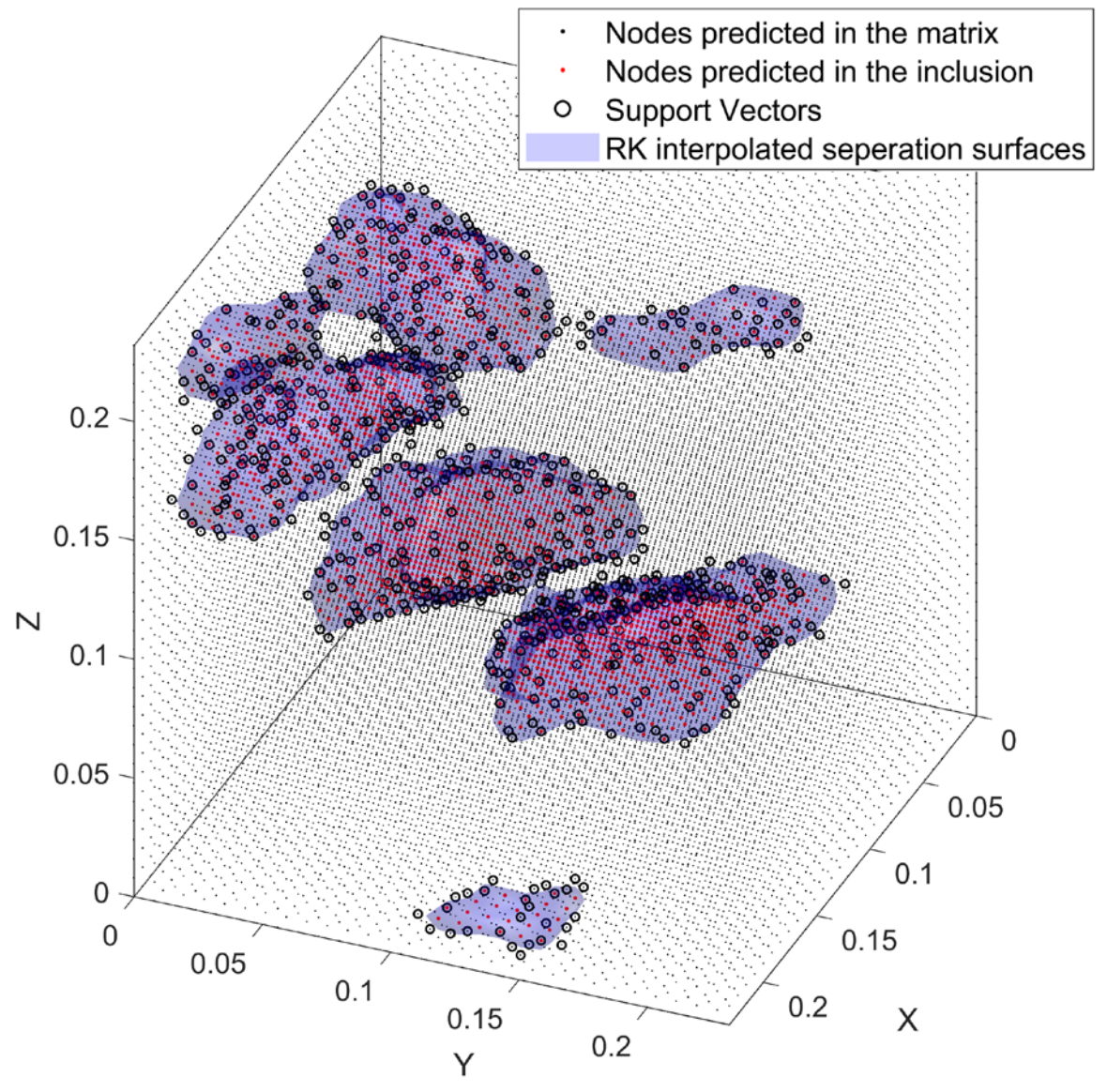

**Figure 44: SVM training and prediction results and RK interpolated decision boundaries (Unit: mm)**

Figure 45 illustrates the results of identified interface nodes and the 3D RK discrete model of the test volume, where the black (small) nodes, blue nodes, and red nodes represent the epoxy material points, alumina material points, and points on the material interfaces, respectively. The 3D SVM-RK discretization model contains in total 17,648 discretized nodes, among which 2330 nodes are on the material interfaces.

The produced 3D SVM-RK discrete model is utilized for a uniaxial tensile test in the z-direction. The model's bottom surface is fixed in all three directions, while a z-directional displacement of 0.01 mm is prescribed at the top surface of the model. The other surfaces of the specimen are assigned traction-free boundary conditions. The proposed IM-RKPM is employed for the numerical solution, and Figure 46 depicts the displacement solution in all three directions.

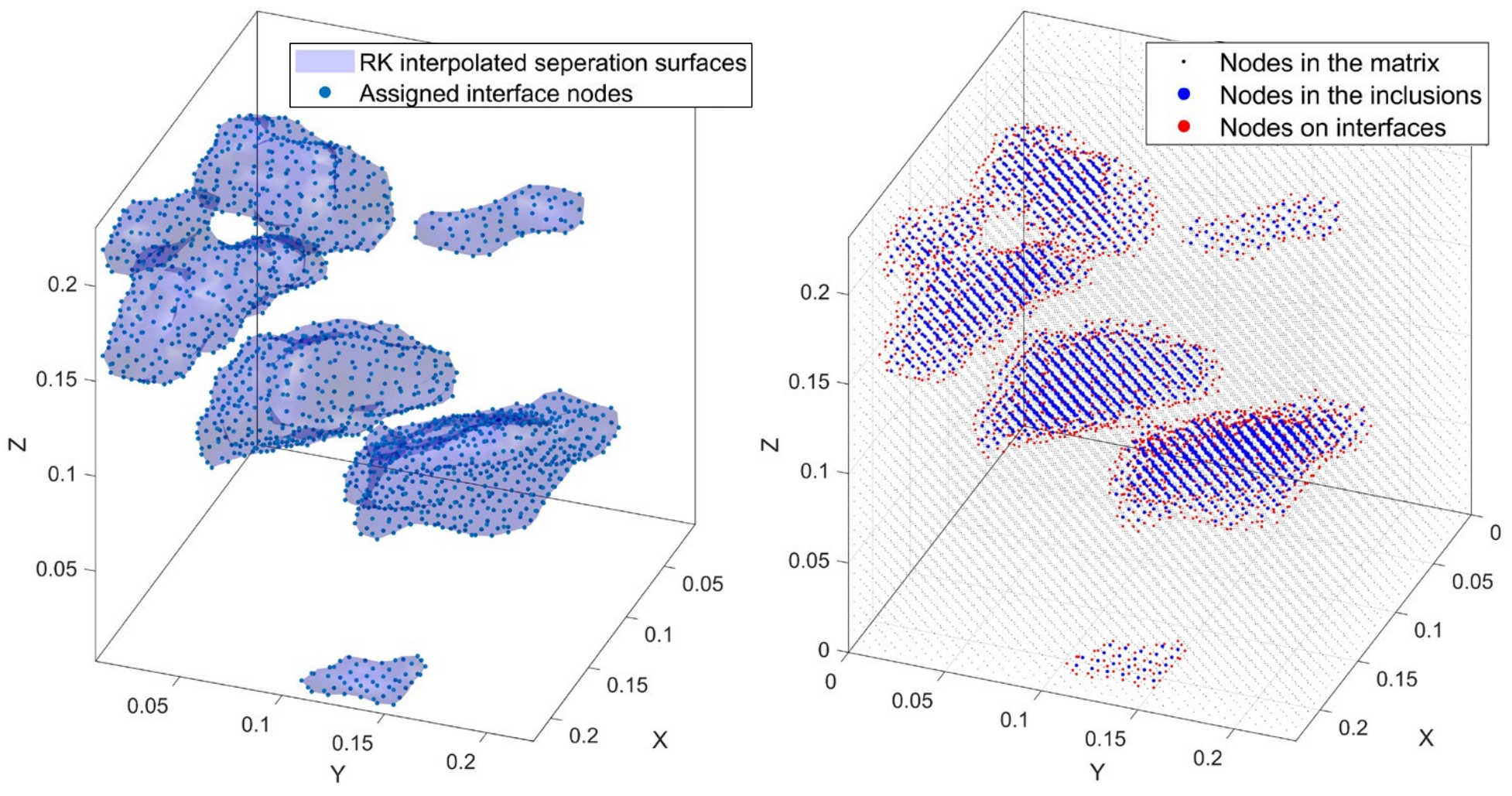

(a) 3D interface node assignment result (b) 3D RK discretized numerical model

**Figure 45: 3D interface node assignment result and RK discretized numerical model for the test volume (Unit: mm)**

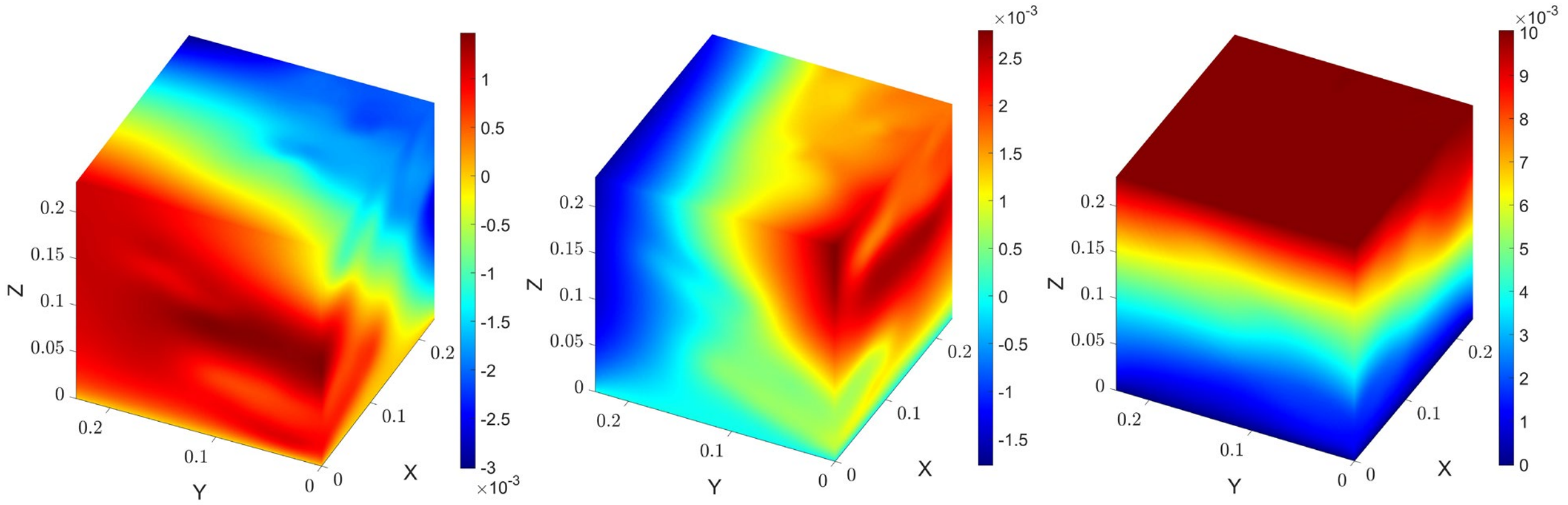

(a) $u_1^h$ (b) $u_2^h$ (c) $u_3^h$

**Figure 46: IM-RKPM approximated displacement solutions (Unit: mm)**

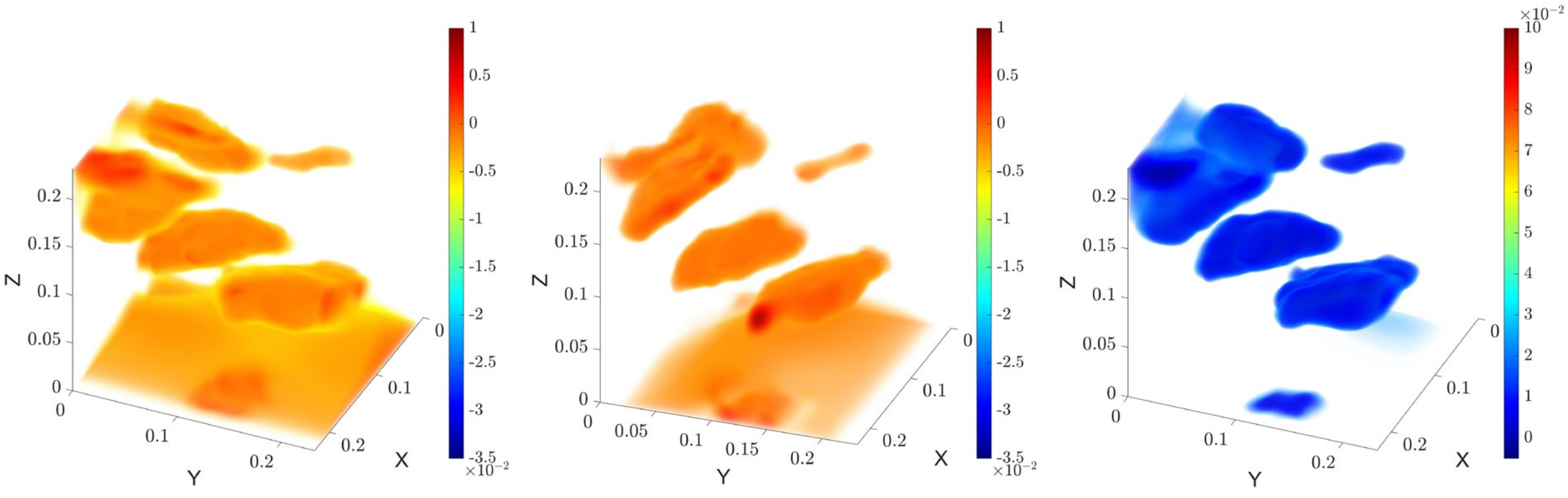

(a) $\varepsilon_{11}$ (b) $\varepsilon_{22}$ (c) $\varepsilon_{33}$

**Figure 47: IM-RKPM approximated normal strain solutions (Unit: mm)**

The predicted normal strains are plotted in Figure 47, to which a transparency filter is applied such that strain with a large magnitude is not visible. As shown in Figure 47, the regions with the relatively small magnitude of strains are consistent with the shapes of the alumina inclusions, which is expected because the alumina inclusions are significantly stiffer than the surrounding epoxy matrix. In Figure 48, Figure 49, and Figure 50, the normal strains are plotted on multiple slices and are compared with the slices of the inclusion contours. The results show that distinctive strain transitions in all three dimensions can be observed across interfaces. In addition, some strain concentrations are observed between two nearby inclusions and around the corner of the inclusions. Overall, this example demonstrates the capability of the proposed SVM-RK image-based model and IM-RKPM in modeling composite material with arbitrarily shaped inclusions in three dimensions.

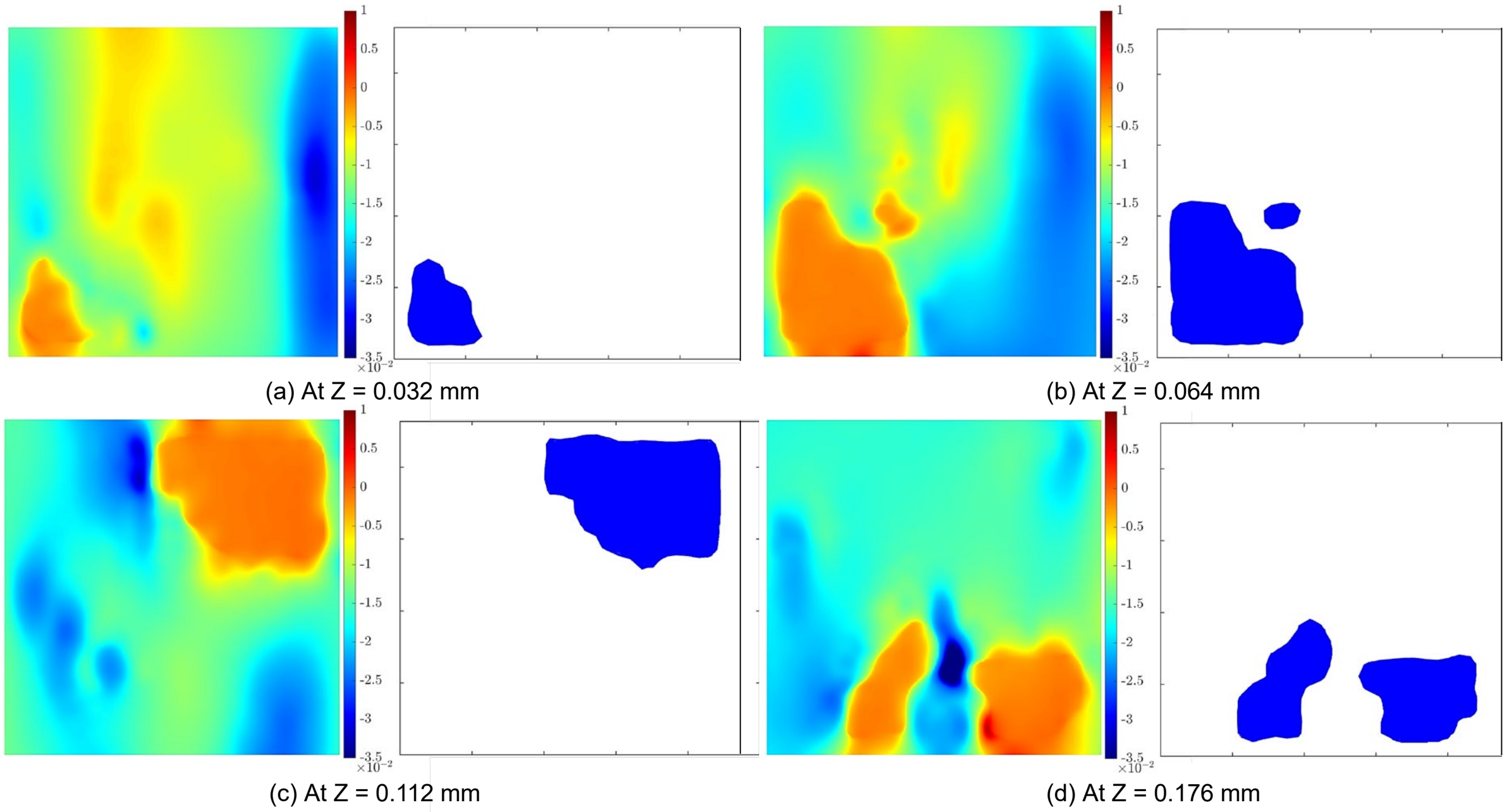

(a) At Z = 0.032 mm (b) At Z = 0.064 mm

(c) At Z = 0.112 mm (d) At Z = 0.176 mm

**Figure 48: Slices of x-directional normal strain results compared to the interface contours**

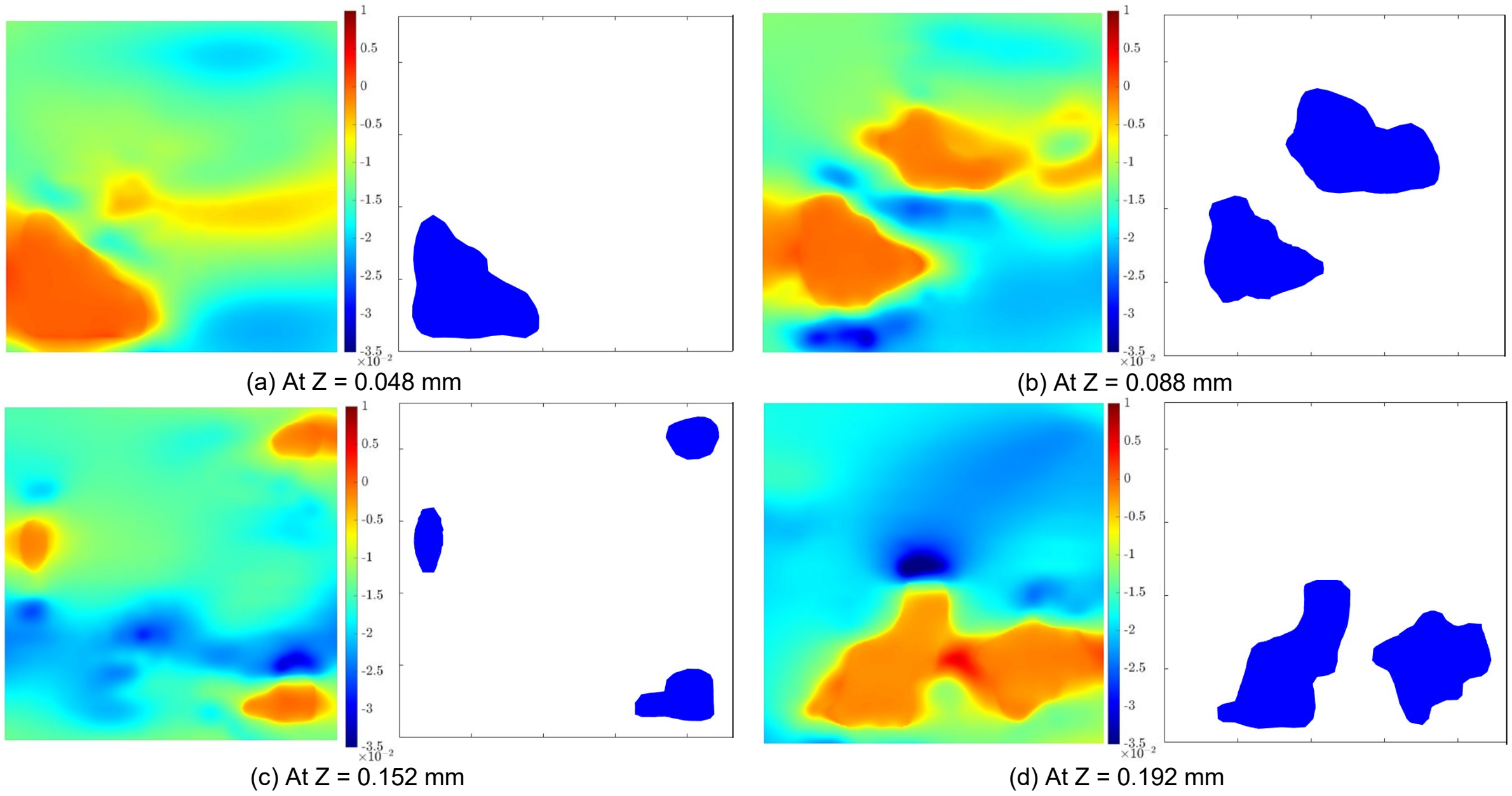

(a) At Z = 0.048 mm (b) At Z = 0.088 mm

(c) At Z = 0.152 mm (d) At Z = 0.192 mm

**Figure 49: Slices of y-directional normal strain results compared to the interface contours**

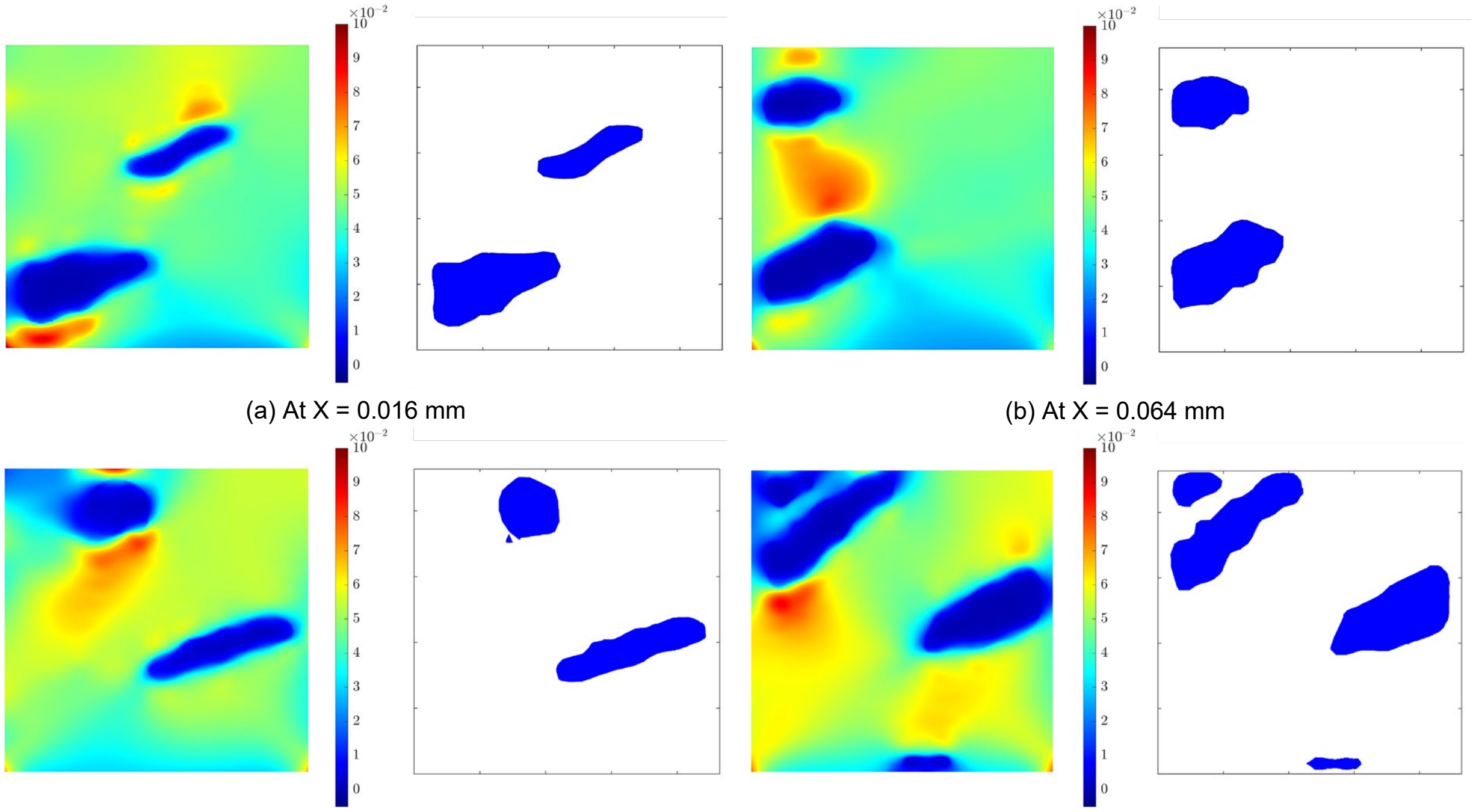

(a) At X = 0.016 mm (b) At X = 0.064 mm

(c) At X = 0.120 mm (d) At X = 0.200 mm

**Figure 50: Slices of z-directional normal strain results compared to the interface contours**

# 6 Conclusion

A Support Vector Machine (SVM) guided model discretization and reproducing kernel approximation, utilizing Micro-CT images of heterogeneous materials as input, is introduced in this work. The trained SVM-RK model generates classification scores for RKPM model discretization from the image, enabling their use as inputs for 1) interface node generation and 2) the interface kernel modification to construct a modified RK approximation of weak discontinuities. The SVM classification scores, representing the signed distances, enables identification of material phase, interface discretization, and interface surface normals, allowing automatic construction of RK approximation with weak discontinuities and interface-conforming gradient smoothing cells for SCNI based domain integration. The proposed image-based SVM-RK model generation process was validated through a synthetic image and a high-resolution surface image obtained from the SEM.

The resulting Interface-Modified Reproducing Kernel Particle Method (IM-RKPM) effectively remedy Gibb's oscillations commonly seen in the conventional RKPM for modeling problems with weak discontinuities. The proposed method incorporates a regularized Heavy-side function defined on the SVM classification score to achieve RK approximation with interface weak discontinuity while avoiding Gibb-type oscillations. These procedures involved in the proposed SVM guided RK approximation with interface weak discontinuities are fully automatic in 3-dimensions and without the need of using duplicated DOFs on the interface nodes common in other interface-enriched meshfree methods [25], [28]. In addition, this IM-RKPM with interface weak discontinuities can be constructed by kernel functions with arbitrary smoothness/roughness while achieving optimal convergence as demonstrated in the numerical examples using both Gauss integration and SCNI.

Finally, the effectiveness of the proposed automated SVM-RK model generation process in conjunction with the IM-RKPM method is demonstrated through numerical examples based on test Micro-CT images in both 2- and 3-dimensions. Notably, the 3D example shows that the proposed approach is applicable for 3D simulations where the SVM-RK model precisely represents the geometry of the inclusion particles, trained based on stacked image slices.

It is worth mentioning that the present work utilizes the standard binary SVM library. However, it is possible to explore more efficient SVM algorithms to enhance the efficiency of the proposed automated numerical model generation process.

## Acknowledgements

This research is supported by the National Science Foundation (#1826221). The authors are grateful to the program manager, Dr. Siddiq Qidwai, for his encouragement and support. Appreciation is extended to Dr. Timothy Stecko of Penn State Center for Quantitative Imaging for technical assistance with micro-CT scanning and Ms. Julie Anderson for technical assistance with SEM imaging.

## Appendix A. Construction of interface-conforming gradient smoothing cells

Since the interface locations are determined by the RK interpolated score function in Eq. (40), the outward unit normal of the interfaces at an interface point $\boldsymbol{x}_K^*$ can be calculated as follows:

$$\vec{\boldsymbol{n}}(\boldsymbol{x}_K^*) = \left. \frac{\boldsymbol{\nabla}\tilde{S}(\boldsymbol{x})}{\left\| \boldsymbol{\nabla}\tilde{S}(\boldsymbol{x}) \right\|} \right|_{\boldsymbol{x}=\boldsymbol{x}_K^*} \tag{55}$$

To construct interface-conforming gradient smoothing cells for SCNI domain integration, a mirroring technique is utilized. For all interface nodes $\boldsymbol{x}_K^* \in \mathbb{S}^{IF}$, the mirrored node pair $\{\boldsymbol{x}_K^{*+}, \boldsymbol{x}_K^{*-}\}$ is obtained as follows:

$$\boldsymbol{x}_K^{*\pm} = \boldsymbol{x}_K^* \pm \epsilon \vec{\boldsymbol{n}}(\boldsymbol{x}_K^*) \tag{56}$$

where $\epsilon$ is a small perturbation number, and the interface normal $\vec{\boldsymbol{n}}(\boldsymbol{x}_K^*)$ is defined in Eq. (55). In this work, $\epsilon = 10^{-3}\ell$ is chosen, where $\ell$ is the image voxel size. As an illustration example, mirrored nodes are shown in the black box in Figure 51 (a), and the resulting gradient smoothing cells are shown in Figure 51 (b). Note that these mirrored node pairs are only used to generate the "interface conforming" Voronoi cells; they are not the RK nodes, and they don't carry degrees of freedom.

This approach allows for the use of conventional techniques, such as Voronoi tessellation, which will result in the two smoothing cells adjacent to either side of the interface having a common boundary along the interface location as shown in Figure 51 (b). In addition, the material class for the smoothing cells can be assigned according to the centered mirrored nodes' material classes. As a result, smoothing cells away from the material interfaces are uniformly arranged, and the material interfaces are well represented by the adjacent two layers of smoothing cells.

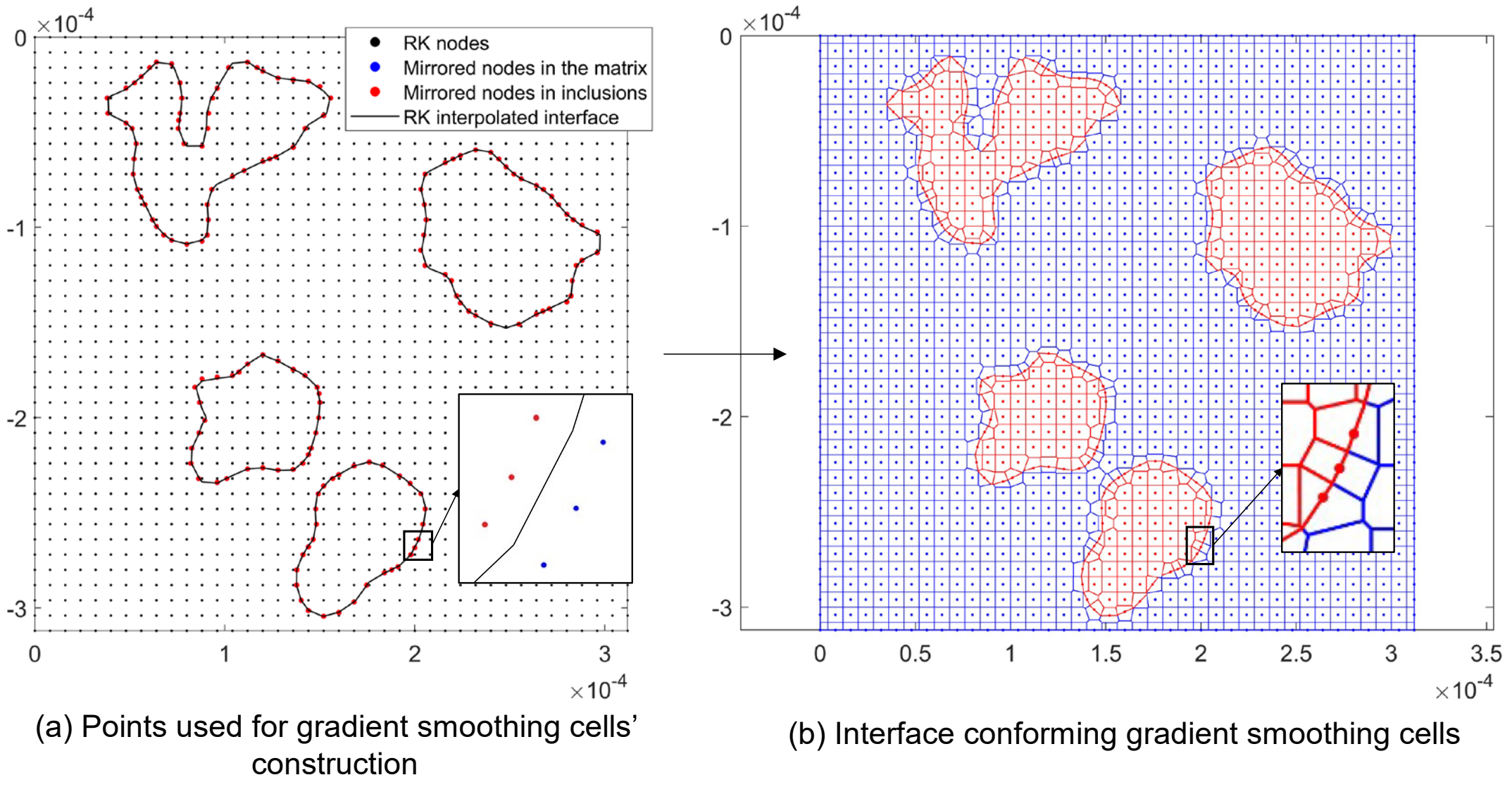

**Figure 51: Interface conforming gradient smoothing cells' construction** (Perturbation distances are magnified $\mathbf{10^2}$ times for mirrored nodes in (a) for demonstration purpose)